\documentclass[letterpaper]{article} 

\usepackage{aaai2027}  
\copyrighttext{\textit{Preprint}. Copyright \copyright\ 2026 by the authors.}

\usepackage[hyphens]{url}  
\usepackage{graphicx} 
\usepackage{natbib}  
\usepackage{caption} 
\usepackage{algorithm}
\usepackage{algorithmic}

\usepackage{newfloat}
\usepackage{listings}
\usepackage[most]{tcolorbox}

\DeclareCaptionStyle{ruled}{
    labelfont=normalfont,
    labelsep=colon,
    strut=off
} 

\floatstyle{ruled}
\newfloat{listing}{tb}{lst}{}
\floatname{listing}{Listing}

\usepackage{booktabs}
\usepackage{tabularx}
\usepackage{amsmath}
\usepackage{amssymb}
\usepackage{multirow}
\usepackage{array}

\definecolor{reliclinkgreen}{RGB}{0,160,80}
\usepackage[
    colorlinks=true,
    linkcolor=reliclinkgreen,
    citecolor=reliclinkgreen,
    urlcolor=reliclinkgreen,
    pdfborder={0 0 0},
    bookmarksopen=true
]{hyperref}

\newsavebox{\relictablebox}
\newcommand{\fitTable}[1]{%
    \sbox{\relictablebox}{#1}%
    \ifdim\wd\relictablebox>\linewidth
        \resizebox{\linewidth}{!}{\usebox{\relictablebox}}%
    \else
        \usebox{\relictablebox}%
    \fi
}

\newcommand{\relicTableSetup}{%
    \fontsize{6.5pt}{7.4pt}\selectfont
    \setlength{\tabcolsep}{2.5pt}%
    \renewcommand{\arraystretch}{1.0}%
}

\newcommand{\relicOneColumnTableSetup}{%
    \fontsize{6pt}{7pt}\selectfont
    \setlength{\tabcolsep}{1.2pt}%
    \renewcommand{\arraystretch}{1.0}%
}

\title{
RELIC: Revealed Principles for Learning\\
Interpretable Composable Skills in Multi-Agent Planning
}

\author{
Nguyen Viet Tuan Kiet$^1{}^\star$,
Bui Dinh Pham$^1{}^\star$,
Duong Quoc Chinh$^1{}^\star$,\\
Dao Van Tung$^1$,
Tran Cong Dao$^2{}^\dagger$,
Huynh Thi Thanh Binh$^1{}^\dagger$
}

\affiliations{
$^\star$Co-first authors
$^1$Hanoi University of Science and Techonology, Hanoi, Vietnam\\
$^\dagger$Corresponding authors
$^2$University of Sydney, Sydney, Australia\\
\texttt{ctra0528@uni.sydney.edu.au, binhht@soict.hust.edu.vn}
}

\begin{document}

\maketitle
\pagestyle{plain}
\thispagestyle{plain}


\begin{abstract}
Multi-agent planning becomes substantially harder when agents must improve specialized decision-making skills while keeping their executable implementations private. This setting arises when independently developed agents expose heterogeneous interfaces, observations, and capabilities, yet must coordinate under a shared team objective. Existing approaches commonly rely on centralized optimization, shared policy access, or common skill representations, assumptions that limit knowledge reuse when function signatures differ. We introduce RELIC, a framework for learning interpretable and composable programmatic skills through \emph{revealed principles}. Each agent improves its own executable skill locally, while useful decision logic and coordination patterns are distilled into compact textual principles. Rather than requiring direct program exchange, these abstractions can be re-instantiated under agent-specific interfaces and reused across incompatible implementation spaces. A shared principle memory accumulates transferable knowledge and promotes abstractions that repeatedly improve team-level performance. This separation allows discoveries made by one agent to guide others while preserving local executable implementations and decentralized execution. RELIC therefore supports strategic transfer across both heterogeneous-role and shared-role cooperative teams. Extensive experiments across routing, scheduling, combinatorial optimization, and distributed coordination settings demonstrate RELIC's effectiveness against independent and joint LLM-based search methods, together with consistent benefits across task structures and LLM backbones.
\end{abstract}

%


\section{Introduction}

Large language models (LLMs) are increasingly used to construct structured artifacts for planning and decision making. They can generate executable policies and programs, refine symbolic models, and support long-horizon behavior through reusable decision procedures \citep{liang2023code,silver2024generalized,oswald2024domain}. In parallel, recent work on skill discovery and skill libraries has shown that reusable behaviors can be learned, stored, retrieved, and composed across tasks rather than regenerated for every instance \citep{zhang2023bootstrap,zhu2026offline,wang2026skillx,li2026graph}. These developments make programmatic skills particularly attractive: they are executable, inspectable, and can be revised directly using task-level feedback.

A closely related line of work studies LLM-based automated heuristic design, where the model acts as a search operator over executable code. EoH combines evolutionary search with LLM-generated heuristic ideas and implementations, ReEvo augments this process with reflective feedback, and MCTS-AHD organizes candidate programs through tree search for more systematic exploration \citep{liu2024eoh,ye2024reevo,zheng2025mcts}. More recently, MOTIF moves beyond single-function search by jointly improving multiple interdependent components through turn-based LLM interaction \citep{kiet2026motif}. Despite this progress, these approaches generally optimize within a centrally accessible program space: candidate implementations can be inspected, compared, or reused directly by the optimization framework.

This assumption becomes restrictive in heterogeneous multi-agent planning. Independently developed agents may specialize in different roles, observe different portions of the state, and expose incompatible interfaces. Prior work on adaptive skill synthesis and decentralized multi-agent learning shows that specialized agents can acquire complementary behaviors, but typically relies on shared representations, accessible implementations, or communication within a common interaction space \citep{li2025adaptive,yang2025agentnet,alzubi2026evoskill}. When signatures and information access differ, a successful program may be unusable by another agent even when its strategic idea remains useful. The central challenge is therefore to transfer reusable strategy without assuming that executable skills are directly compatible.

We therefore study cooperative planning in which each agent has an interpretable programmatic skill with an agent-specific signature, while updates are evaluated through team performance. This setting creates two challenges. First, a local revision must be judged in the context of the current team: it may complement a teammate, duplicate its behavior, or disrupt an emerging division of labor. Second, useful knowledge must cross incompatible executable spaces. Although multi-function methods such as MOTIF can jointly optimize heterogeneous functions \citep{kiet2026motif}, direct code transfer remains unsuitable when implementations should not be exposed. Transfer should therefore capture coordination structure rather than a signature-specific function body.

We introduce RELIC (\textbf{RE}vealed Principles for \textbf{L}earning \textbf{I}nterpretable \textbf{C}omposable Skills) to address this setting. Building on LLM-guided program search \citep{zheng2025mcts,kiet2026motif}, each agent maintains a private tree of candidate programs and improves its executable skill locally. A trusted orchestrator evaluates each revision by inserting it into the current team and measuring its contextual contribution to team utility. Successful decision logic is then abstracted into short textual \emph{revealed principles}, which other agents can re-instantiate under their own signatures. RELIC additionally maintains a public principle tree that accumulates transferable coordination knowledge and favors principles that repeatedly yield positive downstream effects.

The intended information boundary is implementation non-disclosure between agents rather than formal privacy. The orchestrator is trusted to execute candidate programs, and revealed principles deliberately expose strategy-level information; RELIC does not provide differential-privacy or cryptographic guarantees. After training, however, execution remains decentralized: each agent acts only through its own local program and interface, while the public principle memory is used to support learning rather than runtime control.

\paragraph{Contributions.} Building on this problem formulation and the
proposed framework, our main contributions are summarized as follows:
\begin{itemize}
    \item We formulate cooperative programmatic skill learning in two coupled settings: heterogeneous agent signatures with restricted implementation sharing, and streaming task environments that require continual adaptation to newly arriving data. This formulation separates executable interoperability from strategic transfer while making team performance environment-dependent over time.
    \item We propose RELIC, a framework that separates private executable skill optimization from cross-agent knowledge transfer. RELIC combines contextual team-level evaluation with revealed principles and a shared principle memory that retains transferable coordination knowledge according to its observed downstream utility.
    \item We provide a broad empirical evaluation across heterogeneous and shared-role multi-agent settings, covering cooperative routing, scheduling, combinatorial optimization, and distributed coordination problems. We compare RELIC with independent LLM-based heuristic search methods and joint multi-function optimization, testing both incompatible and common skill structures.
    \item We analyze the mechanisms underlying RELIC, including principle-mediated transfer, shared principle memory, and private search, and study the framework's robustness across multiple LLM backbones.
\end{itemize}

\section{Related Work}

\subsection{LLM-Based Automated Heuristic Design}

EoH and ReEvo use evolutionary generation and reflection to improve executable heuristics, while MCTS-AHD organizes program revisions as a tree search \citep{liu2024eoh,ye2024reevo,zheng2025mcts}. MOTIF extends this line from a single heuristic to interacting solver components optimized through turn-based LLM agents \citep{kiet2026motif}. These methods demonstrate the value of evaluating local code changes through system-level performance, but ordinarily assume that the optimizer can access the executable components being revised. B2B complements code-centric AHD with a knowledge-first view: interpretable
knowledge becomes the search object, while executable code instantiates and
tests that knowledge \citep{kiet2026b2b}. RELIC adopts this emphasis on
explicit reusable knowledge, but studies transfer across independently owned,
role-specific programs whose callable interfaces may be incompatible.

\subsection{Programmatic Skill Transfer}

Skill-library methods store and retrieve behaviors learned from demonstrations, trajectories, or generated programs \citep{zhang2023bootstrap,zhu2026offline,wang2026skillx,li2026graph}, while multi-agent learning develops specialized or complementary behaviors across agents \citep{li2025adaptive,yang2025agentnet,alzubi2026evoskill}. RELIC addresses the intersection of these directions when callable interfaces differ and implementations are not exchanged: executable programs remain local, whereas reusable decision logic is transferred as text and re-instantiated by the recipient. This is an implementation-sharing boundary, not a formal privacy guarantee. Unlike prior systems that either optimize isolated heuristics or exchange
executable components, RELIC combines persistent knowledge, heterogeneous
roles, and cross-role transfer without making a contributor's program
available to the recipient. A more detailed discussion of related work is provided in
Appendix~\ref{app:related-work}.

\section{Problem Formulation}

\subsection{Heterogeneous Programmatic Skills}

\paragraph{Agent-Specific Executable Skills.}
Consider a cooperative team of $N$ agents indexed by $i\in[N]$. Agent $i$ receives a local input $x_i\in\mathcal{X}_i$ and produces an action $a_i\in\mathcal{A}_i$ through an executable decision skill
\begin{equation}
    \pi_i:\mathcal{X}_i\rightarrow\mathcal{A}_i.
    \label{eq:skill}
\end{equation}

Each skill must conform to an agent-specific signature $\sigma_i$, which specifies its interface, input and output types, and execution constraints. The input spaces need not coincide: $\mathcal{X}_i\neq\mathcal{X}_j$ may reflect roles, sensors, or information access. Consequently, even if $\pi_i$ contains a useful strategy, its implementation cannot be called through $\sigma_j$.

\paragraph{Cross-Signature Transfer Setting.}
This setting differs from conventional single-function automated heuristic
design, such as EoH~\citep{liu2024eoh} and ReEvo~\citep{ye2024reevo}, which
optimizes one executable heuristic at a time. MOTIF~\citep{kiet2026motif}
extends optimization to multiple interacting functions under a shared
objective, but assumes that their implementations are centrally accessible to
and directly reusable by the optimization process. We instead study heterogeneous skills whose
signatures and information spaces may be incompatible and whose executable
implementations are not disclosed across agents. Thus, transferable knowledge
cannot generally take the form of a function body; it must be expressed at a
strategy level that another agent can instantiate through its own $\sigma_i$.

Let $\boldsymbol{\pi}=(\pi_1,\ldots,\pi_N)$ denote a team of skills. On task
instance $\xi$, the agents execute their programs through their local
interfaces and obtain a cooperative return $R(\xi;\boldsymbol{\pi})$.

Training is coordinated by a trusted orchestrator that can execute a candidate
team and observe its aggregate return. Agents do not directly inspect or call
one another's executable programs, although they may receive textual
principles distilled from successful behavior. After training, execution is
decentralized: agent $i$ acts using only $\pi_i$ and the information admitted
by $\sigma_i$. This is an implementation non-disclosure assumption rather than
a formal privacy guarantee.

\subsection{Streaming Task Environments}

\paragraph{Non-Stationary Batch Stream.}
In addition to skill heterogeneity, we consider a streaming environment in
which task data arrive continuously as batches $b=1,\ldots,B$. Instances in
batch $b$ follow an environment-dependent distribution $\mathcal{D}_b$, which
need not be stationary: $\mathcal{D}_b$ may differ from
$\mathcal{D}_{b'}$ for $b\neq b'$. Hence, a team optimized for earlier data may
become suboptimal when new task structures, operating conditions, or agent
interactions appear. The agents must therefore adapt their skills as each new
batch becomes available rather than learn a single fixed team from an offline
dataset.

For a team $\boldsymbol{\pi}$, its expected utility in environment $b$ is
\begin{equation}
    J_b(\boldsymbol{\pi})
    =
    \mathbb{E}_{\xi\sim\mathcal{D}_b}
    \left[
        R(\xi;\boldsymbol{\pi})
    \right].
    \label{eq:objective}
\end{equation}

\paragraph{Online Team Adaptation.}
The learner observes a finite batch
$\mathcal{B}_b=\{\xi_m^{(b)}\}_{m=1}^{M_b}$,
and uses the empirical utility
\begin{equation}
    \widehat{J}_b(\boldsymbol{\pi})
    =
    \frac{1}{M_b}
    \sum_{m=1}^{M_b}
    R(\xi_m^{(b)};\boldsymbol{\pi}).
    \label{eq:empirical-objective}
\end{equation}

At each arrival, the learner updates the current team using the new batch,
optionally with replay data, and
produces an adapted team $\bar{\boldsymbol{\pi}}_b$. The objective is to maintain
high team utility throughout the stream by improving
$J_b(\bar{\boldsymbol{\pi}}_b)$ under the current environment without
discarding behavior that remains useful from earlier batches.
The supervision is team-level: no agent-specific reward is required, and a
local update is valuable only when it improves the team in the context
of the other skills. Together, heterogeneous skills and streaming
environments define the setting addressed by RELIC.


\section{Methodology}


\subsection{Batch-Level Training}

\paragraph{Batch State and Knowledge Flow.}
Let $\mathfrak{R}$ denote the space of free-form textual principles, each of
which describes an applicable context, a decision rule, and an intended
coordination effect. At batch $b$, RELIC maintains a current best team
whose agent-specific incumbents are $\bar{\pi}_{i,b}$:
\begin{equation}
    \bar{\boldsymbol{\pi}}_b
    =
    (\bar{\pi}_{1,b},\ldots,\bar{\pi}_{N,b}),
\end{equation}
a private program tree $\mathcal{T}_{i,b}$ for each agent, local
revealed-principle archives
$\mathcal{A}^{\mathrm{rev}}_{i,b}\subseteq\mathfrak{R}$, and a shared public
principle tree
$\mathcal{P}_b=(\mathcal{V}^{\mathrm{pub}}_b,\mathcal{E}^{\mathrm{pub}}_b)$.
Private trees store executable programs and remain agent-local; the public tree
stores only textual abstractions and their transfer statistics.

RELIC alternates between \emph{private search} and \emph{principle update}.
During private search, one agent revises its program while its teammates remain
fixed, and the orchestrator measures the effect on team utility. During
principle update, successful behavior is distilled into textual principles,
shared through local archives, and selectively added to the public tree.
Recipients re-instantiate these principles through their own signatures rather
than copy the originating programs.

A \emph{revealed principle} $\rho\in\mathfrak{R}$ is a concise natural-language
decision or coordination rule, not executable code or a fixed structured
record. For example, in skilled-task scheduling, a principle may recommend
deferring a task when a less busy teammate has a clear processing advantage.
Different signatures can realize this idea using local, teammate, or global
observations.


\paragraph{Replay-Based Evaluation.}
To reduce myopic updates under a non-stationary task stream, RELIC evaluates revisions on a mixed batch. Let
$\mathcal{R}_b\subseteq\{1,\ldots,b-1\}$
index replay batches, and let
$\mathcal{M}_t\subseteq\mathcal{B}_t$
be a small replay sample. The evaluation batch is
\begin{equation}
    \widetilde{\mathcal{B}}_b
    =
    \mathcal{B}_b
    \cup
    \bigcup_{t\in\mathcal{R}_b}
    \mathcal{M}_t.
    \label{eq:mixed-batch}
\end{equation}

To distinguish this replay-aware score from the current-batch utility in
Equation~\ref{eq:empirical-objective}, define
\begin{equation}
    \widehat{J}^{\mathrm{mix}}_b(\boldsymbol{\pi})
    =
    \frac{1}{|\widetilde{\mathcal{B}}_b|}
    \sum_{\xi\in\widetilde{\mathcal{B}}_b}
    R(\xi;\boldsymbol{\pi}).
    \label{eq:mixed-objective}
\end{equation}
All candidate revisions within batch $b$ are compared on this same evaluation
set. This shared mixture balances adaptation to the arriving batch with
retention of behaviors that remain useful on previously observed tasks.

\subsection{Private Program Search}

\paragraph{Private Search State.}
Each agent $i$ maintains a private program tree
\[
    \mathcal{T}_{i,b}
    =
    (\mathcal{V}_{i,b},\mathcal{E}_{i,b}).
\]
A node $v\in\mathcal{V}_{i,b}$ stores an executable program $\pi(v)$, visit count $N(v)$, value $Q(v)$, and operator-specific statistics $N_{\mathrm{op}}(v,o)$ and $Q_{\mathrm{op}}(v,o)$. The root contains the seed skill, and each child corresponds to an LLM-generated revision of its parent.
Each edge $(u,v)\in\mathcal{E}_{i,b}$ records that child $v$ was generated by
revising parent $u$.

Starting from the root, RELIC descends the tree by selecting a child $v$ of
the current node $u$ according to
\begin{equation}
    U_{\mathrm{node}}(v;u)
    =
    Q(v)
    +
    c_{\mathrm{node}}
    \sqrt{
        \frac{\ln N(u)}{N(v)}
    }.
    \label{eq:node-uct}
\end{equation}
Here $c_{\mathrm{node}}>0$ is the node exploration coefficient. Unvisited
children are selected before Equation~\ref{eq:node-uct}, so its denominator is
positive whenever the score is evaluated.

\begin{figure*}[t]
    \centering
    \includegraphics[width=0.8\linewidth]{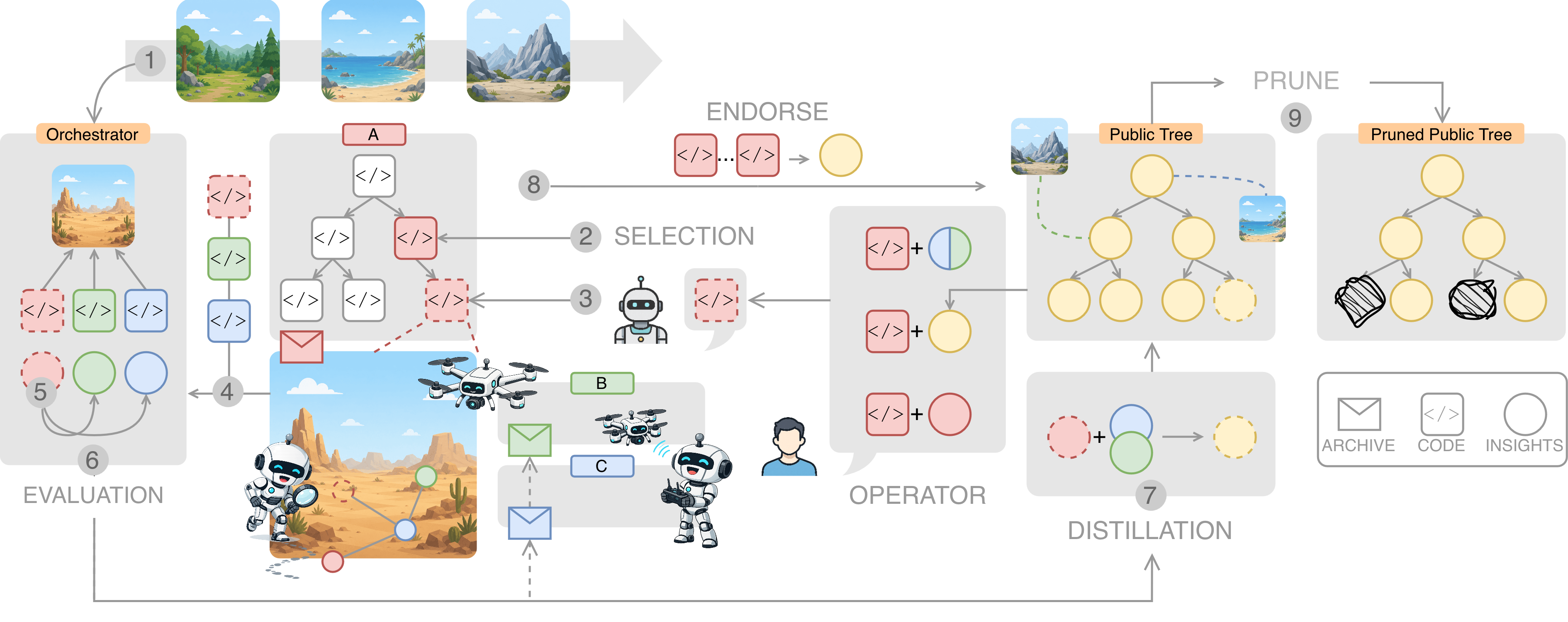}
    \caption{Overview of the RELIC learning cycle. (1) The orchestrator
    receives the next task batch from the environment stream. (2) Private-tree
    search selects an agent program to revise. (3) An LLM applies Lift,
    Bridge, or Reflect to generate a signature-compatible candidate. (4) The
    candidate replaces that agent's incumbent program and is executed with the
    current teammates. (5) The orchestrator derives role-aware credit and
    textual insights from the resulting team behavior. (6) Contextual team
    evaluation updates the private node and operator statistics and accepts an
    improving candidate. (7) Successful behaviors and cross-agent insights are
    distilled into transferable principles in the public tree. (8) A public
    principle is endorsed when its downstream use improves team utility. (9)
    Low-utility leaves are pruned to keep the shared memory compact. Envelopes,
    code boxes, and circles denote local archives, executable programs, and
    textual principles, respectively.}
    \label{fig:relic-overview}
    
\end{figure*}

\paragraph{Operator Selection and Candidate Generation.}
At the expansion node, RELIC considers the operator set
$\mathcal{O}_{i,b}\subseteq
\{\mathrm{Lift},\mathrm{Bridge},\mathrm{Reflect}\}$, where Lift requires a
nonempty local archive, Bridge requires an eligible public principle, and
Reflect is always available. It selects $o\in\mathcal{O}_{i,b}$ using
\begin{equation}
    U_{\mathrm{op}}(o;v)
    =
    Q_{\mathrm{op}}(v,o)
    +
    c_{\mathrm{op}}
    \sqrt{
        \frac{
            \ln
            \sum_{o'\in\mathcal{O}_{i,b}}
            N_{\mathrm{op}}(v,o')
        }{
            N_{\mathrm{op}}(v,o)
        }
    }.
    \label{eq:operator-uct}
\end{equation}
Here $c_{\mathrm{op}}>0$ is the operator exploration coefficient. Each
available operator is tried once before Equation~\ref{eq:operator-uct} is
applied, ensuring $N_{\mathrm{op}}(v,o)>0$.

The two selections answer different questions: which existing implementation should be revised, and which source of transfer or feedback should guide the revision.

Given operator-specific context $C^o_{i,b}$, the LLM produces
\begin{equation}
    \widetilde{\pi}_i
    \sim
    p_\theta
    \left(
        \cdot
        \mid
        \pi(v),
        o,
        C^o_{i,b},
        \sigma_i
    \right).
    \label{eq:generation}
\end{equation}
Here $p_\theta$ is the conditional generation distribution of the LLM with
fixed parameters $\theta$. The parent program $\pi(v)$ and signature
$\sigma_i$ are common to all operators; $C^o_{i,b}$ contains only the
operator-specific transferable knowledge or feedback.

\paragraph{Team Evaluation.}
Only candidates executable through $\sigma_i$ enter team evaluation. Let
$\bar{\boldsymbol{\pi}}_b^{i\leftarrow\widetilde{\pi}_i}$ denote the incumbent
team with agent $i$ replaced by $\widetilde{\pi}_i$. The orchestrator keeps all
other skills fixed and computes
\begin{equation}
    \Delta_b(\widetilde{\pi}_i)=\widehat{J}^{\mathrm{mix}}_b(\bar{\boldsymbol{\pi}}_b^{i\leftarrow\widetilde{\pi}_i})-\widehat{J}^{\mathrm{mix}}_b(\bar{\boldsymbol{\pi}}_b).
    \label{eq:contextual-gain}
\end{equation}

This \emph{contextual marginal gain} measures compatibility with the current
teammates rather than the intrinsic quality of $\widetilde{\pi}_i$ in
isolation. The candidate is added as a new child $v_{\mathrm{new}}$ with
$\pi(v_{\mathrm{new}})=\widetilde{\pi}_i$ and initial value
\[
    Q(v_{\mathrm{new}})=\Delta_b(\widetilde{\pi}_i).
\]
Node and operator statistics are then updated using standard incremental averaging. If the candidate improves the incumbent team score, the orchestrator updates $\bar{\pi}_{i,b}$. This one-agent-at-a-time replacement provides a direct team-level credit signal while ensuring that every intermediate search state remains executable.

The three revision operators differ in the information supplied to the LLM.

\paragraph{Lift.}
Lift adapts a principle revealed by another agent and stored in the recipient's local archive. For
$\rho\in\mathcal{A}^{\mathrm{rev}}_{i,b}$,
\begin{equation}
    C^{\mathrm{Lift}}_{i,b}=\rho.
    \label{eq:lift}
\end{equation}

The LLM preserves the strategic intent of $\rho$ while expressing it only through arguments available under $\sigma_i$. Lift therefore supports direct cross-agent transfer without copying the originating implementation.

\paragraph{Bridge.}
Bridge retrieves an eligible principle node $p^\star_{i,b}$ from the public
tree using the rule defined below and uses its text as revision context:
\begin{equation}
    C^{\mathrm{Bridge}}_{i,b}=\rho(p^\star_{i,b}).
    \label{eq:bridge}
\end{equation}

Unlike a locally revealed principle, a public principle has accumulated evidence from previous downstream use. Bridge therefore connects an agent's private implementation to coordination knowledge that has transferred successfully beyond its original contributor.

\paragraph{Reflect.}
Reflect uses a textual summary
$g_{i,b}(\pi(v))$
produced by the orchestrator:
\begin{equation}
    C^{\mathrm{Reflect}}_{i,b}=g_{i,b}(\pi(v)).
    \label{eq:reflect}
\end{equation}

Here $g_{i,b}$ is an orchestrator-produced textual credit map. Its summary
compares the full team containing $\pi(v)$ with a subset team in which agent
$i$ is removed. It identifies situations in which the agent is decisive,
redundant, or harmful. Reflect thus provides role-aware feedback even though
the optimization signal itself remains team-level.


\subsection{Revealed Principles and Public Tree}

\paragraph{Principle Extraction.}
RELIC converts successful private behaviors into transferable knowledge through reveal, distillation, and cross-distillation.

Let $\mathcal{V}^{\mathrm{new}}_{i,b}\subseteq\mathcal{V}_{i,b}$ be the nodes
created for agent $i$ during batch $b$. The successful new nodes are
\begin{equation}
    \mathcal{S}_{i,b}
    =
    \left\{
        v\in\mathcal{V}^{\mathrm{new}}_{i,b}
        :
        \Delta_b(\pi(v))>0
    \right\}.
    \label{eq:successful-nodes}
\end{equation}

If $\mathcal{S}_{i,b}\neq\varnothing$, RELIC avoids filling public memory with
minor variants by selecting the strongest new node:
\begin{equation}
    v^\star_{i,b}
    =
    \arg\max_{v\in\mathcal{S}_{i,b}}
    \Delta_b(\pi(v)).
    \label{eq:best-new-node}
\end{equation}
When $\mathcal{S}_{i,b}=\varnothing$, agent $i$ contributes no distilled
principle at batch $b$.

The LLM distillation map $D_i$ receives the successful program, its observed
gain, and the current public tree, and returns a textual principle in
$\mathfrak{R}$:
\begin{equation}
    \rho^{\mathrm{dist}}_{i,b}
    =
    D_i
    \left(
        \pi(v^\star_{i,b}),
        \Delta_b(\pi(v^\star_{i,b})),
        \mathcal{P}_b
    \right).
    \label{eq:distillation}
\end{equation}
The new principle is inserted as a public node $p$ with
$\rho(p)=\rho^{\mathrm{dist}}_{i,b}$ and contributor $\kappa(p)=i$. Its public
tree edge records the parent principle used as distillation context, or the
start of a new root branch when no parent principle was used.

\paragraph{Local Reveal and Cross-Distillation.}
The distillation instruction extracts signature-independent decision logic,
its applicable context, and intended team effect, while excluding executable
code and interface-specific variables.

In parallel, the orchestrator applies a reveal map $A_i$ from executable
skills satisfying $\sigma_i$ to $\mathfrak{R}$:
\begin{equation}
    \widehat{\rho}_{i,b}
    =
    A_i(\bar{\pi}_{i,b}).
    \label{eq:reveal}
\end{equation}

The resulting $\widehat{\rho}_{i,b}\in\mathfrak{R}$ is broadcast to each
$\mathcal{A}^{\mathrm{rev}}_{j,b}$ for $j\neq i$ and supplies the knowledge
used by Lift.

Cross-distillation applies a variant of $D_i$ to an incumbent and principles
revealed by its teammates, then inserts the resulting element of
$\mathfrak{R}$ into the public tree by the same rule. This can capture
higher-level coordination relationships, while subsequent Lift and Bridge
revisions empirically test whether they transfer across heterogeneous
interfaces.

\paragraph{Public Principle Retrieval.}
Each public node $p\in\mathcal{V}^{\mathrm{pub}}_b$ stores principle text
$\rho(p)\in\mathfrak{R}$, contributor $\kappa(p)$, retrieval count $N(p)$, and
cumulative positive endorsement $E(p)$. New nodes initialize
$N(p)=E(p)=0$, while edges in $\mathcal{E}^{\mathrm{pub}}_b$ record refinement
relations and determine which nodes are leaves. For agent $i$, the eligible
nodes are
\[
    \mathcal{V}^{\mathrm{pub}}_{i,b}
    =
    \{p\in\mathcal{V}^{\mathrm{pub}}_b:\kappa(p)\neq i\}.
\]
Bridge is available only when this set is nonempty. Eligible nodes with
$N(p)=0$ are retrieved first. After every eligible node has been tried, RELIC
scores them using
\begin{equation}
    U_{\mathrm{pub}}(p)
    =
    \frac{E(p)}{N(p)}
    +
    c_{\mathrm{pub}}
    \sqrt{
        \frac{
            \ln
            \sum_{q\in\mathcal{V}^{\mathrm{pub}}_{i,b}}
            N(q)
        }{
            N(p)
        }
    }.
    \label{eq:public-uct}
\end{equation}
Here $c_{\mathrm{pub}}>0$ is the public-tree exploration coefficient, and the
retrieved node is
$p^\star_{i,b}\in\arg\max_{p\in\mathcal{V}^{\mathrm{pub}}_{i,b}}
U_{\mathrm{pub}}(p)$.

The first term favors principles with strong observed downstream impact, while the second continues to explore less-used principles.

\paragraph{Public-Tree Updates and Scope.}
Every Bridge-derived revision increments the retrieval count of its selected
node. Positive contextual gain additionally increases its endorsement:
\begin{align}
    N(p^\star_{i,b})
    &\leftarrow N(p^\star_{i,b})+1,
    \nonumber\\
    E(p^\star_{i,b})
    &\leftarrow E(p^\star_{i,b})
    +\max\{0,\Delta_b(\widetilde{\pi}_i)\}.
    \label{eq:endorsement}
\end{align}

Repeatedly useful principles are retrieved more often, while low-endorsement
leaves are pruned. The public tree retains textual coordination knowledge and
transfer evidence; private trees retain signature-specific programs.

RELIC uses centralized training evaluation and decentralized execution.
Coordinate updates simplify credit assignment and keep intermediate teams
executable, but may miss improvements requiring simultaneous agent changes.
Textual principles transfer across signatures but offer neither semantic
fidelity nor formal confidentiality.


\section{Experiments}

\paragraph{Baselines and Settings.}
We compare RELIC with EoH-ind. \citep{liu2024eoh}, ReEvo-ind.
\citep{ye2024reevo}, MCTS-AHD-ind. \citep{zheng2025mcts}, HiFo-ind.
\citep{chen2026hifo}, and MOTIF \citep{kiet2026motif}. MOTIF alone natively
optimizes interacting functions with different signatures; the other methods
are adapted as independent role-specific searches (``ind.'') without code or
principle exchange. Each method is run three times under the same data splits,
LLM backbone, proposal budget, and execution limits. Each run uses five
training batches with replay from the two preceding batches and ten proposals
per role per batch; invalid candidates consume budget. For MOTIF, we retain
same-role competitive search but disable its system-aware cross-role round, so
no baseline can access foreign executable code. Benchmark definitions, metrics,
and full settings are provided in Appendices~\ref{app:benchmarks},
\ref{app:evaluation-metrics}, and \ref{app:experiment-details}, respectively.

\subsection{Heterogeneous-Team Performance}

\paragraph{Overall performance.}
Table~\ref{tab:signature-gaps} shows that RELIC is the strongest method overall
across the heterogeneous routing and scheduling tasks, leading most settings
under both the original and alternative signatures. Its advantage persists
when callable interfaces, available observations, and role-specific
responsibilities change, whereas the independently adapted baselines cannot
transfer knowledge across roles. RELIC is particularly effective when one
role's decisions strongly affect its teammates, suggesting that revealed
principles preserve useful coordination logic while allowing each recipient to
re-instantiate it under its own signature. The few exceptions occur mainly on
more weakly coupled objectives, where cross-role transfer provides less
additional benefit.

\newcommand{\relicMainResultsTable}{%
\begin{table*}[t]
\centering
\caption{Main heterogeneous-role results under the original A/B/C signatures.
Results are mean $\pm$ standard deviation. Gap is the relative distance from the best result in each task/size
column; smaller gaps are better.}
\label{tab:main-heterogeneous}
\relicTableSetup

\makebox[\linewidth][l]{\textbf{(a) Multi-Agent Path Planning}}\par\smallskip
\begin{tabular*}{\linewidth}{@{\extracolsep{\fill}}|l|ll|ll|ll|ll|ll|ll|}
\hline
\multirow{3}{*}{Method}
& \multicolumn{4}{l|}{MAPP-PC $\uparrow$}
& \multicolumn{4}{l|}{MAPP-AR $\uparrow$}
& \multicolumn{4}{l|}{MAPP-DS $\uparrow$} \\
\cline{2-13}
& \multicolumn{2}{l|}{50} & \multicolumn{2}{l|}{80}
& \multicolumn{2}{l|}{50} & \multicolumn{2}{l|}{80}
& \multicolumn{2}{l|}{50} & \multicolumn{2}{l|}{80} \\
\cline{2-13}
& Result & Gap $\downarrow$ & Result & Gap $\downarrow$
& Result & Gap $\downarrow$ & Result & Gap $\downarrow$
& Result & Gap $\downarrow$ & Result & Gap $\downarrow$ \\
\hline\hline
EoH-ind.
& 36.60 $\pm$ 1.53 & 6.11\% & 59.20 $\pm$ 2.12 & 4.62\% & 42.12 $\pm$ 0.21 & 1.89\% & 69.97 $\pm$ 0.30 & 3.24\% & 28.79 $\pm$ 3.85 & 26.63\% & 53.29 $\pm$ 7.34 & 19.39\% \\
ReEvo-ind.
& 37.52 $\pm$ 0.91 & 3.75\% & 60.64 $\pm$ 2.46 & 2.30\% & 40.20 $\pm$ 3.32 & 6.36\% & 65.82 $\pm$ 5.26 & 8.98\% & 28.55 $\pm$ 1.55 & 27.24\% & 59.10 $\pm$ 2.01 & 10.60\% \\
MCTS-AHD-ind.
& 36.57 $\pm$ 3.56 & 6.18\% & 58.33 $\pm$ 5.68 & 6.03\% & 39.26 $\pm$ 2.98 & 8.55\% & 65.83 $\pm$ 5.49 & 8.96\% & 29.44 $\pm$ 2.91 & 24.97\% & 57.15 $\pm$ 5.21 & 13.55\% \\
HiFo-ind.
& 35.45 $\pm$ 0.61 & 9.06\% & 56.65 $\pm$ 0.80 & 8.73\% & 41.67 $\pm$ 0.06 & 2.94\% & 68.62 $\pm$ 0.31 & 5.10\% & 27.36 $\pm$ 1.09 & 30.28\% & 55.30 $\pm$ 6.00 & 16.35\% \\
MOTIF
& 36.01 $\pm$ 0.39 & 7.62\% & 57.58 $\pm$ 1.13 & 7.23\% & 42.38 $\pm$ 0.74 & 1.28\% & 69.01 $\pm$ 1.59 & 4.56\% & 24.83 $\pm$ 0.75 & 36.72\% & 47.35 $\pm$ 0.13 & 28.38\% \\
\textbf{RELIC}
& \textbf{38.98 $\pm$ 0.74} & \textbf{0.00\%} & \textbf{62.07 $\pm$ 0.50} & \textbf{0.00\%} & \textbf{42.93 $\pm$ 0.15} & \textbf{0.00\%} & \textbf{72.31 $\pm$ 1.49} & \textbf{0.00\%} & \textbf{39.24 $\pm$ 8.79} & \textbf{0.00\%} & \textbf{66.11 $\pm$ 4.52} & \textbf{0.00\%} \\
\hline
\end{tabular*}

\smallskip
\makebox[\linewidth][l]{\textbf{(b) Multi-Agent Scheduling}}\par\smallskip
\begin{tabular*}{\linewidth}{@{\extracolsep{\fill}}|l|ll|ll|ll|ll|ll|ll|}
\hline
\multirow{3}{*}{Method}
& \multicolumn{4}{l|}{MAS-FS $\downarrow$}
& \multicolumn{4}{l|}{MAS-JD $\downarrow$}
& \multicolumn{4}{l|}{MAS-ST $\uparrow$} \\
\cline{2-13}
& \multicolumn{2}{l|}{50} & \multicolumn{2}{l|}{80}
& \multicolumn{2}{l|}{50} & \multicolumn{2}{l|}{80}
& \multicolumn{2}{l|}{50} & \multicolumn{2}{l|}{80} \\
\cline{2-13}
& Result & Gap $\downarrow$ & Result & Gap $\downarrow$
& Result & Gap $\downarrow$ & Result & Gap $\downarrow$
& Result & Gap $\downarrow$ & Result & Gap $\downarrow$ \\
\hline\hline
EoH-ind.
& 2801.61 $\pm$ 30.06 & 1.45\% & 6497.05 $\pm$ 98.19 & 1.70\% & 240.28 $\pm$ 0.92 & 2.36\% & 865.49 $\pm$ 10.98 & 1.96\% & 64.07 $\pm$ 0.90 & 4.80\% & 77.50 $\pm$ 3.02 & 5.75\% \\
ReEvo-ind.
& 2787.22 $\pm$ 46.73 & 0.93\% & 6453.87 $\pm$ 146.54 & 1.02\% & 240.00 $\pm$ 3.40 & 2.25\% & 865.59 $\pm$ 12.80 & 1.97\% & 58.18 $\pm$ 0.20 & 13.55\% & 65.27 $\pm$ 3.93 & 20.63\% \\
MCTS-AHD-ind.
& \textbf{2761.63 $\pm$ 26.31} & \textbf{0.00\%} & 6426.11 $\pm$ 14.80 & 0.58\% & \textbf{234.73 $\pm$ 2.54} & \textbf{0.00\%} & \textbf{848.88 $\pm$ 9.26} & \textbf{0.00\%} & 61.38 $\pm$ 0.35 & 8.80\% & 71.27 $\pm$ 1.85 & 13.33\% \\
HiFo-ind.
& 2769.03 $\pm$ 4.57 & 0.27\% & 6415.95 $\pm$ 11.49 & 0.43\% & 242.18 $\pm$ 1.72 & 3.17\% & 878.43 $\pm$ 12.69 & 3.48\% & 59.85 $\pm$ 5.42 & 11.07\% & 69.90 $\pm$ 8.08 & 14.99\% \\
MOTIF
& 2766.51 $\pm$ 13.34 & 0.18\% & \textbf{6388.75 $\pm$ 36.53} & \textbf{0.00\%} & 242.18 $\pm$ 2.07 & 3.17\% & 874.69 $\pm$ 14.56 & 3.04\% & 62.33 $\pm$ 3.31 & 7.38\% & 75.00 $\pm$ 3.36 & 8.79\% \\
\textbf{RELIC}
& 2771.49 $\pm$ 1.47 & 0.36\% & 6411.66 $\pm$ 2.42 & 0.36\% & 238.10 $\pm$ 1.02 & 1.44\% & 855.65 $\pm$ 1.81 & 0.80\% & \textbf{67.30 $\pm$ 0.71} & \textbf{0.00\%} & \textbf{82.23 $\pm$ 2.64} & \textbf{0.00\%} \\
\hline
\end{tabular*}
\end{table*}
}

\subsection{Interface and Domain Robustness}

\newcommand{\relicInterfaceTable}{%
\begin{table*}[t]
\centering
\caption{Robustness to alternative A/B/C interfaces. Results are mean $\pm$
standard deviation. Gap is the relative distance from the best result in each task/size
column; smaller gaps are better.}
\label{tab:alternative-interface}
\relicTableSetup

\makebox[\linewidth][l]{\textbf{(a) Multi-Agent Path Planning}}\par\smallskip
\begin{tabular*}{0.96\linewidth}{@{\extracolsep{\fill}}|l|ll|ll|ll|ll|ll|ll|}
\hline
\multirow{3}{*}{Method}
& \multicolumn{4}{l|}{MAPP-PC $\uparrow$}
& \multicolumn{4}{l|}{MAPP-AR $\uparrow$}
& \multicolumn{4}{l|}{MAPP-DS $\uparrow$} \\
\cline{2-13}
& \multicolumn{2}{l|}{50} & \multicolumn{2}{l|}{80}
& \multicolumn{2}{l|}{50} & \multicolumn{2}{l|}{80}
& \multicolumn{2}{l|}{50} & \multicolumn{2}{l|}{80} \\
\cline{2-13}
& Result & Gap $\downarrow$ & Result & Gap $\downarrow$
& Result & Gap $\downarrow$ & Result & Gap $\downarrow$
& Result & Gap $\downarrow$ & Result & Gap $\downarrow$ \\
\hline\hline
EoH-ind.
& 35.25 $\pm$ 0.95 & 4.63\% & 55.61 $\pm$ 1.16 & 7.42\% & 40.23 $\pm$ 1.10 & 3.50\% & 66.29 $\pm$ 2.30 & 4.93\% & 20.67 $\pm$ 3.84 & 14.80\% & 39.55 $\pm$ 5.66 & 16.15\% \\
ReEvo-ind.
& 36.57 $\pm$ 0.38 & 1.06\% & 59.04 $\pm$ 0.59 & 1.71\% & 40.32 $\pm$ 3.10 & 3.29\% & 67.66 $\pm$ 3.62 & 2.97\% & \textbf{24.26 $\pm$ 1.29} & \textbf{0.00\%} & \textbf{47.17 $\pm$ 6.80} & \textbf{0.00\%} \\
MCTS-AHD-ind.
& 34.86 $\pm$ 1.37 & 5.68\% & 55.92 $\pm$ 2.51 & 6.91\% & 38.38 $\pm$ 2.49 & 7.94\% & 64.21 $\pm$ 4.26 & 7.92\% & 20.66 $\pm$ 5.76 & 14.84\% & 39.95 $\pm$ 9.28 & 15.31\% \\
HiFo-ind.
& 34.70 $\pm$ 1.14 & 6.11\% & 55.56 $\pm$ 1.37 & 7.51\% & 40.12 $\pm$ 1.71 & 3.77\% & 67.37 $\pm$ 1.26 & 3.38\% & 20.69 $\pm$ 0.22 & 14.72\% & 37.69 $\pm$ 1.83 & 20.10\% \\
MOTIF
& 35.23 $\pm$ 2.13 & 4.68\% & 57.65 $\pm$ 3.01 & 4.03\% & 40.26 $\pm$ 1.60 & 3.43\% & 67.07 $\pm$ 2.34 & 3.81\% & 21.83 $\pm$ 3.74 & 10.02\% & 41.19 $\pm$ 6.93 & 12.68\% \\
\textbf{RELIC}
& \textbf{36.96 $\pm$ 1.04} & \textbf{0.00\%} & \textbf{60.07 $\pm$ 1.18} & \textbf{0.00\%} & \textbf{41.69 $\pm$ 0.67} & \textbf{0.00\%} & \textbf{69.73 $\pm$ 1.26} & \textbf{0.00\%} & 23.78 $\pm$ 3.08 & 1.98\% & 39.60 $\pm$ 4.56 & 16.05\% \\
\hline
\end{tabular*}

\smallskip
\makebox[\linewidth][l]{\textbf{(b) Multi-Agent Scheduling}}\par\smallskip
\begin{tabular*}{0.96\linewidth}{@{\extracolsep{\fill}}|l|ll|ll|ll|ll|ll|ll|}
\hline
\multirow{3}{*}{Method}
& \multicolumn{4}{l|}{MAS-FS $\downarrow$}
& \multicolumn{4}{l|}{MAS-JD $\downarrow$}
& \multicolumn{4}{l|}{MAS-ST $\uparrow$} \\
\cline{2-13}
& \multicolumn{2}{l|}{50} & \multicolumn{2}{l|}{80}
& \multicolumn{2}{l|}{50} & \multicolumn{2}{l|}{80}
& \multicolumn{2}{l|}{50} & \multicolumn{2}{l|}{80} \\
\cline{2-13}
& Result & Gap $\downarrow$ & Result & Gap $\downarrow$
& Result & Gap $\downarrow$ & Result & Gap $\downarrow$
& Result & Gap $\downarrow$ & Result & Gap $\downarrow$ \\
\hline\hline
EoH-ind.
& 2757.07 $\pm$ 0.16 & 0.56\% & 6384.16 $\pm$ 1.27 & 0.66\% & 236.14 $\pm$ 0.16 & 0.31\% & 853.54 $\pm$ 0.52 & 1.02\% & 59.13 $\pm$ 2.98 & 2.05\% & 69.70 $\pm$ 3.21 & 3.17\% \\
ReEvo-ind.
& 2755.87 $\pm$ 4.35 & 0.52\% & 6385.55 $\pm$ 9.92 & 0.69\% & 236.30 $\pm$ 0.14 & 0.37\% & 856.42 $\pm$ 4.53 & 1.36\% & 54.53 $\pm$ 2.19 & 9.67\% & 64.77 $\pm$ 1.55 & 10.02\% \\
MCTS-AHD-ind.
& 2756.31 $\pm$ 1.17 & 0.53\% & 6382.48 $\pm$ 2.49 & 0.64\% & 236.83 $\pm$ 0.24 & 0.60\% & 855.32 $\pm$ 1.10 & 1.23\% & 56.92 $\pm$ 3.36 & 5.71\% & 66.80 $\pm$ 4.79 & 7.20\% \\
HiFo-ind.
& 2756.94 $\pm$ 0.29 & 0.56\% & 6383.89 $\pm$ 1.13 & 0.66\% & 236.56 $\pm$ 0.60 & 0.48\% & 855.07 $\pm$ 1.70 & 1.20\% & 57.18 $\pm$ 3.85 & 5.28\% & 66.47 $\pm$ 4.23 & 7.65\% \\
MOTIF
& 2754.86 $\pm$ 5.54 & 0.48\% & 6380.89 $\pm$ 5.47 & 0.61\% & \textbf{235.42 $\pm$ 2.30} & \textbf{0.00\%} & \textbf{844.90 $\pm$ 6.08} & \textbf{0.00\%} & 57.72 $\pm$ 3.32 & 4.39\% & 67.23 $\pm$ 4.98 & 6.60\% \\
\textbf{RELIC}
& \textbf{2741.66 $\pm$ 15.57} & \textbf{0.00\%} & \textbf{6342.10 $\pm$ 65.40} & \textbf{0.00\%} & 236.47 $\pm$ 0.44 & 0.45\% & 855.55 $\pm$ 1.79 & 1.26\% & \textbf{60.37 $\pm$ 0.43} & \textbf{0.00\%} & \textbf{71.98 $\pm$ 1.76} & \textbf{0.00\%} \\
\hline
\end{tabular*}
\end{table*}
}

\begin{table*}[t]
\centering
\caption{Gap-only comparison under two A/B/C function signatures.
Arrows indicate the task-objective direction; every reported gap is
nonnegative and lower is better. Objective means and standard deviations are
reported in the appendix.}
\label{tab:signature-gaps}
\relicTableSetup
\setlength{\tabcolsep}{1.1pt}%
\resizebox{\linewidth}{!}{%
\begin{tabular}{|>{\raggedright\arraybackslash}p{0.14\linewidth}|*{12}{>{\raggedright\arraybackslash}p{0.062\linewidth}|}}
\hline
\multirow{3}{*}{Variant}
& \multicolumn{6}{c|}{Multi-Agent Path Planning}
& \multicolumn{6}{c|}{Multi-Agent Scheduling} \\
\cline{2-13}
& \multicolumn{2}{c|}{MAPP-PC $\uparrow$}
& \multicolumn{2}{c|}{MAPP-AR $\uparrow$}
& \multicolumn{2}{c|}{MAPP-DS $\uparrow$}
& \multicolumn{2}{c|}{MAS-FS $\downarrow$}
& \multicolumn{2}{c|}{MAS-JD $\downarrow$}
& \multicolumn{2}{c|}{MAS-ST $\uparrow$} \\
\cline{2-13}
& \multicolumn{1}{c|}{50} & \multicolumn{1}{c|}{80}
& \multicolumn{1}{c|}{50} & \multicolumn{1}{c|}{80}
& \multicolumn{1}{c|}{50} & \multicolumn{1}{c|}{80}
& \multicolumn{1}{c|}{50} & \multicolumn{1}{c|}{80}
& \multicolumn{1}{c|}{50} & \multicolumn{1}{c|}{80}
& \multicolumn{1}{c|}{50} & \multicolumn{1}{c|}{80} \\
\hline\hline
\multicolumn{13}{|c|}{%
\textbf{A:} Myopic-greedy (local state only)
\quad \textbf{B:} Teammate-aware (coordinate via teammate state)
\quad \textbf{C:} Global planner (team-wide state)} \\
\hline
EoH-ind.
& 6.11\% & 4.62\% & 1.89\% & 3.24\% & 26.63\% & 19.39\%
& 1.45\% & 1.70\% & 2.36\% & 1.96\% & 4.80\% & 5.75\% \\
ReEvo-ind.
& 3.75\% & 2.30\% & 6.36\% & 8.98\% & 27.24\% & 10.60\%
& 0.93\% & 1.02\% & 2.25\% & 1.97\% & 13.55\% & 20.63\% \\
MCTS-AHD-ind.
& 6.18\% & 6.03\% & 8.55\% & 8.96\% & 24.97\% & 13.55\%
& \textbf{0.00\%} & 0.58\% & \textbf{0.00\%} & \textbf{0.00\%} & 8.80\% & 13.33\% \\
HiFo-ind.
& 9.06\% & 8.73\% & 2.94\% & 5.10\% & 30.28\% & 16.35\%
& 0.27\% & 0.43\% & 3.17\% & 3.48\% & 11.07\% & 14.99\% \\
MOTIF
& 7.62\% & 7.23\% & 1.28\% & 4.56\% & 36.72\% & 28.38\%
& 0.18\% & \textbf{0.00\%} & 3.17\% & 3.04\% & 7.38\% & 8.79\% \\
\textbf{RELIC}
& \textbf{0.00\%} & \textbf{0.00\%} & \textbf{0.00\%} & \textbf{0.00\%}
& \textbf{0.00\%} & \textbf{0.00\%} & 0.36\% & 0.36\% & 1.44\% & 0.80\%
& \textbf{0.00\%} & \textbf{0.00\%} \\
\hline
\multicolumn{13}{|c|}{%
\textbf{A:} Deadline sentinel (prioritize low slack)
\quad \textbf{B:} Value harvester (high value, low contention)
\quad \textbf{C:} System regulator (coverage/load/utilization)} \\
\hline
EoH-ind.
& 4.63\% & 7.42\% & 3.50\% & 4.93\% & 14.80\% & 16.15\%
& 0.56\% & 0.66\% & 0.31\% & 1.02\% & 2.05\% & 3.17\% \\
ReEvo-ind.
& 1.06\% & 1.71\% & 3.29\% & 2.97\% & \textbf{0.00\%} & \textbf{0.00\%}
& 0.52\% & 0.69\% & 0.37\% & 1.36\% & 9.67\% & 10.02\% \\
MCTS-AHD-ind.
& 5.68\% & 6.91\% & 7.94\% & 7.92\% & 14.84\% & 15.31\%
& 0.53\% & 0.64\% & 0.60\% & 1.23\% & 5.71\% & 7.20\% \\
HiFo-ind.
& 6.11\% & 7.51\% & 3.77\% & 3.38\% & 14.72\% & 20.10\%
& 0.56\% & 0.66\% & 0.48\% & 1.20\% & 5.28\% & 7.65\% \\
MOTIF
& 4.68\% & 4.03\% & 3.43\% & 3.81\% & 10.02\% & 12.68\%
& 0.48\% & 0.61\% & \textbf{0.00\%} & \textbf{0.00\%} & 4.39\% & 6.60\% \\
\textbf{RELIC}
& \textbf{0.00\%} & \textbf{0.00\%} & \textbf{0.00\%} & \textbf{0.00\%}
& 1.98\% & 16.05\% & \textbf{0.00\%} & \textbf{0.00\%} & 0.45\% & 1.26\%
& \textbf{0.00\%} & \textbf{0.00\%} \\
\hline
\end{tabular}%
}
\end{table*}

\paragraph{Robustness across LLM backbones.}
Table~\ref{tab:distributed-backbone} shows that RELIC remains competitive with
all three LLM backbones. Qwen-3.6-Flash obtains the best mean on DGC, DiMES,
and SensorDCSP, whereas GPT-4o-mini is best on SHDS. Because the leading
backbone changes across domains, the results indicate that RELIC's gains do not
depend on a single generator, although their magnitude remains
model- and problem-dependent. This stability is important in distributed tasks,
where local choices interact through shared constraints.

\newcommand{\relicBackboneTable}{%
\begin{table*}[t]
\centering
\caption{Distributed coordination and robustness across LLM backbones
(120 variables). Results are mean $\pm$ standard deviation. Gap is the relative
distance from the best result in each problem column; smaller gaps are better.}
\label{tab:distributed-backbone}
\label{tab:backbone-robustness}
\relicTableSetup
\resizebox{\linewidth}{!}{%
\begin{tabular}{|l|l|ll|ll|ll|ll|}
\hline
\multirow{2}{*}{Group}
& \multirow{2}{*}{Method}
& \multicolumn{2}{l|}{DGC $\downarrow$}
& \multicolumn{2}{l|}{DiMES $\downarrow$}
& \multicolumn{2}{l|}{SensorDCSP $\downarrow$}
& \multicolumn{2}{l|}{SHDS $\downarrow$} \\
\cline{3-10}
& & Result & Gap $\downarrow$ & Result & Gap $\downarrow$
& Result & Gap $\downarrow$ & Result & Gap $\downarrow$ \\
\hline\hline
\multirow{2}{*}{Human}
& DSA-B (Q1)
& 3336.46 $\pm$ 150.65 & 292.86\%
& 9453.24 $\pm$ 35.79 & 397.52\%
& 422.48 $\pm$ 12.79 & 966.08\%
& 4452.50 $\pm$ 31.35 & 59.28\% \\
& LSGA (Q1)
& 2006.35 $\pm$ 42.64 & 136.24\%
& 7398.02 $\pm$ 8.20 & 289.35\%
& 59.33 $\pm$ 15.13 & 49.72\%
& 4701.27 $\pm$ 132.56 & 68.18\% \\
\hline
\multirow{8}{*}{LLM}
& EoH
& 926.29 $\pm$ 6.53 & 9.07\%
& 1921.26 $\pm$ 20.68 & 1.11\%
& 167.23 $\pm$ 35.02 & 321.99\%
& 3291.39 $\pm$ 463.66 & 17.74\% \\
& ReEvo
& 907.47 $\pm$ 16.03 & 6.85\%
& 1916.31 $\pm$ 0.00 & 0.85\%
& 141.42 $\pm$ 8.00 & 256.86\%
& 3559.08 $\pm$ 0.00 & 27.32\% \\
& MCTS-AHD
& 912.40 $\pm$ 39.18 & 7.43\%
& 1951.56 $\pm$ 23.44 & 2.71\%
& 138.85 $\pm$ 19.51 & 250.37\%
& 3374.23 $\pm$ 320.18 & 20.71\% \\
& HiFo
& 941.37 $\pm$ 16.20 & 10.84\%
& 1922.86 $\pm$ 17.71 & 1.20\%
& 127.99 $\pm$ 20.70 & 222.97\%
& 3454.96 $\pm$ 180.35 & 23.59\% \\
& MOTIF
& 901.05 $\pm$ 20.53 & 6.10\%
& 1902.98 $\pm$ 5.98 & 0.15\%
& 73.26 $\pm$ 42.24 & 84.85\%
& 3266.25 $\pm$ 684.94 & 16.84\% \\
& \textbf{RELIC (GPT-4o-mini)}
& 859.70 $\pm$ 68.16 & 1.23\%
& 1901.68 $\pm$ 15.30 & 0.08\%
& 41.40 $\pm$ 15.08 & 4.47\%
& \textbf{2795.42 $\pm$ 223.44} & \textbf{0.00\%} \\
& \textbf{RELIC (DeepSeek-V4-Flash)}
& 894.62 $\pm$ 38.20 & 5.34\%
& 1907.14 $\pm$ 34.91 & 0.37\%
& 51.45 $\pm$ 38.60 & 29.82\%
& 2872.18 $\pm$ 398.53 & 2.75\% \\
& \textbf{RELIC (Qwen-3.6-Flash)}
& \textbf{849.28 $\pm$ 0.00} & \textbf{0.00\%}
& \textbf{1900.08 $\pm$ 16.60} & \textbf{0.00\%}
& \textbf{39.63 $\pm$ 10.13} & \textbf{0.00\%}
& 2800.30 $\pm$ 400.10 & 0.17\% \\
\hline
\end{tabular}%
}
\end{table*}
}

\relicBackboneTable

\newcommand{\relicGeneralizationTable}{%
\begin{center}
\captionof{table}{Generalization beyond the main heterogeneous-role interfaces.
Results are mean $\pm$ standard deviation. Gap is the relative distance from
the best result in each task/size column; smaller gaps are better.}
\label{tab:generalization}
\relicOneColumnTableSetup

\fitTable{%
\begin{tabular}{|l|ll|ll|ll|ll|}
\hline
\multirow{3}{*}{Method}
& \multicolumn{4}{l|}{HMTSP $\downarrow$}
& \multicolumn{4}{l|}{FSTSP $\downarrow$} \\
\cline{2-9}
& \multicolumn{2}{l|}{100} & \multicolumn{2}{l|}{200}
& \multicolumn{2}{l|}{100} & \multicolumn{2}{l|}{200} \\
\cline{2-9}
& Result & Gap $\downarrow$ & Result & Gap $\downarrow$
& Result & Gap $\downarrow$ & Result & Gap $\downarrow$ \\
\hline\hline
EoH
& 75.28 $\pm$ 0.32 & 0.63\% & 135.08 $\pm$ 1.48 & 1.34\% & 15.40 $\pm$ 0.73 & 80.33\% & 26.21 $\pm$ 3.05 & 117.33\% \\
ReEvo
& 74.82 $\pm$ 0.16 & 0.01\% & \textbf{133.30 $\pm$ 0.11} & \textbf{0.00\%} & 15.58 $\pm$ 4.70 & 82.44\% & 27.93 $\pm$ 10.39 & 131.59\% \\
MCTS-AHD
& 75.13 $\pm$ 0.38 & 0.43\% & 134.26 $\pm$ 1.54 & 0.72\% & 11.68 $\pm$ 3.78 & 36.77\% & 19.29 $\pm$ 8.35 & 59.95\% \\
HiFo
& \textbf{74.81 $\pm$ 0.18} & \textbf{0.00\%} & 133.35 $\pm$ 0.04 & 0.04\% & 15.85 $\pm$ 0.06 & 85.60\% & 28.25 $\pm$ 0.33 & 134.25\% \\
MOTIF
& 75.13 $\pm$ 0.38 & 0.43\% & 134.26 $\pm$ 1.54 & 0.72\% & 9.65 $\pm$ 0.79 & 13.00\% & 13.27 $\pm$ 1.16 & 10.03\% \\
\textbf{RELIC}
& 74.90 $\pm$ 0.03 & 0.12\% & 133.43 $\pm$ 0.11 & 0.10\% & \textbf{8.54 $\pm$ 0.12} & \textbf{0.00\%} & \textbf{12.06 $\pm$ 0.12} & \textbf{0.00\%} \\
\hline
\end{tabular}%
}

\smallskip
\fitTable{%
\begin{tabular}{|l|ll|ll|ll|ll|}
\hline
\multirow{3}{*}{Method}
& \multicolumn{4}{l|}{FJAGV $\downarrow$}
& \multicolumn{4}{l|}{MRCPS $\downarrow$} \\
\cline{2-9}
& \multicolumn{2}{l|}{100} & \multicolumn{2}{l|}{200}
& \multicolumn{2}{l|}{100} & \multicolumn{2}{l|}{200} \\
\cline{2-9}
& Result & Gap $\downarrow$ & Result & Gap $\downarrow$
& Result & Gap $\downarrow$ & Result & Gap $\downarrow$ \\
\hline\hline
EoH
& 20707 $\pm$ 723 & 14.54\% & 90134 $\pm$ 3084 & 15.39\% & 933.95 $\pm$ 14.56 & 0.71\% & 1983.25 $\pm$ 44.28 & 1.91\% \\
ReEvo
& 21557 $\pm$ 117 & 19.24\% & 93435 $\pm$ 159 & 19.62\% & 928.95 $\pm$ 14.74 & 0.17\% & 1968.96 $\pm$ 46.76 & 1.17\% \\
MCTS-AHD
& 20437 $\pm$ 1289 & 13.05\% & 88912 $\pm$ 5089 & 13.83\% & \textbf{927.40 $\pm$ 1.51} & \textbf{0.00\%} & \textbf{1946.11 $\pm$ 4.17} & \textbf{0.00\%} \\
HiFo
& 21551 $\pm$ 77 & 19.21\% & 92775 $\pm$ 503 & 18.77\% & 934.08 $\pm$ 23.03 & 0.72\% & 1974.40 $\pm$ 54.48 & 1.45\% \\
MOTIF
& 20272 $\pm$ 2252 & 12.14\% & 87476 $\pm$ 8857 & 11.99\% & 933.36 $\pm$ 13.90 & 0.64\% & 1972.68 $\pm$ 51.35 & 1.37\% \\
\textbf{RELIC}
& \textbf{18078 $\pm$ 241} & \textbf{0.00\%} & \textbf{78110 $\pm$ 767} & \textbf{0.00\%} & 929.28 $\pm$ 5.46 & 0.20\% & 1969.82 $\pm$ 19.63 & 1.22\% \\
\hline
\end{tabular}%
}

\end{center}
}

\paragraph{Generalization to combinatorial optimization.}
Table~\ref{tab:generalization} evaluates four additional problems whose A/B/C
roles perform different parts of a constructive solver. RELIC obtains the best
mean in four of eight columns---both FSTSP and both FJAGV sizes---and remains
close to the best result on HMTSP and MRCPS. The gains on FSTSP and FJAGV show
that revealed-principle transfer extends beyond routing and scheduling, while
the close results elsewhere show that role-wise search remains competitive when
cross-role coordination is less decisive. This pattern suggests that transfer
is most valuable when one role's constructive choice directly shapes the
feasible decisions available to the others.

\relicGeneralizationTable

\subsection{Mechanism Analysis}

We next isolate the contribution of the public principle tree, the three revision operators, and private-tree selection.

\newcommand{\relicAblationTable}{%
\begin{center}
\captionof{table}{Component ablation. Results are mean $\pm$ standard deviation. Gap
is the relative distance from the best variant in each task/size column;
smaller gaps are better. Without the public tree, principles are stored in a
flat pool and selected uniformly at random.}
\label{tab:component-ablation}
\relicOneColumnTableSetup

\makebox[\linewidth][l]{\textbf{(a) Multi-Agent Path Planning}}\par\smallskip
\fitTable{%
\begin{tabular}{|l|ll|ll|ll|ll|}
\hline
\multirow{3}{*}{Variant}
& \multicolumn{4}{l|}{MAPP-PC $\uparrow$}
& \multicolumn{4}{l|}{MAPP-AR $\uparrow$} \\
\cline{2-9}
& \multicolumn{2}{l|}{50} & \multicolumn{2}{l|}{80}
& \multicolumn{2}{l|}{50} & \multicolumn{2}{l|}{80} \\
\cline{2-9}
& Result & Gap $\downarrow$ & Result & Gap $\downarrow$
& Result & Gap $\downarrow$ & Result & Gap $\downarrow$ \\
\hline\hline
\textbf{RELIC Full}
& \textbf{38.98 $\pm$ 0.74} & \textbf{0.00\%} & \textbf{62.07 $\pm$ 0.50} & \textbf{0.00\%} & \textbf{42.93 $\pm$ 0.15} & \textbf{0.00\%} & \textbf{72.31 $\pm$ 1.49} & \textbf{0.00\%} \\
w/o Public Tree
& 37.37 $\pm$ 0.84 & 4.13\% & 58.72 $\pm$ 1.10 & 5.40\% & 42.74 $\pm$ 1.05 & 0.44\% & 69.04 $\pm$ 2.33 & 4.52\% \\
w/o Bridge
& 37.52 $\pm$ 0.28 & 3.75\% & 59.63 $\pm$ 0.70 & 3.93\% & 40.62 $\pm$ 1.12 & 5.38\% & 68.10 $\pm$ 2.10 & 5.82\% \\
w/o Lift
& 37.45 $\pm$ 0.49 & 3.93\% & 59.38 $\pm$ 1.50 & 4.33\% & 41.77 $\pm$ 0.44 & 2.70\% & 69.67 $\pm$ 3.32 & 3.65\% \\
w/o Reflect
& 35.27 $\pm$ 1.00 & 9.52\% & 57.53 $\pm$ 1.40 & 7.31\% & 40.45 $\pm$ 1.59 & 5.78\% & 66.25 $\pm$ 2.23 & 8.38\% \\
w/o Private UCB
& 37.11 $\pm$ 0.30 & 4.80\% & 60.94 $\pm$ 0.70 & 1.82\% & 41.63 $\pm$ 0.13 & 3.03\% & 70.18 $\pm$ 0.58 & 2.95\% \\
\hline
\end{tabular}%
}

\smallskip
\makebox[\linewidth][l]{\textbf{(b) Multi-Agent Scheduling}}\par\smallskip
\fitTable{%
\begin{tabular}{|l|ll|ll|ll|ll|}
\hline
\multirow{3}{*}{Variant}
& \multicolumn{4}{l|}{MAS-JD $\downarrow$}
& \multicolumn{4}{l|}{MAS-ST $\uparrow$} \\
\cline{2-9}
& \multicolumn{2}{l|}{50} & \multicolumn{2}{l|}{80}
& \multicolumn{2}{l|}{50} & \multicolumn{2}{l|}{80} \\
\cline{2-9}
& Result & Gap $\downarrow$ & Result & Gap $\downarrow$
& Result & Gap $\downarrow$ & Result & Gap $\downarrow$ \\
\hline\hline
\textbf{RELIC Full}
& \textbf{238.10 $\pm$ 1.02} & \textbf{0.00\%} & \textbf{855.65 $\pm$ 1.81} & \textbf{0.00\%} & \textbf{67.30 $\pm$ 0.71} & \textbf{0.00\%} & \textbf{82.23 $\pm$ 2.64} & \textbf{0.00\%} \\
w/o Public Tree
& 241.86 $\pm$ 1.50 & 1.58\% & 872.39 $\pm$ 1.00 & 1.96\% & 59.40 $\pm$ 3.40 & 11.74\% & 68.93 $\pm$ 6.25 & 16.17\% \\
w/o Bridge
& 242.98 $\pm$ 1.00 & 2.05\% & 880.46 $\pm$ 6.00 & 2.90\% & 57.82 $\pm$ 3.40 & 14.09\% & 67.15 $\pm$ 8.31 & 18.34\% \\
w/o Lift
& 241.90 $\pm$ 0.90 & 1.60\% & 872.15 $\pm$ 1.00 & 1.93\% & 58.52 $\pm$ 4.10 & 13.05\% & 70.37 $\pm$ 5.27 & 14.42\% \\
w/o Reflect
& 238.99 $\pm$ 1.60 & 0.37\% & 862.02 $\pm$ 7.00 & 0.74\% & 63.33 $\pm$ 1.00 & 5.90\% & 72.80 $\pm$ 0.93 & 11.47\% \\
w/o Private UCB
& 241.32 $\pm$ 2.60 & 1.35\% & 871.26 $\pm$ 1.00 & 1.82\% & 61.90 $\pm$ 4.00 & 8.02\% & 74.20 $\pm$ 7.72 & 9.77\% \\
\hline
\end{tabular}%
}
\end{center}
}

\relicAblationTable

\paragraph{Component ablation.}
Table~\ref{tab:component-ablation} shows that RELIC's components play
complementary roles. Reflect supplies role-aware feedback when local choices
have downstream effects, and Bridge converts validated public principles into
cross-role guidance. The public tree preserves transferable knowledge, whereas
private UCB maintains local specialization; strong performance relies on
combination. The consistent degradation across removals further suggests that
RELIC's gains emerge from coordinated interaction among these mechanisms rather
than from any single dominant component.

\begingroup
\par\smallskip
\captionof{table}{Operator statistics. Selection is the call frequency, Success and Downside are the positive- and negative-gain rates, Mean Gain averages successful calls, and Gain Share is the fraction of total positive gain.}
\label{tab:operator-analysis}
\centering
\relicOneColumnTableSetup
\resizebox{\linewidth}{!}{%
\begin{tabular}{|l|l|l|l|l|l|}
\hline
Operator
& Selection
& Success
& Mean Gain
& Gain Share
& Downside \\
\hline\hline
Lift    & 49.70\% & 8.90\% & 5.20\% & 48.10\% & 69.90\% \\
Bridge  & 27.10\% & 9.50\% & 4.00\% & 29.80\% & 65.80\% \\
Reflect & 23.20\% & 9.00\% & 3.60\% & 22.10\% & 73.00\% \\
\hline
\end{tabular}%
}
\par
\endgroup

\newpage
\begin{figure}[t]
\centering
\includegraphics[width=1.0\columnwidth]{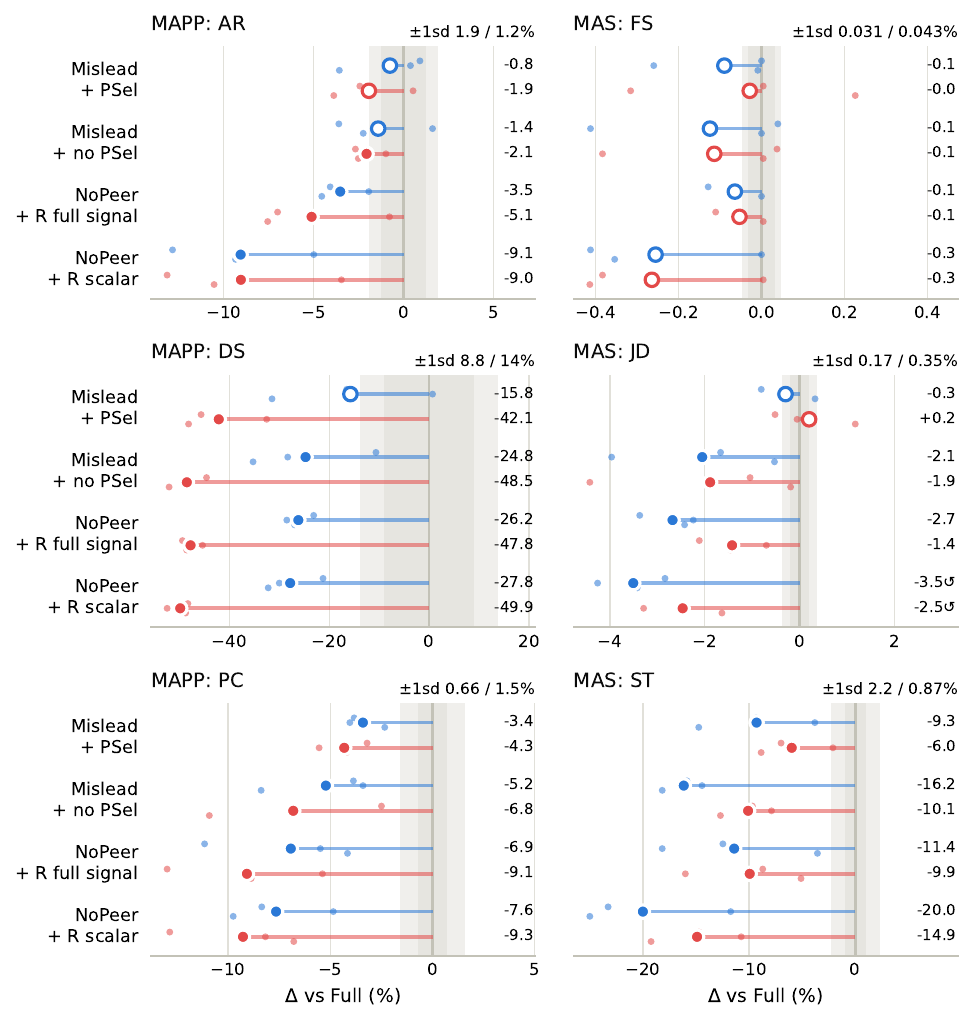}
\caption{Communication interventions on the heterogeneous routing and scheduling benchmarks at test sizes 50 (left) and 80 (right). Points report the relative change from full RELIC and horizontal intervals show $\pm 1$ standard deviation. \emph{Mislead} injects corrupted principles, whereas \emph{NoPeer} removes peer principles; principle selection and full Reflect are evaluated as corresponding mitigation mechanisms.}
\label{fig:communication-interventions}
\end{figure}

\subsection{Communication and Operator Dynamics}

\paragraph{Communication interventions.}
Figure~\ref{fig:communication-interventions} reports the relative change from
full RELIC when the shared channel is corrupted (\emph{Mislead}) or removed
(\emph{NoPeer}). Misleading guidance can propagate across the team because a
shared principle may be re-instantiated by several agents, aligning their local
programs around the same faulty coordination pattern. This effect is strongest
on tightly coupled tasks, where one poor decision changes the opportunities
available to other roles. Principle selection consistently reduces the damage:
misleading principles fail to accumulate downstream endorsement and are
therefore down-selected (Eq.~\ref{eq:endorsement}). It consequently acts as a
team-level filter that limits, although does not eliminate, error propagation.
Under \emph{NoPeer}, structured Reflect feedback is also less harmful than a
single scalar signal, showing that role-aware diagnosis preserves useful credit
information even when direct peer knowledge is unavailable.

\paragraph{Operator behavior.}
Table~\ref{tab:operator-analysis} separates selection frequency from positive
contextual gain. Lift accounts for nearly half of all selections and $48.1\%$
of positive contribution, while Bridge has the highest success rate. Reflect
is selected less often and has the largest downside rate, consistent with a
diagnostic operator that can expose failures without immediately improving the
team. The similar success rates but different contribution shares indicate
distinct search roles rather than interchangeable prompts. Importantly, RELIC
does not impose a fixed hierarchy among the three operators: each participates
as an equal-status mechanism under the same search controller and contributes
through a different pathway. Lift supplies broad principle-mediated transfer,
Bridge provides reliable cross-role re-instantiation, and Reflect diagnoses
coordination failures that guide subsequent revisions. Their value therefore
comes from complementary system-level contributions rather than from one
operator consistently dominating the others.


\section{Conclusion}

We introduced RELIC, a framework that improves private executable skills while transferring coordination knowledge as revealed principles. Across routing, scheduling, combinatorial optimization, and distributed coordination, RELIC is strongest when agents create substantial cross-role externalities and remains competitive on more weakly coupled tasks. The ablations show that private search, public memory, and the equal-status Lift, Bridge, and Reflect operators make complementary contributions, while communication interventions demonstrate that principle selection can limit the propagation of misleading shared knowledge. RELIC nevertheless assumes a trusted training orchestrator, updates one agent at a time, and provides implementation non-disclosure rather than formal privacy. Future work should study joint updates, stronger principle validation, secure evaluation, and cheaper contextual credit assignment while preserving decentralized execution.

\clearpage


\bibliography{ref2}



\onecolumn


\setlength{\oddsidemargin}{0.25in}
\setlength{\evensidemargin}{0.25in}
\setlength{\textwidth}{6.0in}
\setlength{\topmargin}{0.25in}
\setlength{\textheight}{8.0in}
\setlength{\columnwidth}{\textwidth}
\setlength{\linewidth}{\textwidth}

\hsize=\textwidth
\vsize=\textheight

\makeatletter
\setlength{\@colroom}{\textheight}
\setlength{\@colht}{\textheight}
\makeatother

\appendix
\setcounter{secnumdepth}{2}

\hrule

\begin{center}
    {\LARGE\bfseries Appendix\par}
    \vspace{0.4em}

    {\Large\itshape
    RELIC: Revealed Principles for Learning\\
    Interpretable Composable Skills in Multi-Agent Planning\par}
\end{center}

\hrule

\vspace{1em}

\makeatletter

\renewcommand{\section}{
    \@startsection{section}{1}{\z@}%
    {-2.0ex plus -0.5ex minus -.2ex}%
    {3pt plus 2pt minus 1pt}%
    {\Large\bf\raggedright}
}

\makeatother

\renewcommand{\thesection}{\Alph{section}}
\renewcommand{\thesubsection}{\thesection.\arabic{subsection}}
\renewcommand{\labelenumi}{\arabic{enumi}.}

\makeatletter

\newcommand{\appendixcontents}{%
    {\large\bfseries Contents\par}
    \vspace{0.4em}
    \@starttoc{atoc}
}

\newcommand{\appendixtocentry}[4]{%
    \addtocontents{atoc}{%
        \protect\contentsline
        {#1}
        {\protect\numberline{#2}#3}
        {\thepage}
        {#4}
    }
}

\newcommand{\appsection}[1]{%
    \section{#1}%
    \edef\appcurrenthref{\@currentHref}%
    \appendixtocentry{section}{\thesection}{#1}{\appcurrenthref}
}

\newcommand{\appsubsection}[1]{%
    \subsection{#1}%
    \edef\appcurrenthref{\@currentHref}%
    \appendixtocentry{subsection}{\thesubsection}{#1}{\appcurrenthref}
}

\makeatother

\appendixcontents

\newpage

\appsection{Related Work}
\label{app:related-work}

\appsubsection{Knowledge-Centric Automated Heuristic Design}
\label{app:related-ahd}

\paragraph{Executable heuristics as search objects.}
LLM-based automated heuristic design (AHD) replaces manual rule engineering
with search over executable programs. EoH~\citep{liu2024eoh} couples
evolutionary selection with LLM-generated algorithmic ideas and code, thereby
making the language model a variation operator over a population of
heuristics. ReEvo~\citep{ye2024reevo} augments evolutionary search with
reflective verbal feedback, allowing weaknesses of previously evaluated
heuristics to inform subsequent proposals. MCTS-AHD~\citep{zheng2025mcts}
instead organizes program revisions in a Monte Carlo tree, preserving
alternative improvement trajectories and balancing their exploitation and
exploration. HiFo-Prompt~\citep{chen2026hifo} complements these population-
and tree-based approaches by prompting with both hindsight from evaluated
candidates and foresight about promising future modifications.

These methods establish executable code as an effective, interpretable search
representation: a candidate can be inspected, run directly, and evaluated
without a learned surrogate. Their standard unit of optimization, however, is
one heuristic or one centrally managed search space. When several programs
interact, independently improving each program does not reveal whether a
revision complements or duplicates the behavior of its current teammates.
Moreover, knowledge is normally reused inside a compatible implementation
space; a useful function body cannot be directly transferred when another
agent exposes a different signature or observation set. RELIC retains
executable local search but evaluates every revision in its current team
context and transfers the underlying decision principle rather than the
source program.

\paragraph{Knowledge as an explicit intermediate representation.}
B2B~\citep{kiet2026b2b} moves AHD toward a knowledge-first formulation:
explicit algorithmic knowledge is proposed, instantiated as code, and revised
using execution feedback. This separation is important because natural
language can describe design intent at a level that is less tied to a
particular implementation than source code. Related work on language-model
program generation also demonstrates that code can represent reusable
policies~\citep{liang2023code}, generalized planning procedures
\citep{silver2024generalized}, and planning-domain models
\citep{oswald2024domain}.

RELIC builds on this knowledge--code separation but assigns it a different
role. Its revealed principles are not only explanations used to improve the
program from which they were extracted. They are cross-agent transfer
artifacts: a recipient interprets the same strategy through its own inputs,
outputs, and execution constraints. RELIC further records whether such
re-instantiation improves downstream team utility, so persistent knowledge is
selected by observed transfer value rather than solely by the quality of its
originating program.

\appsubsection{Multi-Component and Multi-Agent Heuristic Optimization}
\label{app:related-multicomponent}

\paragraph{Optimizing interacting solver components.}
The value of a heuristic component is often conditional on the other
components with which it is composed. MOTIF~\citep{kiet2026motif} addresses
this dependence by using turn-based LLM agents to optimize multiple
interacting functions under a shared system objective. It therefore moves
beyond isolated single-function AHD and provides a particularly relevant
comparison for heterogeneous solver roles. RELIC shares MOTIF's system-level
view: a local candidate should be assessed through the performance of the
complete solver rather than through an agent-specific proxy.

The two frameworks address different information settings. Multi-component
optimization ordinarily assumes that the optimization process can inspect the
components being revised and can carry implementation-level information
between its turns. RELIC instead studies independently owned programs: the
trusted evaluator can execute the complete team, but a recipient agent cannot
inspect or call a teammate's function. This restriction rules out direct
cross-role code comparison, recombination, and reuse. RELIC consequently
separates private executable search from a public, text-only knowledge path
and makes cross-signature re-instantiation an explicit part of the learning
problem.

\paragraph{Specialization and decentralized coordination.}
Cooperative multi-agent learning provides several mechanisms for producing
specialized behaviors. Adaptive Skill Synthesis~\citep{li2025adaptive}
constructs skills that support cooperative planning, while
AgentNet~\citep{yang2025agentnet} uses decentralized evolutionary
coordination among LLM-based agents. EvoSkill~\citep{alzubi2026evoskill}
automates skill discovery for multi-agent systems, illustrating how an agent
population can acquire complementary capabilities rather than a single
monolithic policy.

These directions motivate learning at the level of specialized agents, but
specialization alone does not solve interoperability. Agents may still rely
on a common interaction representation, exchange rich behavioral artifacts,
or evolve within a jointly accessible system. RELIC focuses on the narrower
but practically distinct case in which local executable interfaces are
incompatible and implementations are not exchanged across agents. Its
contribution is therefore not a new communication protocol for runtime
action selection; it is a training-time mechanism for moving reusable
coordination logic between private program spaces while retaining
decentralized execution.

\appsubsection{Programmatic Skill Libraries and Cross-Agent Transfer}
\label{app:related-skill-transfer}

\paragraph{Acquiring and storing reusable skills.}
Skill-library methods avoid solving every task from scratch by converting
experience into reusable behaviors. Bootstrap Your Own Skills
\citep{zhang2023bootstrap} uses LLM guidance to acquire skills that can be
composed for new tasks. Offline Discovery of Interpretable Skills
\citep{zhu2026offline} extracts understandable skills from multi-task
trajectories. SkillX~\citep{wang2026skillx} automatically constructs a skill
knowledge base for agents, and Graph of Skills~\citep{li2026graph} represents
dependencies among skills to support structure-aware retrieval at larger
library scale. Collectively, these works show that persistent skill memories
can improve reuse, retrieval, and composition beyond one learning episode.

Most skill libraries assume that a stored artifact can be invoked, adapted,
or grounded through a sufficiently compatible interface. That assumption is
fragile for heterogeneous cooperative teams: two roles may differ not only in
their observations, but also in their action semantics and callable
signatures. A trajectory or function that is executable for one role may be
invalid for another even when it encodes a useful coordination idea. RELIC
therefore stores transferable strategy separately from executable behavior.
Private trees retain signature-specific programs, whereas the public tree
contains natural-language principles and evidence about their successful
reuse.

\begin{table}[t]
\centering
\caption{Positioning of RELIC relative to representative related-work
directions. \textbf{Exec.}: search or synthesis produces executable
artifacts; \textbf{Team}: candidates are optimized through a multi-component
team objective; \textbf{Het.}: heterogeneous callable interfaces or roles are
explicitly supported; \textbf{Xfer}: knowledge transfers across agents or
components; \textbf{No code}: transfer does not require exposing the source
executable implementation to the recipient; \textbf{Memory}: transferable
knowledge persists beyond one local revision; \textbf{Stream}: skills adapt
over arriving task batches. $\checkmark$ denotes explicit support,
$\triangle$ partial or indirect support, and -- an aspect not central to the
method.}
\label{tab:related-positioning}
\relicTableSetup
\resizebox{\linewidth}{!}{%
\begin{tabular}{|>{\raggedright\arraybackslash}p{0.32\linewidth}|
                    c|c|c|c|c|c|c|}
\hline
\textbf{Representative direction}
& \textbf{Exec.}
& \textbf{Team}
& \textbf{Het.}
& \textbf{Xfer}
& \textbf{No code}
& \textbf{Memory}
& \textbf{Stream} \\
\hline\hline
Evolutionary AHD:
EoH~\citep{liu2024eoh}, ReEvo~\citep{ye2024reevo}
& $\checkmark$ & -- & -- & -- & -- & $\triangle$ & -- \\
\hline
Tree- and prompt-guided AHD:
MCTS-AHD~\citep{zheng2025mcts}, HiFo-Prompt~\citep{chen2026hifo}
& $\checkmark$ & -- & -- & -- & -- & $\triangle$ & -- \\
\hline
Knowledge-centric AHD:
B2B~\citep{kiet2026b2b}
& $\checkmark$ & -- & -- & $\triangle$ & -- & $\checkmark$ & -- \\
\hline
Multi-component optimization:
MOTIF~\citep{kiet2026motif}
& $\checkmark$ & $\checkmark$ & $\checkmark$ & $\triangle$ & -- & $\triangle$ & -- \\
\hline
Programmatic skill libraries:
BYOS~\citep{zhang2023bootstrap}, Offline Skills~\citep{zhu2026offline},
SkillX~\citep{wang2026skillx}, Graph of Skills~\citep{li2026graph}
& $\triangle$ & -- & $\triangle$ & $\checkmark$ & $\triangle$ & $\checkmark$ & $\triangle$ \\
\hline
Multi-agent skill learning:
Adaptive Skill Synthesis~\citep{li2025adaptive},
AgentNet~\citep{yang2025agentnet}, EvoSkill~\citep{alzubi2026evoskill}
& $\triangle$ & $\checkmark$ & $\triangle$ & $\checkmark$ & -- & $\triangle$ & $\triangle$ \\
\hline
\textbf{RELIC}
& $\checkmark$ & $\checkmark$ & $\checkmark$ & $\checkmark$
& $\checkmark$ & $\checkmark$ & $\checkmark$ \\
\hline
\end{tabular}%
}
\end{table}

\paragraph{Transfer under an implementation-sharing boundary.}
Text is a deliberately lossy transfer medium. It can omit interface-specific
details and expose the strategic relationship between decisions, such as
deferring work to a better-suited teammate or avoiding demand already covered
by another role. The recipient must then ground that relationship using only
variables admitted by its own signature. This makes transfer possible where
literal code reuse is impossible, but it introduces semantic drift: an
apparently relevant principle may be mistranslated or may cease to apply under
the recipient's observations. RELIC addresses this risk empirically through
contextual team evaluation, downstream endorsement, and pruning rather than
assuming that every textual abstraction transfers faithfully.

This boundary should not be confused with formal privacy. Revealed principles
intentionally disclose strategy-level information, and a trusted
orchestrator executes all candidate programs. RELIC provides implementation
non-disclosure between learning agents, not differential privacy,
cryptographic confidentiality, or protection against inference from shared
principles. Its advantage over direct skill exchange is interoperability and
controlled abstraction, while stronger privacy guarantees remain a separate
research direction.

\appsubsection{Positioning RELIC}
\label{app:positioning}

Table~\ref{tab:related-positioning} positions RELIC along the capabilities
most relevant to our setting. The table distinguishes explicit support
($\checkmark$), partial or indirect support ($\triangle$), and aspects that
are not central to a line of work (--). The comparison is intentionally about
method scope rather than absolute capability: for example, a single-function
AHD method can be embedded in a multi-agent pipeline, but multi-component
credit and cross-role transfer are not part of its native formulation.

RELIC is distinguished by the conjunction of these properties rather than by
any one column in isolation. It performs team-level search over executable
skills, but keeps each implementation local; it enables knowledge transfer,
but requires the recipient to re-instantiate that knowledge under a different
signature; and it retains principles according to measured downstream
utility while continuing to adapt over a task stream. This combination
targets cooperative settings in which strategic reuse is valuable precisely
because direct executable reuse is unavailable.

\newpage
\appsection{Benchmark Problems}
\label{app:benchmarks}

This section formalizes the benchmark problems used to evaluate RELIC. We distinguish three groups: six cooperative routing and scheduling problems with heterogeneous interfaces, four combinatorial optimization problems with specialized constructive roles, and four distributed coordination benchmarks. In every case, RELIC optimizes local executable decision functions, while the simulator evaluates the complete joint execution under a global objective.

\appsubsection{Cooperative Multi-Agent Routing}
\label{app:path-planning}

We refer to this cooperative routing family in the paper as
\emph{Multi-Agent Path Planning} (MAPP). It is implemented as a synchronous,
stepwise rollout on a
complete metric graph
\[
    G=(V,E), \qquad V=\{0,1,\ldots,N\},
\]
where node $0$ is a common depot and $\ell_{ij}\geq 0$ is the Euclidean
travel cost.  Each agent $k\in\mathcal{K}=\{A,B,C\}$ receives the same route
budget $B$ and starts at the depot.  A non-depot move from $i$ to $j$ is admitted
only when
\[
    \ell_{ij}+\ell_{j0}\leq b_k,
\]
where $b_k$ is the remaining budget before the move.  Thus every accepted
partial route can still return to the depot.  Equivalently, the closed route
$P_k=(v_{k,0},\ldots,v_{k,T_k})$, with
$v_{k,0}=v_{k,T_k}=0$, satisfies
\begin{equation}
    \sum_{t=1}^{T_k}
    \ell_{v_{k,t-1},v_{k,t}}
    \leq B,
    \qquad
    \forall k\in\mathcal{K}.
    \label{eq:app-route-budget}
\end{equation}
The three benchmarks share this movement rule but use different sample-path
rewards.

\paragraph{Multi-Agent Prize Collection (MAPP-PC).}
This benchmark is a synchronous Team Orienteering variant
\citep{chao1996team}.  Each non-depot node $i$ carries prize $q_i\geq0$.
Let $y_i=1$ iff at least one agent reaches $i$.  Simultaneous arrivals at the
same node collect its prize only once.  The evaluator therefore computes
\begin{equation}
    R_{\mathrm{PC}}(P)
    =
    \sum_{i\in V\setminus\{0\}}q_i
    \mathbf{1}\!\left\{
        \exists k,t:\;v_{k,t}=i
    \right\},
    \qquad
    \max_P R_{\mathrm{PC}}(P),
    \label{eq:mapp-pc}
\end{equation}
subject to Eq.~\eqref{eq:app-route-budget}.

\paragraph{\textit{Agent roles and callable interfaces}.}
The three agents choose from the same currently feasible destination set but
receive different representations of it.  Agent A receives a sparse
prize dictionary together with the full distance matrix, Agent B receives a
structured graph and the current teammate positions, and Agent C receives a
compact distance row and dense prize vector.  The learned policies must
expose the following exact Python interfaces:
\begin{lstlisting}[language=Python,numbers=none,xleftmargin=1em]
def select_next_A(
    current: int,
    unvisited_prizes: dict,
    dist_mat: np.ndarray,
    budget_left: float,
) -> int:

def select_next_B(
    my_state: dict,
    teammate_positions: list,
    graph: dict,
    budget_left: float,
) -> int:

def select_next_C(
    current: int,
    dist_row: np.ndarray,
    remaining_prizes: np.ndarray,
    remaining_budget: float,
) -> int:
\end{lstlisting}
Each function returns a node identifier, with node \(0\) denoting return to
the depot.  The evaluator owns the state transition and admits a proposed
non-depot node only if it is in range, has not previously been visited by that
agent, and preserves enough budget to return to the depot; exceptions and
malformed outputs are mapped to the depot.

\paragraph{\textit{Instance generation}.}
Training uses five batches of ten instances with $N=40$ nodes, including a
depot fixed at $(0.5,0.5)$; all customer coordinates are sampled uniformly
from $[0,1]^2$.  Customer prizes are generated from a mixture of two or three
Gaussian hotspots whose centers lie in $[0.1,0.9]^2$, with standard
deviations in $[0.08,0.15]$ and weights in $[0.6,1.0]$, plus a uniform
baseline of $0.1$.  Non-depot prizes are normalized to unit mean.  The
per-agent budget is $1.2$ times a geometric estimate combining the mean
depot distance and the expected nearest-neighbor spacing.  The test sets
contain 20 independently generated instances at each of $N\in\{50,80\}$
under the same structural distribution and a disjoint seed.

\paragraph{Multi-Agent Area Reconnaissance (MAPP-AR).}
This benchmark is an evaluator-specific informative-sensing surrogate inspired
by multi-robot informative path planning \citep{singh2009efficient}.  Each
non-depot node initially has unit residual uncertainty,
\[
    u_i^{(0)}
    =
    \begin{cases}
        0, & i=0,\\
        1, & i\neq 0.
    \end{cases}
\]
For spatial correlation parameter $\sigma>0$, define
\begin{equation}
    \rho_{ij}
    =
    \exp\left(
        -\frac{\ell_{ij}^{2}}{2\sigma^{2}}
    \right).
\end{equation}
When a scan event occurs at node $s$, the implementation updates every node
sequentially as
\begin{equation}
    u_j \leftarrow u_j(1-\rho_{sj}),
    \qquad
    u_s\leftarrow0,
    \qquad
    u_0\leftarrow0.
    \label{eq:mapp-ar-update}
\end{equation}
The scan events of all agents are applied in deterministic rollout order, so
repeated scans contribute repeated multiplicative updates.  If
$U_0=\sum_i u_i^{(0)}$ and $U_T=\sum_i u_i^{(T)}$, the reported objective is
\begin{equation}
    R_{\mathrm{AR}}=U_0-U_T,
    \qquad
    \max_P R_{\mathrm{AR}},
    \label{eq:mapp-ar}
\end{equation}
which is equivalent to minimizing final residual uncertainty for the fixed
initial field.

\paragraph{\textit{Agent roles and callable interfaces}.}
The role split mirrors MAPP-PC, but prizes are replaced by current residual
uncertainty.  Agent A has the full distance matrix and a sparse uncertainty
map, Agent B has the structured team-and-graph view, and Agent C has dense
local-distance and uncertainty vectors:
\begin{lstlisting}[language=Python,numbers=none,xleftmargin=1em]
def select_next_A(
    current: int,
    uncertainty_dict: dict,
    dist_mat: np.ndarray,
    budget_left: float,
) -> int:

def select_next_B(
    my_state: dict,
    teammate_positions: list,
    graph: dict,
    budget_left: float,
) -> int:

def select_next_C(
    current: int,
    dist_row: np.ndarray,
    uncertainty_vector: np.ndarray,
    remaining_budget: float,
) -> int:
\end{lstlisting}
The returned integer is the next scan node, or \(0\) to return to the depot.
The evaluator applies the same route-feasibility guard as MAPP-PC and retains
exclusive authority over the uncertainty update in
Eq.~\eqref{eq:mapp-ar-update}.

\paragraph{\textit{Instance generation}.}
As in MAPP-PC, every instance places a central depot and samples the remaining
coordinates uniformly in the unit square.  Training consists of five batches
of ten $N=40$ instances.  To induce a streaming shift in sensing locality,
the correlation length is fixed within a batch and follows
$\sigma\in(0.08,0.12,0.10,0.15,0.09)$ across the five batches.  Budgets use
the same geometric construction as MAPP-PC, with the target number of visits
scaled by $1/(3\times2.5)$.  Testing uses 20 instances at each of
$N\in\{50,80\}$ and $\sigma=0.11$, an intermediate value not used in
training.

\paragraph{Multi-Agent Dynamic Servicing (MAPP-DS).}
This benchmark adapts dynamic team orienteering
\citep{kirac2025dynamic} to recurring, time-sensitive requests.  Request
$\rho$ is represented by location $i_\rho$, integer arrival step $\tau_\rho$,
value $w_\rho>0$, and time-to-live $\delta_\rho\geq0$; it is active through
\[
    \tau_\rho\leq t\leq\tau_\rho+\delta_\rho.
\]
One graph hop consumes one simulation step, irrespective of its Euclidean
length, while the route budget still consumes Euclidean distance.  On arrival
at node $i$ at time $t$, the evaluator serves and removes the highest-value
active request at $i$, if one exists.  For the arrival stream
$\Omega$ stored in an instance, the sample-path objective is
\begin{equation}
    R_{\mathrm{DS}}(P;\Omega)
    =
    \sum_{\rho\in\Omega}w_\rho z_\rho,
    \qquad
    \max_P R_{\mathrm{DS}}(P;\Omega),
    \label{eq:mapp-ds}
\end{equation}
where $z_\rho=1$ iff request $\rho$ is selected while active.  The random
stream is generated once from the saved instance seed, making method
comparisons deterministic; averaging over held-out instances estimates
performance across streams.

\paragraph{\textit{Agent roles and callable interfaces}.}
At every decision epoch, each role observes only the most valuable active
request at each location.  Agent A receives a sparse
\(\{\text{node}:(\text{value},\text{time-to-live})\}\) mapping and the full
distance matrix; Agent B receives these attributes inside a structured graph
along with teammate positions; Agent C receives dense value and deadline
vectors with its current distance row:
\begin{lstlisting}[language=Python,numbers=none,xleftmargin=1em]
def select_next_A(
    current: int,
    request_dict: dict,
    dist_mat: np.ndarray,
    budget_left: float,
) -> int:

def select_next_B(
    my_state: dict,
    teammate_positions: list,
    graph: dict,
    budget_left: float,
) -> int:

def select_next_C(
    current: int,
    dist_row: np.ndarray,
    value_vector: np.ndarray,
    deadline_vector: np.ndarray,
    remaining_budget: float,
) -> int:
\end{lstlisting}
Each policy returns the next node identifier, with \(0\) requesting depot
return.  The evaluator validates route feasibility and, upon an accepted
arrival, selects and removes the highest-value request that remains active at
that node.

\paragraph{\textit{Instance generation}.}
Training contains five batches of ten instances with $N=40$ nodes, a central
depot, and uniformly sampled customer coordinates.  The batch-wise mean
arrival-rate regimes are
$\bar\lambda\in(0.08,0.12,0.10,0.15,0.09)$.  At each non-depot node,
$\lambda_i$ is sampled from
$\operatorname{Gamma}(2,\bar\lambda/2)$; its exponential value-rate
parameter is
$\mu_i=\operatorname{clip}(1.5-0.5\lambda_i/\bar\lambda,0.5,3.0)$, coupling
frequent locations with larger expected values.  A saved per-instance seed
deterministically regenerates the stream: node $i$ fires at each step with
probability $1-e^{-\lambda_i}$, request values are exponential with mean
$1/\mu_i$, and lifetimes are uniform integers from 3 to 10 steps.  Test sets
contain 20 instances at each of $N\in\{50,80\}$ with the unseen intermediate
regime $\bar\lambda=0.11$.

\appsubsection{Multi-Agent Scheduling}
\label{app:scheduling}

The scheduling family uses event-driven decision epochs.  Each learned
function controls one machine, worker, or processing stage and chooses only
from the jobs currently released and ready.

\paragraph{Multi-Agent Job Dispatch (MAS-JD).}
This is identical-parallel-machine scheduling with release dates and total
weighted tardiness, denoted $P|r_j|\sum_j w_jT_j$
\citep{kacem2012branch}.  There are three machines, one per role.  Job $j$
has release time $r_j$, processing time $p_j$, due date $d_j$, weight
$w_j\geq0$, and completion time $C_j$.  Each job is assigned exactly once,
machines process at most one job at a time, and
\[
    T_j=\max\{0,C_j-d_j\}.
\]
The reported cost is
\begin{equation}
    C_{\mathrm{JD}}
    =
    \sum_{j\in\mathcal{J}}w_jT_j,
    \qquad
    \min C_{\mathrm{JD}}.
    \label{eq:mas-jd}
\end{equation}
The search engine receives $-C_{\mathrm{JD}}$ so that internal scores remain
higher-is-better, whereas the tables report the positive lower-is-better cost.

\paragraph{\textit{Agent roles and callable interfaces}.}
Each role owns one of the three identical machines but receives a different
dispatch view.  Agent A sees its local availability and a dictionary
\(\{\text{job}:(p_j,d_j,w_j)\}\); Agent B additionally sees teammate
availability; Agent C receives the complete machine-busy vector and a dense
job-feature matrix whose columns are \((p_j,d_j,w_j)\):
\begin{lstlisting}[language=Python,numbers=none,xleftmargin=1em]
def select_next_A(
    my_busy_until: float,
    t_now: float,
    job_pool: dict,
    p_mine: float,
) -> int:

def select_next_B(
    my_busy_until: float,
    teammate_busy_until: list,
    job_pool: dict,
    t_now: float,
) -> int:

def select_next_C(
    busy_vector: np.ndarray,
    job_ids: np.ndarray,
    job_features: np.ndarray,
    t_now: float,
) -> int:
\end{lstlisting}
The return value must be an identifier in the released, unassigned job pool,
or the integer sentinel \(-1\) to idle until the next event.  Exceptions,
non-integral outputs, and other identifiers invalidate the rollout rather than
silently changing the schedule.

\paragraph{\textit{Instance generation}.}
The five training batches each contain ten 30-job instances and use
exponential release-time scales
$(1.0,0.6,1.2,0.8,1.5)$ to vary congestion over the stream.  Release times
are sorted, processing times follow
$\operatorname{Gamma}(2,0.5)$ clipped to $[0.2,5.0]$, and due dates are
$d_j=r_j+\eta_jp_j$ with $\eta_j\sim U[1.5,4.0]$.  Weights are drawn from
the multiset $\{1,1,2,3,5\}$.  The test sets contain 20 instances at each of
50 and 80 jobs, use a disjoint seed, and fix the unseen release scale to
$0.9$.

\paragraph{Multi-Agent Skilled-Task Scheduling (MAS-ST).}
This benchmark adapts multi-skill workforce scheduling
\citep{firat2012multiskill} to value-bearing tasks.  Task $j$ has release time
$r_j$, category $c_j$, nominal duration $p_j$, value $v_j$, and deadline
$d_j$.  Worker $k$ has category-dependent skill $s_{kc_j}$, yielding
\begin{equation}
    p^{\mathrm{eff}}_{jk}
    =
    \frac{p_j}{\max\{s_{k,c_j},\varepsilon\}}.
    \label{eq:mas-st-duration}
\end{equation}
Each task is assigned at most once and each worker executes at most one task
at a time.  A late task is allowed to finish and consumes worker time, but
earns zero value.  Hence
\begin{equation}
    R_{\mathrm{ST}}
    =
    \sum_{j\in\mathcal{J}}
    v_j\mathbf{1}\{C_j\leq d_j\},
    \qquad
    \max R_{\mathrm{ST}}.
    \label{eq:mas-st}
\end{equation}
There is therefore no hard constraint $C_j\leq d_j$; the deadline enters only
through the reward indicator.

\paragraph{\textit{Agent roles and callable interfaces}.}
Agent A observes only its own skill vector and availability, Agent B augments
that local state with the corresponding states of its teammates, and Agent C
receives the global skill and availability matrices.  For A and B,
\(\texttt{task\_pool}[j]=(c_j,p_j,v_j,d_j)\); C's task-feature rows contain
\((j,c_j,p_j,v_j,d_j)\):
\begin{lstlisting}[language=Python,numbers=none,xleftmargin=1em]
def select_next_A(
    my_skills: np.ndarray,
    my_busy_until: float,
    task_pool: dict,
    t_now: float,
) -> int:

def select_next_B(
    my_state: dict,
    teammate_states: list,
    task_pool: dict,
    t_now: float,
) -> int:

def select_next_C(
    skill_matrix: np.ndarray,
    busy_vector: np.ndarray,
    task_features: np.ndarray,
    t_now: float,
) -> int:
\end{lstlisting}
Each function returns a currently released task for which its worker is
qualified, or \(-1\) to idle.  The evaluator alone assigns the task, computes
the skill-adjusted duration, and advances the event queue.

\paragraph{\textit{Instance generation}.}
Training uses five ten-instance batches of 30 tasks with exponential
release-time scales $(1.0,0.7,1.3,0.9,1.1)$.  Task categories are sampled
uniformly from three skills; nominal durations follow
$\operatorname{Gamma}(2,0.5)$ clipped to $[0.3,4.0]$, values are drawn from
$\{1,2,3,5\}$, and deadlines use
$d_j=r_j+\eta_jp_j$ with $\eta_j\sim U[2,5]$.  Each $3\times3$ skill matrix
starts with diagonal specialization $0.8$ and off-diagonal competence $0.4$,
adds independent Gaussian noise with standard deviation $0.08$, and is
clipped to $[0.1,1.0]$.  Testing contains 20 instances at each of 50 and 80
tasks with release scale $1.0$ and independently sampled skill matrices.

\paragraph{Multi-Agent Flow-Shop Scheduling (MAS-FS).}
This is a three-machine flow shop with release dates and a total weighted
completion-time objective \citep{schulz1996weightedflow}.  Every job $j$ has
release time $r_j$, processing time $p_{js}$ at stage
$s\in\{1,2,3\}$, and weight $w_j$.  There is one machine at each stage and
jobs visit the stages in order.  The evaluator permits a different dispatch
order at each stage (a non-permutation schedule), subject to
\begin{equation}
    S_{j1}\geq r_j,
    \qquad
    S_{j,s+1}\geq C_{js},
    \label{eq:mas-fs-precedence}
\end{equation}
and non-overlap on every stage.  All jobs must finish, and the reported cost is
\begin{equation}
    C_{\mathrm{FS}}
    =
    \sum_{j\in\mathcal{J}}w_jC_{j3},
    \qquad
    \min C_{\mathrm{FS}}.
    \label{eq:mas-fs}
\end{equation}
As in Job Dispatch, RELIC internally maximizes $-C_{\mathrm{FS}}$ but reports
the positive cost.

\paragraph{\textit{Agent roles and callable interfaces}.}
The roles are fixed to stages rather than interchangeable machines.  Agent A
dispatches the first-stage buffer from local timing information; Agent B
additionally observes upstream occupancy and downstream congestion; Agent C
receives the full three-stage busy vector and a global job table.  Each buffer
maps a job identifier to its processing time at that stage and its weight:
\begin{lstlisting}[language=Python,numbers=none,xleftmargin=1em]
def select_next_A(
    my_buffer: dict,
    my_busy_until: float,
    t_now: float,
    dummy: float,
) -> int:

def select_next_B(
    my_buffer: dict,
    upstream_busy: float,
    downstream_buf_size: int,
    my_busy_until: float,
) -> int:

def select_next_C(
    my_buffer: dict,
    pipeline_state: np.ndarray,
    job_table: np.ndarray,
    t_now: float,
) -> int:
\end{lstlisting}
The final argument of A is a fixed compatibility placeholder.  A policy
returns a job in its current stage buffer or \(-1\) to idle.  The evaluator
enforces stage precedence and machine non-overlap and requires every job to
complete all three stages.

\paragraph{\textit{Instance generation}.}
Each of five training batches contains ten 25-job instances.  Half of the
jobs arrive in a burst uniformly over $[0,0.5]$, while the remainder arrive
after $0.5$ with exponentially distributed increments of scale $0.4$.
Batch-specific mean processing profiles
$(0.3,1.2,0.8)$, $(0.3,0.6,1.5)$, $(0.5,1.5,0.3)$,
$(1.0,0.3,1.2)$, and $(0.4,0.4,0.4)$ move the bottleneck among stages.
Stages with mean below $0.5$ use a tight uniform duration distribution;
other stages use clipped Gamma durations, and job weights are drawn from
$\{1,1,2,3,5\}$.  Testing uses 20 instances at each of 50 and 80 jobs with
the unseen mild-B-bottleneck profile $(1.0,1.2,1.0)$.
\appsubsection{Heterogeneous-Role Combinatorial Optimization}
\label{app:heterogeneous-co}

The second suite uses three specialized decision modules inside one
constraint-owning constructive solver.  Policies nominate identifiers, while
the evaluator enforces uniqueness, precedence, capacity, and deterministic
repair.  The coefficients below are the actual scoring coefficients used in
the repository, not generic placeholders.

\paragraph{Heterogeneous Multi-Depot Multiple Traveling Salesman Problem (HMTSP).}
This benchmark extends the heterogeneous multi-depot multiple-TSP, in which
vehicle-dependent costs make the tours non-interchangeable
\citep{oberlin2009hmdmtsp}.  Vehicle $k\in\{A,B,C\}$ has depot $d_k$, travel
multiplier $\gamma_k$, service multiplier $s_{k,c}$ for task type $c$, and
soft budget $B_k$.  If
$P_k=(d_k,v_{k,1},\ldots,v_{k,n_k},d_k)$, its evaluator cost is
\begin{equation}
    L_k
    =
    \sum_{h=1}^{n_k}
    \left(
        \gamma_k\ell_{v_{k,h-1},v_{k,h}}
        +s_{k,c(v_{k,h})}
    \right)
    +
    \gamma_k\ell_{v_{k,n_k},d_k}.
    \label{eq:hmtsp-route-cost}
\end{equation}
Every customer is assigned to exactly one route.  Budget excess is penalized
rather than treated as a hard infeasibility:
\begin{equation}
    C_{\mathrm{HMTSP}}
    =
    \sum_k L_k
    +0.5\max_k L_k
    +5\sum_k\max\{L_k-B_k,0\},
    \qquad
    \min C_{\mathrm{HMTSP}}.
    \label{eq:hmtsp}
\end{equation}
Agent A receives incremental costs and remaining budget; Agent B additionally
receives candidate task types and its current route cost; Agent C receives
all three current route costs.  The evaluator returns
$-C_{\mathrm{HMTSP}}$ to the optimizer and reports the positive cost.

\paragraph{\textit{Agent roles and callable interfaces}.}
The roles coincide with the three heterogeneous vehicles but deliberately
expose different decision contexts: A emphasizes residual budget, B exposes
task types and accumulated local cost, and C exposes cross-vehicle route
balance.  Their exact constructive interfaces are:
\begin{lstlisting}[language=Python,numbers=none,xleftmargin=1em]
def select_next_A(
    current: int,
    candidate_ids: np.ndarray,
    candidate_costs: np.ndarray,
    budget_left: float,
) -> int:

def select_next_B(
    candidate_ids: np.ndarray,
    candidate_costs: np.ndarray,
    task_types: np.ndarray,
    route_cost: float,
) -> int:

def select_next_C(
    current: int,
    candidate_ids: np.ndarray,
    candidate_costs: np.ndarray,
    team_route_costs: np.ndarray,
) -> int:
\end{lstlisting}
The return value is an unassigned customer identifier; \(-1\) may skip the
current vehicle's turn.  Invalid choices use a deterministic feasible
fallback, and if all vehicles skip, the evaluator forces the globally
cheapest remaining assignment to guarantee completion.

\paragraph{\textit{Instance generation}.}
For each instance, customer coordinates and three depot coordinates are
sampled independently and uniformly from $[0,1]^2$, and each customer is
assigned one of three task types uniformly.  Vehicle service and travel
multipliers are initially sampled from $U[0.8,1.3]$; vehicle $k$ then
receives a specialization multiplier from $U[0.25,0.55]$ for task type $k$.
Each soft route budget is set to $0.42N$.  Training comprises five batches of
ten 50-customer instances.  Test files contain 20 independently generated
instances at each of 100 and 200 customers.

\paragraph{Flying Sidekick Traveling Salesman Problem (FSTSP).}
FSTSP coordinates one truck and one drone
\citep{murray2015fstsp}.  In each synchronized construction epoch, Agent A
may select one drone-eligible customer whose conservative out-and-back flight
fits the endurance limit; Agent B may select an intermediate truck customer;
and Agent C selects an endurance-feasible rendezvous using the truck and drone
arrival times.  Every customer is removed from the unserved set exactly once,
and both vehicles synchronize at the selected recovery node.  If
$T_{\mathrm{sync}}$ is the elapsed synchronized completion time and
$D_{\mathrm{truck}}$ is total truck distance, the implemented objective is
\begin{equation}
    C_{\mathrm{FSTSP}}
    =
    T_{\mathrm{sync}}
    +0.05D_{\mathrm{truck}},
    \qquad
    \min C_{\mathrm{FSTSP}}.
    \label{eq:fstsp}
\end{equation}
The completion-time term includes the final truck return to the depot.

\paragraph{\textit{Agent roles and callable interfaces}.}
The three functions implement distinct stages of one synchronized sortie:
Agent A nominates a drone customer, Agent B nominates a truck customer after
that reservation, and Agent C chooses the recovery node from synchronized
truck and drone arrival times:
\begin{lstlisting}[language=Python,numbers=none,xleftmargin=1em]
def select_next_A(
    candidate_ids: np.ndarray,
    drone_roundtrip_times: np.ndarray,
    truck_detours: np.ndarray,
    endurance: float,
) -> int:

def select_next_B(
    current: int,
    candidate_ids: np.ndarray,
    truck_distances: np.ndarray,
    drone_customer: int,
) -> int:

def select_next_C(
    candidate_ids: np.ndarray,
    truck_etas: np.ndarray,
    drone_etas: np.ndarray,
    endurance: float,
) -> int:
\end{lstlisting}
A and B may return \(-1\) when skipping is feasible; C must select one of the
endurance-feasible recovery identifiers.  Candidate arrays are constructed
by the evaluator, which repairs malformed outputs deterministically and owns
all service and synchronization updates.

\paragraph{\textit{Instance generation}.}
The shared depot is fixed at $(0.5,0.5)$ and all customer coordinates are
uniform in $[0,1]^2$.  Each customer is independently drone-eligible with
probability $0.7$.  Drone endurance is sampled from $U[0.65,0.9]$, drone
speed from $U[1.8,2.5]$, and truck speed is fixed to $1$.  Training uses
five batches of ten 50-customer instances; testing uses 20 instances at
each of 100 and 200 customers with a disjoint generator seed.

\paragraph{Flexible Job Shop Scheduling with Automated Guided Vehicles (FJAGV).}
FJAGV jointly schedules machine processing and material handling, following
the integrated machine/AGV setting of \citet{bilge1995agv} and the modern
FJSP-AGV decomposition of \citet{cheng2025fjspagv}.  Each job $j$ contains an
ordered sequence of operations.  Agent A assigns every newly
precedence-ready operation to an eligible machine; Agent C assigns pending
transport requests to one of two unit-capacity AGVs; after delivery, Agent B
dispatches operations waiting at each free machine.  The event-driven
evaluator enforces job precedence, one operation per machine, one transport
per AGV, and delivery before processing.  Let $C_j$ be job completion,
$d_j$ its due date, $w_j$ its weight, and $D_{\mathrm{AGV}}$ total loaded plus
deadhead travel.  The exact cost is
\begin{equation}
    C_{\mathrm{FJAGV}}
    =
    C_{\max}
    +0.2\sum_jw_j\max\{C_j-d_j,0\}
    +0.02D_{\mathrm{AGV}},
    \qquad
    \min C_{\mathrm{FJAGV}}.
    \label{eq:fjagv}
\end{equation}

\paragraph{\textit{Agent roles and callable interfaces}.}
Agent A performs operation-to-machine assignment, Agent B sequences delivered
operations at a free machine, and Agent C dispatches a free AGV to a pending
transport request.  This decomposition produces three different identifier
spaces and observation schemas:
\begin{lstlisting}[language=Python,numbers=none,xleftmargin=1em]
def select_next_A(
    operation_id: int,
    machine_ids: np.ndarray,
    processing_times: np.ndarray,
    machine_ready_times: np.ndarray,
) -> int:

def select_next_B(
    machine_id: int,
    operation_ids: np.ndarray,
    processing_times: np.ndarray,
    due_dates: np.ndarray,
) -> int:

def select_next_C(
    agv_id: int,
    request_ids: np.ndarray,
    travel_times: np.ndarray,
    ready_times: np.ndarray,
) -> int:
\end{lstlisting}
The returned identifier must belong to the supplied feasible machine,
operation, or request array, respectively.  The evaluator replaces malformed
outputs with a deterministic feasible choice and remains responsible for
precedence, transport, resource occupancy, and event timing.

\paragraph{\textit{Instance generation}.}
Every job contains three ordered operations and there are five machines.
For each operation, between two and five eligible machines are selected
without replacement; machine processing times are integers from 4 to 24.
The job due date is its sum of fastest eligible processing times multiplied
by a factor from $U[1.8,3.2]$, and its weight is uniform on
$\{1,\ldots,5\}$.  The five machine locations and one depot are sampled in
$[0,1]^2$; pairwise Euclidean distances are multiplied by $5$ to obtain
travel times for the two AGVs.  The training stream has five ten-instance
batches with 50 jobs, while test sets contain 20 instances at each of 100
and 200 jobs.

\paragraph{Multi-Project Multi-Skilled Resource-Constrained Project Scheduling (MRCPS).}
This benchmark is a compact multi-skill resource-constrained multi-project
scheduling problem \citep{torba2024msrcmpsp}.  Each non-preemptive task $i$
belongs to project $p(i)$, has nominal duration $a_i$, requires skill
$c(i)$, and may start only after all predecessors complete.  A worker $w$ is
qualified iff skill level $s_{w,c(i)}>0$; assigning that worker gives
\begin{equation}
    a^{\mathrm{eff}}_{iw}
    =
    \frac{a_i}{0.7+0.15s_{w,c(i)}}.
    \label{eq:mrcps-duration}
\end{equation}
At each event time, Agent A selects a project containing a schedulable task,
Agent B selects such a task, and Agent C assigns a qualified free worker.
Let $C_p=\max_{i:p(i)=p}C_i$, due date $d_p$, project weight $w_p$, and worker
cost rate $\kappa_w$.  With
$C_{\mathrm{labor}}=\sum_i\kappa_{w(i)}a^{\mathrm{eff}}_{i,w(i)}$, the
implemented objective is
\begin{equation}
    C_{\mathrm{MRCPS}}
    =
    0.25\max_p C_p
    +\sum_pw_p\max\{C_p-d_p,0\}
    +0.01C_{\mathrm{labor}},
    \qquad
    \min C_{\mathrm{MRCPS}}.
    \label{eq:mrcps}
\end{equation}

\paragraph{\textit{Agent roles and callable interfaces}.}
The constructive hierarchy is explicit in the three return spaces: Agent A
chooses a project, Agent B chooses a ready task within that project, and Agent
C chooses a currently free worker qualified for that task:
\begin{lstlisting}[language=Python,numbers=none,xleftmargin=1em]
def select_next_A(
    project_ids: np.ndarray,
    due_dates: np.ndarray,
    remaining_task_counts: np.ndarray,
    t_now: float,
) -> int:

def select_next_B(
    project_id: int,
    task_ids: np.ndarray,
    durations: np.ndarray,
    successor_counts: np.ndarray,
) -> int:

def select_next_C(
    task_id: int,
    worker_ids: np.ndarray,
    skill_levels: np.ndarray,
    worker_costs: np.ndarray,
) -> int:
\end{lstlisting}
Each function must return an identifier from its supplied feasible array.
Malformed decisions trigger deterministic fallback selection; the evaluator
then commits the nested decision, computes the effective duration and labor
cost, and updates resource and precedence state.

\paragraph{\textit{Instance generation}.}
Instances contain five approximately equal-sized projects, four skill types,
and ten workers.  Task durations are uniform integers from 3 to 17 and
required skills are uniform over the four types.  Tasks within each project
form a precedence chain; every earlier non-immediate task additionally
becomes a predecessor with probability $0.08$.  Worker skill levels are
uniform integers from 0 to 3, with a repair guaranteeing at least one
qualified worker per skill, and worker costs follow $U[1,3]$.  Project due
dates multiply an estimated two-worker project load by a factor in
$[0.9,1.5]$, and project weights are uniform integers from 1 to 5.
Training contains five batches of ten 50-task instances; testing contains
20 instances at each of 100 and 200 tasks.

\appsubsection{Distributed Constraint Optimization Problem Benchmarks}
\label{app:dcop}

We additionally evaluate RELIC on Distributed Constraint Optimization Problem (DCOP) domains \citep{fioretto2018dcop}. A DCOP is represented by
\[
    \langle\mathcal{X},\mathcal{D},\mathcal{F},\mathcal{A},\alpha\rangle,
\]
where $\mathcal{X}$ is the set of variables, $\mathcal{D}$ their finite domains, $\mathcal{F}$ local cost or utility functions, $\mathcal{A}$ the agents, and $\alpha$ maps each variable to its controlling agent. Using a cost convention, the global objective is
\begin{equation}
    \min_{\mathbf{x}}
    \sum_{f\in\mathcal{F}}
    f(\mathbf{x}_{\mathrm{scope}(f)}).
    \label{eq:dcop-general}
\end{equation}
Each agent controls only its local variable(s), while constraints couple decisions across agents.

\paragraph{Distributed Graph Coloring (DGC).}
Distributed graph coloring is a standard DCOP benchmark
\citep{fioretto2018dcop}.  Given an undirected graph $G=(V,E)$ and a color
set $\mathcal{C}$, each vertex $i\in V$ induces a variable
\[
    x_i\in\mathcal{C},
    \qquad
    |\mathcal{C}|=3.
\]
Every edge $(i,j)$ has nonnegative conflict weight $w_{ij}$.  The evaluator
uses
\begin{equation}
    f_{ij}(x_i,x_j)
    =
    \begin{cases}
        w_{ij}, & x_i=x_j,\\
        0, & x_i\neq x_j.
    \end{cases}
\end{equation}
Thus
\begin{equation}
    C_{\mathrm{DGC}}(\mathbf{x})
    =
    \sum_{(i,j)\in E}
    w_{ij}\mathbf{1}\{x_i=x_j\},
    \qquad
    \min_{\mathbf{x}}C_{\mathrm{DGC}}(\mathbf{x}).
    \label{eq:dgc}
\end{equation}

\paragraph{\textit{Agent roles and callable interfaces}.}
At each synchronous sweep, \texttt{cur} contains the current colors of the
role's \(m\) owned vertices, \texttt{delta} is the corresponding
\(m\times3\) local-cost landscape, \texttt{u} is an evaluator-supplied
deterministic random tensor, and \texttt{t} is the sweep index.  Agent A
specializes in high-risk conflict repair, Agent B in improvement opportunity,
and Agent C in coordination stability:
\begin{lstlisting}[language=Python,numbers=none,xleftmargin=1em]
def select_values_A(
    cur: np.ndarray,
    delta: np.ndarray,
    u: np.ndarray,
    t: int,
    violation_now: np.ndarray,
    repair_gain: np.ndarray,
) -> np.ndarray:

def select_values_B(
    cur: np.ndarray,
    delta: np.ndarray,
    u: np.ndarray,
    t: int,
    opportunity_gain: np.ndarray,
    flexibility: np.ndarray,
) -> np.ndarray:

def select_values_C(
    cur: np.ndarray,
    delta: np.ndarray,
    u: np.ndarray,
    t: int,
    peer_value_hist: np.ndarray,
    peer_churn_rate: np.ndarray,
    local_trend: np.ndarray,
) -> np.ndarray:
\end{lstlisting}
Each role returns an integer vector of shape \((m,)\) with entries in
\(\{0,1,2\}\).  Candidate code may use \texttt{u} but may not invoke an
independent random-number generator.  The strict evaluator rejects execution
errors, malformed shapes, and values outside the color domain.

\paragraph{\textit{Instance generation}.}
All instances have 120 variables, three colors, and 50 synchronous sweeps.
Training uses five batches of ten fixed-edge-count Erd\H{o}s--R\'enyi graphs
with mean degrees $(3,5,7,9,12)$; edge weights are uniform integers from 1
to 100.  Initial colors are uniform, and variables are assigned in balanced
thirds to the constraint-risk, improvement-opportunity, and
coordination-coupling roles using incident edge statistics.  Zero-weight
self-loops pad training edge arrays to a common replay-compatible shape but
do not affect costs.  The test split contains 50 independently generated
graphs with mean degree $6$.

\paragraph{Distributed Multi-Event Scheduling (DiMES).}
The source DiMES problem coordinates event times across distributed
participants \citep{maheswaran2004dimes}.  Our evaluator uses its
private-events-as-variables (PEAV) form: a variable $x_i$ represents one
person--meeting attendance and selects a slot
\begin{equation}
    x_i\in\{0,\ldots,19\}.
\end{equation}
Let $m(i)$ and $p(i)$ be the meeting and person associated with variable
$i$.  Each person attends exactly two meetings; let $\bar i$ denote the other
attendance variable of the same person.  The implemented meeting time is the
majority slot
\begin{equation}
    q_m(\mathbf{x})
    =
    \arg\max_{h\in\{0,\ldots,19\}}
    \sum_{i:m(i)=m}\mathbf{1}\{x_i=h\},
\end{equation}
with the lowest indexed slot breaking ties.  If $a_{ih}$ is the preference
cost of attendance $i$ at slot $h$ and
$\tau_{m,m'}$ is inter-meeting travel time, the exact soft cost is
\begin{equation}
    \begin{aligned}
    C_{\mathrm{DiMES}}
    ={}&
    100\sum_i\mathbf{1}\{x_i\neq q_{m(i)}(\mathbf{x})\}
    \\
    &+
    10\sum_{\{i,\bar i\}}
    \max\!\left\{
        0,\,
        \tau_{m(i),m(\bar i)}
        -|x_i-x_{\bar i}|
    \right\}
    \\
    &+
    \sum_i a_{i,x_i},
    \qquad
    \min_{\mathbf{x}}C_{\mathrm{DiMES}}.
    \end{aligned}
    \label{eq:dimes}
\end{equation}
The second sum contains each person's unordered attendance pair once.

\paragraph{\textit{Agent roles and callable interfaces}.}
DiMES uses the same three semantic DCOP roles but applies them to attendance
variables.  Here \texttt{cur} is the current slot vector,
\texttt{delta}\(\in\mathbb{R}^{m\times20}\) gives the conditional local cost
of every slot, \texttt{u} supplies deterministic exploration, and
\texttt{t} is the synchronous sweep index.  A emphasizes current constraint
risk, B exploitable cost reduction, and C peer agreement and stability:
\begin{lstlisting}[language=Python,numbers=none,xleftmargin=1em]
def select_values_A(
    cur: np.ndarray,
    delta: np.ndarray,
    u: np.ndarray,
    t: int,
    violation_now: np.ndarray,
    repair_gain: np.ndarray,
) -> np.ndarray:

def select_values_B(
    cur: np.ndarray,
    delta: np.ndarray,
    u: np.ndarray,
    t: int,
    opportunity_gain: np.ndarray,
    flexibility: np.ndarray,
) -> np.ndarray:

def select_values_C(
    cur: np.ndarray,
    delta: np.ndarray,
    u: np.ndarray,
    t: int,
    peer_value_hist: np.ndarray,
    peer_churn_rate: np.ndarray,
    local_trend: np.ndarray,
) -> np.ndarray:
\end{lstlisting}
Each function returns a length-\(m\) integer array of proposed slots in
\(\{0,\ldots,19\}\).  The code must consume evaluator-provided
\texttt{u} for any randomized behavior; the strict runtime rejects invalid
signatures, exceptions, shapes, or domain values.

\paragraph{\textit{Instance generation}.}
Every instance has 60 people, two meeting attendances per person
($N=120$ PEAV variables), 20 time slots, and 50 synchronous sweeps.
The five ten-instance training batches use respectively
$M\in(16,14,12,10,8)$ meetings; attendance counts differ by at most one
within an instance, and sampling prevents a person from attending the same
meeting twice.  Meetings are assigned among four locations.  Symmetric
inter-location travel times are uniform integers from 6 to 10, attendance
preference costs follow $U[0,10]$, and initial slots are uniform.  Balanced
semantic ownership prioritizes travel pressure, preference-saving
opportunity, or meeting-size coupling.  Testing uses 50 instances with
$M=12$, hence ten attendances per meeting.

\paragraph{Sensor Network Distributed Constraint Satisfaction Problem (SensorDCSP).}
SensorDCSP is a distributed sensor-allocation benchmark introduced by
\citet{bejar2005sensordcsp}.  In our soft optimization adaptation, each
sensor $s$ has domain $\{0,1,2,3\}$: value $0$ means sleep and values
$1$--$3$ index up to three visible candidate targets
$g_s(x_s)\in\mathcal{T}\cup\{-1\}$.  Let
\[
    n_t(\mathbf{x})
    =
    \sum_s\mathbf{1}\{g_s(x_s)=t\}
\]
be target coverage, $v_t>0$ target value, and $I_{ss'}=1$ when sensors
$s,s'$ are incompatible.  Every target ideally receives exactly $k=3$
compatible sensors.  The evaluator minimizes
\begin{equation}
    C_{\mathrm{Sensor}}
    =
    \sum_{t\in\mathcal{T}}
    v_t\mathbf{1}\{n_t(\mathbf{x})<3\}
    +
    \sum_{t\in\mathcal{T}}\max\{0,n_t(\mathbf{x})-3\}
    +
    100\!\!\!\!\!\!\!\!\!\!\!\!
    \sum_{\substack{s<s'\\g_s(x_s)=g_{s'}(x_{s'})\neq-1}}\!\!\!\!\!\!\!\!\!\!\!\!
    I_{ss'},
    \qquad
    \min_{\mathbf{x}}C_{\mathrm{Sensor}}.
    \label{eq:sensordcsp}
\end{equation}
The first term charges the full target value whenever coverage is below three,
the second discourages excess coverage, and the third penalizes incompatible
sensor pairs assigned to the same target.

\paragraph{\textit{Agent roles and callable interfaces}.}
All roles receive their current domain values, the
\(m\times4\) conditional-cost matrix \texttt{delta}, deterministic
exploration tensor \texttt{u}, sweep index, candidate-target table, and domain
validity mask.  Agent A emphasizes coverage repair and incompatibility risk,
Agent B target value and supply scarcity, and Agent C coverage dynamics and
coordination:
\begin{lstlisting}[language=Python,numbers=none,xleftmargin=1em]
def select_values_A(
    cur: np.ndarray,
    delta: np.ndarray,
    u: np.ndarray,
    t: int,
    candidate_target: np.ndarray,
    valid_mask: np.ndarray,
    target_value: np.ndarray,
    target_coverage: np.ndarray,
    incompatibility_cost: np.ndarray,
) -> np.ndarray:

def select_values_B(
    cur: np.ndarray,
    delta: np.ndarray,
    u: np.ndarray,
    t: int,
    candidate_target: np.ndarray,
    valid_mask: np.ndarray,
    target_value: np.ndarray,
    target_coverage: np.ndarray,
    target_supply: np.ndarray,
) -> np.ndarray:

def select_values_C(
    cur: np.ndarray,
    delta: np.ndarray,
    u: np.ndarray,
    t: int,
    candidate_target: np.ndarray,
    valid_mask: np.ndarray,
    target_coverage: np.ndarray,
    coverage_change: np.ndarray,
    incompatibility_cost: np.ndarray,
) -> np.ndarray:
\end{lstlisting}
The output is a length-\(m\) integer vector in \(\{0,1,2,3\}\).
\texttt{valid\_mask} distinguishes visible-target actions from padded domain
entries; a padded entry resolves to no target and therefore contributes no
coverage.  As in the other DCOP domains, all stochasticity must derive from
\texttt{u}; malformed executables are rejected under strict evaluation.

\paragraph{\textit{Instance generation}.}
Each instance contains 120 sensors, 16 targets, a four-valued sensor domain,
required coverage $k=3$, compatibility probability $0.8$, and 50 synchronous
sweeps.  For coverage coefficient $c$, every sensor--target visibility edge
is sampled independently with probability $c/16$; each sensor retains at most
three visible targets as actionable domain values, in addition to sleep.
Target values follow $U[1,100]$, pairwise sensor compatibility is sampled
symmetrically, and initial domain values are uniform.  Training uses five
ten-instance batches with
$c\in(1.2,1.8,2.4,3.0,3.6)$; balanced ownership is derived from value risk,
available value, and coupling.  Testing contains 50 instances with the
intermediate coefficient $c=2.1$.

\paragraph{Smart Home Device Scheduling (SHDS).}
SHDS originates from the DCOP benchmark for coordinating smart-device
schedules \citep{kluegel2017shds}.  Our evaluator uses a compact
start-time abstraction over a 24-slot horizon.  Device $i$ has duration
$d_i\in\{1,\ldots,6\}$, power $p_i$, preferred start $\hat x_i$, and chooses
$x_i\in\{0,\ldots,17\}$.  Define aggregate load
\begin{equation}
    L_h(\mathbf{x})
    =
    \sum_i
    p_i\mathbf{1}\{x_i\leq h<x_i+d_i\},
    \qquad
    h=0,\ldots,23.
\end{equation}
For price $\theta_h$ and capacity $\kappa_h$, the implemented cost is
\begin{equation}
    C_{\mathrm{SHDS}}
    =
    \sum_{h=0}^{23}\theta_hL_h(\mathbf{x})
    +5\sum_i|x_i-\hat x_i|
    +200\sum_{h=0}^{23}\max\{0,L_h(\mathbf{x})-\kappa_h\},
    \qquad
    \min_{\mathbf{x}}C_{\mathrm{SHDS}}.
    \label{eq:shds}
\end{equation}
This differs from the richer physical simulator of the source SHDS dataset:
homes are used to define semantic ownership and peer coupling, while the
reported objective uses one neighborhood-wide load, price, and capacity
curve.

\paragraph{\textit{Agent roles and callable interfaces}.}
Every role receives current start values, its \(m\times18\) conditional-cost
landscape, deterministic exploration tensor, sweep index, and device duration
and power.  Agent A focuses on capacity-violation repair using current load,
Agent B on price and preference opportunity using capacity slack, and Agent C
on load-trend stabilization:
\begin{lstlisting}[language=Python,numbers=none,xleftmargin=1em]
def select_values_A(
    cur: np.ndarray,
    delta: np.ndarray,
    u: np.ndarray,
    t: int,
    duration: np.ndarray,
    power: np.ndarray,
    load_now: np.ndarray,
    capacity: np.ndarray,
) -> np.ndarray:

def select_values_B(
    cur: np.ndarray,
    delta: np.ndarray,
    u: np.ndarray,
    t: int,
    duration: np.ndarray,
    power: np.ndarray,
    pref_start: np.ndarray,
    price: np.ndarray,
    capacity_slack: np.ndarray,
) -> np.ndarray:

def select_values_C(
    cur: np.ndarray,
    delta: np.ndarray,
    u: np.ndarray,
    t: int,
    duration: np.ndarray,
    power: np.ndarray,
    load_now: np.ndarray,
    load_change: np.ndarray,
    capacity: np.ndarray,
) -> np.ndarray:
\end{lstlisting}
Each function returns a length-\(m\) integer vector of start indices in
\(\{0,\ldots,17\}\).  Independent randomness is prohibited, and the strict
runtime rejects exceptions, incorrect shapes, and values outside the
start-time domain.

\paragraph{\textit{Instance generation}.}
All instances contain 120 devices assigned uniformly to 24 homes, an
18-value start-time domain over 24 slots, and 50 synchronous sweeps.
Durations are uniform integers from 1 to 6, powers follow $U[0.5,3.0]$, and
preferred and initial starts are uniform on $\{0,\ldots,17\}$.  The price
curve combines Gaussian-shaped morning and evening peaks centered at slots
8 and 19 with independent jitter in $[-0.1,0.1]$.  Capacity is constant
over time and equals a tightness multiplier times the perfectly smoothed
average load.  The five ten-instance training batches use multipliers
$(1.1,1.25,1.4,1.6,1.8)$; risk, energy-saving opportunity, and same-home
coupling determine balanced role ownership.  The test split contains 50
instances with tightness $1.35$.

\newpage
\appsection{Methodological Details}
\label{app:methodology-details}

\definecolor{relicpromptframe}{RGB}{64,176,112}
\definecolor{relicpromptback}{RGB}{241,252,246}
\definecolor{relicprompttitle}{RGB}{24,132,77}
\definecolor{relicpromptaccent}{RGB}{17,105,61}
\tcbset{
    relicprompt/.style={
        enhanced,
        breakable,
        colback=relicpromptback,
        colframe=relicpromptframe,
        colbacktitle=relicprompttitle,
        coltitle=white,
        fonttitle=\bfseries,
        boxrule=0.7pt,
        arc=2mm,
        left=2mm,
        right=2mm,
        top=1.5mm,
        bottom=1.5mm,
        before skip=7pt,
        after skip=8pt
    }
}

This section supplies the procedural and prompting details omitted from the
main text.  We retain its notation throughout: \(N\) denotes the number of
agents, \(b\) the batch index, \(\bar{\boldsymbol{\pi}}_b\) the incumbent
team, \(\mathcal{T}_{i,b}\) agent \(i\)'s private program tree,
\(\mathcal{A}^{\mathrm{rev}}_{i,b}\) its revealed-principle archive,
\(\mathcal{P}_b\) the public principle tree, \(\sigma_i\) its callable
interface, and \(\rho\in\mathfrak{R}\) a textual principle.  All optimization
scores are converted by the problem adapter to a higher-is-better convention,
including benchmarks whose reported metric is a cost.

\appsubsection{Complete Batch-Level Procedure}
\label{app:batch-procedure}

\paragraph{Initialization and persistent state.}
For each agent \(i\), the root of \(\mathcal{T}_{i,1}\) contains the
human-written seed program satisfying \(\sigma_i\).  The initial incumbent
\(\bar{\pi}_{i,1}\) is that root program, all
\(\mathcal{A}^{\mathrm{rev}}_{i,1}\) are empty, and
\(\mathcal{P}_1\) contains only a nonselectable system root.  Private trees,
local archives, and the public tree persist across batches; they are not
restarted when the task distribution changes.  This persistence separates
RELIC from restarting an independent AHD run on every
\(\mathcal{B}_b\).

\paragraph{Replay construction.}
Equation~\eqref{eq:mixed-batch} defines the common evaluation set
\(\widetilde{\mathcal{B}}_b\).  In the reported configuration, at most two
past batches are sampled and at most three instances are replayed from each
selected batch.  Per-instance arrays are concatenated with the current batch,
whereas instance-independent fields are retained from
\(\mathcal{B}_b\).  The sampled replay indices are fixed by the run seed.
Crucially, the same \(\widetilde{\mathcal{B}}_b\) is held fixed for all
candidate comparisons in batch \(b\); otherwise
\(\Delta_b\) in Eq.~\eqref{eq:contextual-gain} would conflate program quality
with evaluation-set noise.

\paragraph{Incumbent revalidation.}
At the start of \(b>1\), each agent's previously successful programs are
re-evaluated in the current team context on
\(\widetilde{\mathcal{B}}_b\).  Exact duplicate source strings are evaluated
once.  The best valid program becomes the new \(\bar{\pi}_{i,b}\), and its
stored marginal is reset to zero because it now defines the comparison
baseline.  Revalidation permits a formerly useful branch to regain incumbent
status after a distribution shift and prevents stale historical scores from
being compared directly with current-batch scores.

Algorithm~\ref{alg:app-relic} gives the complete protocol.  The proposal
budget \(H\) counts attempted LLM revisions; an invalid generation therefore
consumes budget even though it does not create a private node.  The reported
experiments use \(H=10\) proposals per agent per batch and maximum private-tree
depth eight.

\begin{algorithm}[t]
\caption{Complete Batch-Level RELIC Procedure}
\label{alg:app-relic}
\begin{algorithmic}[1]
\REQUIRE Seed team \(\bar{\boldsymbol{\pi}}_1\), batches
\(\{\mathcal{B}_b\}_{b=1}^{B}\), proposal budget \(H\)
\STATE Initialize \(\{\mathcal{T}_{i,1},
\mathcal{A}^{\mathrm{rev}}_{i,1}\}_{i=1}^{N}\) and \(\mathcal{P}_1\)
\FOR{\(b=1,\ldots,B\)}
    \STATE Construct the fixed mixed batch
    \(\widetilde{\mathcal{B}}_b\)
    \IF{\(b>1\)}
        \STATE Revalidate previously successful private programs and update
        \(\bar{\boldsymbol{\pi}}_b\)
    \ENDIF
    \FOR{\(i=1,\ldots,N\)}
        \FOR{\(h=1,\ldots,H\)}
            \STATE Form the available operator set
            \(\mathcal{O}_{i,b}\)
            \STATE Select expansion node \(v\in\mathcal{T}_{i,b}\) and
            operator \(o\in\mathcal{O}_{i,b}\)
            \STATE Generate
            \(\widetilde{\pi}_i\) under \(\sigma_i\) using
            \(C^o_{i,b}\)
            \IF{\(\widetilde{\pi}_i\) is invalid or nonexecutable}
                \STATE Record the operator attempt and continue
            \ENDIF
            \STATE Evaluate
            \(\Delta_b(\widetilde{\pi}_i)\) with all other programs fixed
            \STATE Insert a private child and update node/operator statistics
            \IF{its absolute team score exceeds the incumbent score}
                \STATE Set
                \(\bar{\pi}_{i,b}\leftarrow\widetilde{\pi}_i\)
            \ENDIF
        \ENDFOR
    \ENDFOR
    \STATE Distill the strongest positive new behavior from each agent
    \STATE Reveal each incumbent as text and update recipients'
    \(\mathcal{A}^{\mathrm{rev}}_{i,b}\)
    \STATE Cross-distill coordination principles and update
    \(\mathcal{P}_b\)
    \STATE Endorse successfully used public principles and prune weak leaves
    \STATE Append \(\mathcal{B}_b\) to the replay buffer
\ENDFOR
\RETURN \(\bar{\boldsymbol{\pi}}_{B+1}\) and
\(\mathcal{P}_{B+1}\)
\end{algorithmic}
\end{algorithm}

\paragraph{Coordinate updates.}
Only one callable is replaced during a candidate evaluation.  Consequently,
the difference in Eq.~\eqref{eq:contextual-gain} has an unambiguous coordinate:
it measures agent \(i\)'s revision against a common teammate context rather
than comparing two jointly changed teams.  Accepted incumbents remain
executable after every update.  The trade-off is that two revisions that are
useful only when adopted simultaneously may be missed.

\appsubsection{Private Search, Credit, and Candidate Lifecycle}
\label{app:private-search-details}

\paragraph{Two-level tree policy.}
The node rule in Eq.~\eqref{eq:node-uct} and the operator rule in
Eq.~\eqref{eq:operator-uct} are applied sequentially.  At a private node,
every currently available operator is attempted before its UCT score is used.
A node is expandable when it has an untried available operator or no child;
otherwise descent follows the child maximizing \(U_{\mathrm{node}}\), until
an expandable node or the depth limit is reached.  Availability is
context-dependent:
\[
    \mathcal{O}_{i,b}
    =
    \{\mathrm{Reflect}\}
    \cup
    \{\mathrm{Lift}:
        \mathcal{A}^{\mathrm{rev}}_{i,b}\neq\varnothing\}
    \cup
    \{\mathrm{Bridge}:
        \mathcal{V}^{\mathrm{pub}}_{i,b}\neq\varnothing\}.
\]
Thus the first batch begins with Reflect; Lift and Bridge become available
only after text has entered the corresponding memories.  In communication
ablations that deny peer information to \(i\), both Lift and Bridge are
removed, while Reflect remains available because its summary is produced by
the trusted evaluator rather than another agent.

\paragraph{What is and is not back-propagated.}
For a valid child \(v_{\mathrm{new}}\), RELIC stores
\(Q(v_{\mathrm{new}})=\Delta_b(\pi(v_{\mathrm{new}}))\).  Ancestor visit
counts are incremented, but ancestor \(Q\)-values are not replaced by a
maximum-descendant or Monte Carlo return.  This convention preserves the
interpretation of \(Q(v)\) as the measured contextual gain of the program
stored at \(v\).  For the selected parent and operator, the running statistic
is updated by
\begin{equation}
    Q_{\mathrm{op}}(v,o)
    \leftarrow
    Q_{\mathrm{op}}(v,o)
    +
    \frac{
        \Delta_b(\widetilde{\pi}_i)-Q_{\mathrm{op}}(v,o)
    }{
        N_{\mathrm{op}}(v,o)
    }.
    \label{eq:app-operator-update}
\end{equation}
An unavailable-context, generation, validation, or execution failure
increments the attempted-operator count but creates no child and supplies no
synthetic reward.

\paragraph{Marginal credit versus incumbent selection.}
Two scores serve different purposes.  The marginal
\(\Delta_b(\widetilde{\pi}_i)\) is relative to the incumbent team at the
candidate's evaluation time and is used as search credit, a success test for
distillation, and public endorsement.  Incumbent selection instead compares
the absolute value
\[
    \widehat{J}^{\mathrm{mix}}_b
    \left(
        \bar{\boldsymbol{\pi}}_b^{i\leftarrow\widetilde{\pi}_i}
    \right).
\]
This distinction matters after an earlier proposal has improved
\(\bar{\pi}_{i,b}\): marginals produced before and after that update have
different baselines, whereas their absolute team scores remain directly
comparable on the fixed mixed batch.  A valid but non-improving candidate is
retained as an alternative private branch; only an absolute improvement
changes the incumbent.

\paragraph{Executable validation.}
Generated text is reduced to the requested function and supplied with
\texttt{import numpy as np} when needed.  Before execution, the adapter checks
the role-specific contract \(\sigma_i\); strict adapters require the exact
function name, positional argument order, synchronous definition, and
permitted imports.  The program is then compiled and evaluated only through
the benchmark adapter.  The adapter, not the learned code, owns the simulator
state, feasibility checks, and objective computation.  Section~B lists the
exact role signatures and return contracts.  Exceptions, timeouts, malformed
types, invalid identifiers, and incomplete rollouts are handled according to
those documented benchmark contracts and never expose mutable simulator state
to another agent.

\paragraph{Reflect credit map.}
For parent \(\pi(v)\), the trusted orchestrator executes a full team and a
reference team in which role \(i\)'s learned contribution is removed.  For
resource-substitutable settings, the role is absent; for mandatory pipeline
stages, the physical stage remains and uses a fixed non-learned fallback.
The problem adapter computes per-instance full-minus-reference gaps internally
and converts them into a domain-specific summary of dominant, redundant, and
harmful cases.  Only this text
\(g_{i,b}(\pi(v))\) is returned to the role.  Neither teammate code nor the
raw per-instance score vector appears in the Reflect prompt.

\appsubsection{Prompt Construction and Revision Operators}
\label{app:prompts}

All code-generation prompts factor into a shared system instruction, a
role-specific target block, and an operator-specific payload.  This
factorization keeps correctness constraints constant while changing only the
source of guidance \(C^o_{i,b}\) in Eq.~\eqref{eq:generation}.  Angle-bracketed
fields below are filled at runtime.  The human seed is included as a tested
starting point, not as a fixed skeleton or a restriction to parameter tuning.

\begin{tcolorbox}[relicprompt,title={Shared system instruction and target block}]
\small
\textbf{System.} You are an expert algorithm designer. Write a Python
function that serves as a decision-making skill in a multi-agent planning
problem. The function must be correct, concise, self-contained, total, and
deterministic for every contract-valid input. Never raise an exception,
return \texttt{None}, or choose outside a supplied feasible collection. Use
an idle sentinel only when the contract permits it and required work can
still complete. Only use NumPy, imported as \texttt{np}. Do not call
\texttt{np.random}; use only randomness explicitly supplied in the
arguments. No other imports.

\medskip
\textbf{Target block.}\\
\texttt{[Target Function]}\\
\texttt{Company: <agent \(i\)>}\\
\texttt{Signature: <\(\sigma_i\)>}\\
\texttt{Description: <argument and return contract>}\\
\texttt{[Human Seed Baseline]}\\
\texttt{<human-written seed strategy description>}\\
The seed is a tested starting point, not a design constraint. Preserve
correctness and useful invariants, but search beyond constant tuning by
exploiting the full target signature.\\
\texttt{Hint: <role-specific design hint>}
\end{tcolorbox}

\paragraph{Lift prompt.}
Lift samples a revealed teammate principle from
\(\mathcal{A}^{\mathrm{rev}}_{i,b}\); it does not see the source program or a
public-tree score.  The sampled principle and contributor label are recorded
as provenance, while only the principle text affects generation.

\begin{tcolorbox}[relicprompt,title={Lift: instantiate a locally revealed principle}]
\small
\textbf{Inputs:} the shared target block, current parent program
\(\pi(v)\), and sampled \(\rho\in\mathcal{A}^{\mathrm{rev}}_{i,b}\).

\medskip
\texttt{[Principle to Implement]}\\
\texttt{<revealed principle \(\rho\)>}

\texttt{[Current Skill (parent)]}\\
\texttt{<source of \(\pi(v)\)>}

\medskip
Write a \textbf{new} version of the skill that implements the principle
within the target signature. You may retain useful parts of the parent, but
the core logic must express the principle. Return only the complete function,
starting with the exact target signature.
\end{tcolorbox}

\paragraph{Bridge prompt.}
Bridge uses a public node \(p^\star_{i,b}\) selected by
Eq.~\eqref{eq:public-uct}.  Unlike Lift, its principle has downstream transfer
statistics and a public-tree lineage.  A role is never allowed to retrieve a
principle for which it is the recorded contributor.

\begin{tcolorbox}[relicprompt,title={Bridge: revise through a public principle}]
\small
\textbf{Inputs:} the shared target block, current parent program
\(\pi(v)\), and public principle \(\rho(p^\star_{i,b})\).

\medskip
\texttt{[Current Skill (parent)]}\\
\texttt{<source of \(\pi(v)\)>}

\texttt{[Public Principle] (proposed by <contributor>)}\\
\texttt{<\(\rho(p^\star_{i,b})\)>}

\medskip
Write a \textbf{new} version that (1) keeps useful structural elements of
the parent and (2) incorporates the public principle through variables
available under \(\sigma_i\). Return only the complete function, starting
with the exact target signature.
\end{tcolorbox}

\paragraph{Reflect prompt.}
Reflect is self-revision guided by the adapter-authored credit map from the
full-versus-reference comparison.  It is not a free-form request to
``improve'' the code: the prompt explicitly separates cases where the role is
already decisive from cases where its behavior duplicates teammate coverage.

\begin{tcolorbox}[relicprompt,title={Reflect: reshape behavior using contextual credit}]
\small
\textbf{Inputs:} the shared target block, current parent program
\(\pi(v)\), and \(g_{i,b}(\pi(v))\).

\medskip
\texttt{[Current Skill (parent)]}\\
\texttt{<source of \(\pi(v)\)>}

\texttt{[Reflection]}\\
We measured your actual contribution by comparing the team containing your
learned skill with the adapter's fixed reference behavior for this role. The
per-instance gap identifies where you pull weight and where teammates already
cover effectively.\\
\texttt{<domain-specific summary \(g_{i,b}(\pi(v))\)>}\par

\medskip
Write a \textbf{new} version that (1) preserves behavior on dominant cases,
where the role is difficult to replace, and (2) reshapes behavior on
redundant cases by changing coverage, horizon, or priorities. Return only the
complete function, starting with the exact target signature.
\end{tcolorbox}

\paragraph{Output normalization.}
The code extractor accepts either a Python fenced block or raw source,
discards text preceding the requested function, and inserts the NumPy import
if absent.  This normalization is only syntactic: it does not repair
algorithmic logic or silently rewrite the signature.  Candidate validity and
behavior are still determined by the adapter checks and actual rollouts.

\appsubsection{Revelation, Distillation, and Public Memory}
\label{app:principle-lifecycle}

\paragraph{Three distinct knowledge paths.}
RELIC creates textual knowledge through three operations that should not be
conflated:
\begin{enumerate}
    \item \emph{Reveal} abstracts every current incumbent
    \(\bar{\pi}_{i,b}\) for delivery to other agents' local archives.  These
    texts feed Lift but are not automatically public nodes.
    \item \emph{Distillation} applies Eq.~\eqref{eq:distillation} only to the
    strongest positive new node \(v^\star_{i,b}\) from
    Eq.~\eqref{eq:best-new-node}.  It contributes at most one public
    principle per agent per batch.
    \item \emph{Cross-distillation} combines an agent's own incumbent with
    revealed principles from other agents and may contribute up to two
    higher-level public principles.
\end{enumerate}
In every path, raw code stays with its owner and the orchestrator.  The only
cross-agent payload is text in \(\mathfrak{R}\).

\begin{tcolorbox}[relicprompt,title={Reveal: abstract an incumbent for local delivery}]
\small
\texttt{[Source Signature]} \texttt{<\(\sigma_i\)>}\\
\texttt{[Source Code]} \texttt{<\(\bar{\pi}_{i,b}\)>}

\medskip
Extract the core algorithmic principle behind this function. Output only a
concise two-to-five-sentence description of the strategy, independent of
variable names and parameter types. Do not include code.
\end{tcolorbox}

\begin{tcolorbox}[relicprompt,title={Distill: create a transferable public principle}]
\small
You are agent \(i\). The following skill improved team performance by
\(\Delta_b(\pi(v^\star_{i,b}))\).

\texttt{[Skill Code]} \texttt{<\(\pi(v^\star_{i,b})\)>}\\
\texttt{[Existing Public Principles]} \texttt{<public summary>}

\medskip
Extract one principle that is generic, contains no code, variable names, or
signature details, is useful to agents with different signatures, and is not
a duplicate of existing principles. Output:

\texttt{TYPE: STRATEGY or PATTERN or COORDINATION}\\
\texttt{PRINCIPLE: <two-to-four sentences>}
\end{tcolorbox}

\begin{tcolorbox}[relicprompt,title={Cross-distill: infer coordination from principles only}]
\small
You are agent \(i\). Each teammate's best skill has been abstracted into a
principle.

\texttt{[My Best Skill (code)]} \texttt{<\(\bar{\pi}_{i,b}\)>}\\
\texttt{[Other agent -- abstracted principle]} \texttt{<principle text>}\\
\texttt{<repeat for each delivered teammate principle>}

\medskip
You cannot see teammate code. Identify abstract patterns or strategies that
appear across multiple agents and are therefore plausible robust
coordination principles. Output one or two entries, each formatted as:

\texttt{TYPE: STRATEGY or PATTERN or COORDINATION}\\
\texttt{PRINCIPLE: <two-to-four sentences>}
\end{tcolorbox}

\paragraph{Archive delivery and persistence.}
After all incumbents are abstracted, the orchestrator constructs one
batch-level delivery map.  Recipient \(j\) receives only principles from
\(i\neq j\), and the same delivered view is used both to append
\(\mathcal{A}^{\mathrm{rev}}_{j,b}\) and to construct its cross-distillation
prompt.  Archives are append-only during a standard run, so a later Lift can
sample a principle revealed in an earlier batch.  Contributor identity is
bookkeeping metadata; communication experiments may anonymize the displayed
label without changing the true contributor used to prevent self-transfer.

\paragraph{Public-tree lineage.}
A distilled principle normally becomes a child of the public root.  If its
source program was produced by Bridge using public node \(p^\star\), the new
principle is attached below \(p^\star\), recording a refinement relation.
Cross-distilled principles begin new root branches.  The edge therefore
encodes knowledge provenance, not executable control flow and not a claim
that one text logically entails another.

\paragraph{Retrieval, endorsement, and pruning.}
Bridge excludes nodes contributed by its recipient and uses the public UCT
rule in Eq.~\eqref{eq:public-uct}.  Retrieval increments \(N(p^\star)\).
Only a positive contextual gain contributes
\(\max\{0,\Delta_b(\widetilde{\pi}_i)\}\) to \(E(p^\star)\), as in
Eq.~\eqref{eq:endorsement}; a principle is therefore valued by measured
downstream use rather than textual plausibility.  Beginning after the first
two batches, pruning considers only leaves, removes those below endorsement
threshold \(0.5\), and preserves at least five public nodes.  Internal nodes
remain because deleting them would destroy the recorded lineage of surviving
descendants.

\appsubsection{Information Boundary and Reproducibility Contract}
\label{app:information-boundary}

\paragraph{Trusted orchestration.}
The orchestrator is the only component allowed to hold all executable
programs simultaneously.  It compiles candidates, constructs complete or
subset teams, runs the simulator, and returns aggregate scalar scores or
adapter-authored text.  Agent \(i\) may inspect its own parent and seed code,
its callable contract \(\sigma_i\), delivered principles, selected public
text, and its Reflect summary.  It cannot inspect
\(\pi_j\), \(\mathcal{T}_{j,b}\), search statistics, or candidate populations
for \(j\neq i\).  Table~\ref{tab:app-information-boundary} summarizes this
boundary.

\begin{table}[t]
\centering
\caption{Information visible during RELIC training.  ``Orchestrator only''
means the item is used for trusted evaluation but is not inserted into an
agent prompt.}
\label{tab:app-information-boundary}
\begin{tabular}{p{0.39\linewidth}cc}
\toprule
\textbf{Information} & \textbf{Agent \(i\)} & \textbf{Orchestrator} \\
\midrule
Own program and private tree & \(\checkmark\) & \(\checkmark\) \\
Teammate executable programs & -- & \(\checkmark\) \\
Own signature and role description & \(\checkmark\) & \(\checkmark\) \\
Revealed/public principle text & \(\checkmark\) & \(\checkmark\) \\
Teammate private search statistics & -- & -- \\
Aggregate team score & limited feedback & \(\checkmark\) \\
Per-instance full/subset gaps & summary only & \(\checkmark\) \\
Simulator state transitions & runtime observation only & \(\checkmark\) \\
\bottomrule
\end{tabular}
\end{table}

\paragraph{Determinism and controlled randomness.}
The system prompt prohibits uncontrolled imports and
\texttt{np.random}.  When randomized decisions are part of a benchmark
contract, the evaluator supplies deterministic randomness explicitly through
the function arguments (for example, \texttt{u} in the DCOP interfaces of
Section~B.4).  Dataset generation, replay sampling, and communication
interventions use recorded run seeds.  The same candidate team is therefore a
pure function of its programs and stored instance data, apart from the
upstream LLM proposal itself.

\paragraph{Training and deployment.}
Training evaluation is centralized because computing
\(\widehat{J}^{\mathrm{mix}}_b\), subset comparisons, and public endorsement
requires a trusted view of the joint system.  Deployment is decentralized:
the final executable for agent \(i\) is called only through \(\sigma_i\) and
the observations listed in Section~B.  Neither
\(\mathcal{P}_{B+1}\) nor another agent's source code is required at runtime.
The public tree is a learning-time memory, not a centralized execution
controller.

\paragraph{Scope of the privacy claim.}
The boundary above provides implementation non-disclosure between learning
agents.  It does not hide code from the orchestrator, and textual principles
can intentionally reveal strategy-level information.  RELIC therefore makes
no differential-privacy, secure-computation, or cryptographic confidentiality
claim.  Likewise, re-instantiating a principle under \(\sigma_i\) provides no
formal semantic-equivalence guarantee; its usefulness is established only by
the contextual evaluation and endorsement process.

\newpage
\appsection{Experimental Details}
\label{app:experiment-details}

\newcommand{\relicExperimentTableSetup}{%
    \footnotesize
    \setlength{\tabcolsep}{3.5pt}%
    \setlength{\extrarowheight}{1.2pt}%
    \renewcommand{\arraystretch}{1.08}%
}

\appsubsection{Evaluation Metrics and Statistical Aggregation}
\label{app:evaluation-metrics}

\paragraph{Raw benchmark objectives.}
Every candidate team is evaluated by the corresponding simulator from
Section~\ref{app:benchmarks}.  Table~\ref{tab:app-objectives} records the
quantity reported in the paper.  The search implementation uses a single
higher-is-better convention: rewards are retained and costs are negated before
they enter the private-tree return, mixed-team comparison, or public
endorsement.  Tables in the main paper convert the final score back to the raw
benchmark convention, so the arrows there describe the actual task objective.
No metric is normalized using test-set statistics.

\begin{table}[H]
\centering
\caption{Reported objective for every benchmark.  ``Search score'' is the
scalar maximized by the implementation.}
\label{tab:app-objectives}
\relicExperimentTableSetup
\begin{tabularx}{\linewidth}{|
    >{\raggedright\arraybackslash}p{0.14\linewidth}|
    >{\raggedright\arraybackslash}X|
    >{\centering\arraybackslash}p{0.075\linewidth}|
    >{\centering\arraybackslash}p{0.14\linewidth}|}
\hline
Benchmark & Reported quantity & Better & Search score \\
\hline\hline
MAPP-PC & Collected prize of the completed joint plan & $\uparrow$ & \(J\) \\
MAPP-AR & Aggregate reduction in route uncertainty & $\uparrow$ & \(J\) \\
MAPP-DS & Value of dynamically served requests & $\uparrow$ & \(J\) \\
\hline
MAS-JD & Total weighted tardiness & $\downarrow$ & \(-C\) \\
MAS-FS & Total weighted completion time & $\downarrow$ & \(-C\) \\
MAS-ST & Value of tasks completed by their deadlines & $\uparrow$ & \(J\) \\
\hline
HMTSP & Composite heterogeneous multi-tour cost & $\downarrow$ & \(-C\) \\
FSTSP & Composite truck--drone delivery cost & $\downarrow$ & \(-C\) \\
FJAGV & Composite flexible-job-shop and AGV cost & $\downarrow$ & \(-C\) \\
MRCPS & Composite multi-resource project-scheduling cost & $\downarrow$ & \(-C\) \\
\hline
DGC & Graph-coloring conflict cost & $\downarrow$ & \(-C\) \\
DiMES & Soft distributed meeting-scheduling cost & $\downarrow$ & \(-C\) \\
SensorDCSP & Soft coverage and incompatibility cost & $\downarrow$ & \(-C\) \\
SHDS & Energy, preference, and capacity-violation cost & $\downarrow$ & \(-C\) \\
\hline
\end{tabularx}
\smallskip
\parbox{\linewidth}{\footnotesize
\emph{Notation:} \(J\) is a reward retained by the maximization interface;
\(C\) is a reported cost and enters the search as \(-C\).}
\end{table}

\paragraph{Run-level aggregation.}
For one run, the selected three-role team is frozen after the fifth training
batch and evaluated on every held-out instance at a given test size.  Its
reported result is the arithmetic mean over those instances.  We repeat the
complete search three times with independent run seeds and report the mean
\(\pm\) standard deviation of the three run-level means.  Thus, the displayed
standard deviation reflects end-to-end search variability---including
stochastic LLM proposals and seeded sampling---rather than variation among
instances inside one test set.

\paragraph{Relative gap.}
Let \(m_{a,c}\) be method \(a\)'s three-run mean in result column \(c\), and
let \(m^\star_c\) be the best mean in that column.  The gap reported in the
main tables is
\[
  \operatorname{Gap}_{a,c}
  =100
  \begin{cases}
  (m^\star_c-m_{a,c})/(|m^\star_c|+\epsilon), & c\text{ is maximized},\\
  (m_{a,c}-m^\star_c)/(|m^\star_c|+\epsilon), & c\text{ is minimized},
  \end{cases}
\]
where \(\epsilon\) is only a zero-denominator safeguard.  Consequently, the
best method in a column has \(0\%\) gap and smaller is always better,
independently of the raw objective direction.  Rankings, boldface, and gaps
are computed from unrounded means; the table displays rounded values.

\appsubsection{Data Splits and Common Evaluation Protocol}
\label{app:data-evaluation-protocol}

\paragraph{Training and held-out data.}
All methods receive identical serialized instances and split boundaries.
Training uses five ordered batches, with ten fresh instances per batch.  A
candidate in batch \(b\) is evaluated on the current batch together with
replay sampled from the two preceding batches; the held-out test instances
are never used for proposal generation, tree updates, model selection,
reflection, revelation, or public-tree endorsement.  Complete generative
distributions and role interfaces appear in Section~\ref{app:benchmarks};
Table~\ref{tab:app-data-summary} gives the split sizes needed to reproduce the
evaluation protocol.

\begin{table}[H]
\centering
\caption{Common data protocol by benchmark family.  \(5\times10\) denotes
five training batches of ten instances.  Every entry uses a fixed held-out
test split, shared by all methods.}
\label{tab:app-data-summary}
\relicExperimentTableSetup
\begin{tabularx}{\linewidth}{|
    >{\raggedright\arraybackslash}p{0.245\linewidth}|
    >{\centering\arraybackslash}p{0.14\linewidth}|
    >{\centering\arraybackslash}p{0.14\linewidth}|
    >{\raggedright\arraybackslash}X|}
\hline
Family & Training split & Held-out split & Train \(\rightarrow\) test scale \\
\hline\hline
MAPP (3 tasks) & \(5\times10\) & \(20\)/size &
\(40\rightarrow\{50,80\}\) nodes \\
MAS (3 tasks) & \(5\times10\) & \(20\)/size &
\(30\rightarrow\{50,80\}\) jobs/tasks for JD/ST;
\(25\rightarrow\{50,80\}\) jobs for FS \\
Heterogeneous CO (4 tasks) & \(5\times10\) & \(20\)/size &
\(50\rightarrow\{100,200\}\) nodes/tasks \\
DCOP (4 tasks) & \(5\times10\) & \(50\) &
\(120\rightarrow120\) agents; 50 synchronous sweeps \\
\hline
\end{tabularx}
\end{table}

\paragraph{Contextual candidate evaluation.}
The unit of optimization is a role program, but the unit of measurement is a
three-program team.  When method \(a\) proposes \(\widetilde{\pi}_i\), the
evaluator replaces only role \(i\), freezes the current teammates, and
executes the resulting team on the same batch/replay sample used by competing
candidates in that update.  This is the contextual return formalized in
Section~\ref{app:batch-procedure}.  It prevents a role from being rewarded
under an incompatible hypothetical team.  Invalid code, a missing required
return, non-finite output, forbidden mutation, or an execution failure is
assigned the adapter's invalid score and still consumes one proposal.

\paragraph{Randomness control.}
Each run seed initializes Python and NumPy before data/replay sampling and
algorithm construction.  Stochasticity required by a role contract is passed
as ordinary input data---for example, the pre-generated tensor
\texttt{u} in the DCOP interfaces---rather than drawn from hidden global
state inside a candidate.  This makes repeated execution of the same team on
the same stored instance deterministic.  The LLM remains stochastic at
temperature \(1.0\), motivating the three complete runs.

\appsubsection{Compared Methods and Hyperparameters}
\label{app:baseline-details}

\paragraph{Role-isolated adaptations.}
EoH-ind.~\citep{liu2024eoh}, ReEvo-ind.~\citep{ye2024reevo},
MCTS-AHD-ind.~\citep{zheng2025mcts}, and HiFo-ind.~\citep{chen2026hifo}
maintain a separate search state for each role.  A role-specific search sees
its own signature, seed implementation, ancestors, and the contextual team
score obtained while its teammates are frozen; it never sees another role's
code, population, reflection, or search statistics.  This adaptation lets
each method optimize heterogeneous callable interfaces without adding a
communication mechanism that the original method does not provide.

\paragraph{MOTIF control.}
MOTIF~\citep{kiet2026motif} natively supports interacting functions.  We keep
its P1/P2 same-role competitive step, including comparison between candidates
with the same signature, but disable the system-aware cross-role round.
Accordingly, MOTIF is a strong multi-function control while satisfying the
same foreign-code non-disclosure boundary as the independent adaptations.

\paragraph{Human-designed DCOP baselines in Table 2.}
The two ``Human'' rows in Table~2 of the main paper are established,
non-LLM DCOP algorithms.  DSA-B~\citep{zhang2005distributed} is the B variant
of Distributed Stochastic Search: an agent adopts a locally improving
assignment stochastically and may also make an equal-cost lateral move to
escape a plateau.  LSGA~\citep{chen2020genetic} is the Local Search Genetic
Algorithm framework; the reported row uses its DSA-based instantiation
(LSGA-DSA), which augments distributed local search with population-level
crossover and mutation.  We execute both algorithms for the same 50
synchronous sweeps and on the same held-out DCOP instances as the learned
teams.  They are fixed human-designed reference solvers and therefore incur
no LLM calls or proposal budget.  The ``Q1'' annotation in Table~2 identifies
the journal venue of each originating work; it is not an algorithmic variant
or an experimental condition.

\paragraph{Shared settings.}
Table~\ref{tab:app-common-settings} lists parameters held fixed across
methods.  The primary experiments use \texttt{gpt-4o-mini}; the backbone
study substitutes \texttt{deepseek-v4-flash} or
\texttt{qwen3.6-flash} without changing the data, prompts' required output
contracts, proposal budget, or evaluator.  All providers are accessed through
the same OpenAI-compatible wrapper.  No baseline receives a larger candidate
budget when it makes fewer auxiliary LLM calls.

\begin{table}[H]
\centering
\caption{Settings shared by RELIC and all baselines.}
\label{tab:app-common-settings}
\relicExperimentTableSetup
\begin{tabularx}{\linewidth}{|
    >{\raggedright\arraybackslash}p{0.265\linewidth}|
    >{\centering\arraybackslash}p{0.18\linewidth}|
    >{\raggedright\arraybackslash}X|}
\hline
Setting & Value & Interpretation \\
\hline\hline
Roles per team & 3 & One independently callable program per role \\
Training batches & 5 & Ordered batch-level search updates \\
Fresh training instances & 10/batch & Identical serialized instances for every method \\
Replay horizon & 2 batches & Samples are drawn only from the two preceding batches \\
Proposals & 10/role/batch & Invalid proposals are charged to this budget \\
Candidate proposals/run & 150 & \(3\) roles \(\times5\) batches \(\times10\) proposals \\
Independent runs & 3 & Complete search repeated from independently seeded runs \\
Primary backbone & \texttt{gpt-4o-mini} & Same model for every method in a comparison \\
LLM temperature & 1.0 & Used for code generation and auxiliary text calls \\
Model selection & training only & Test instances are evaluated only after team selection \\
\hline
\end{tabularx}
\end{table}

\paragraph{Method-specific search settings.}
We use the local reference implementations with the hyperparameters in
Table~\ref{tab:app-method-settings}.  Values not shown are consequences of
the common protocol rather than method-specific tuning.  In particular, all
population- and tree-based methods still receive only ten new executable
proposals per role and batch.

\begin{table}[H]
\centering
\caption{Method-specific hyperparameters used in the experiments.}
\label{tab:app-method-settings}
\relicExperimentTableSetup
\begin{tabularx}{\linewidth}{|
    >{\raggedright\arraybackslash}p{0.19\linewidth}|
    >{\raggedright\arraybackslash}p{0.25\linewidth}|
    >{\raggedright\arraybackslash}X|}
\hline
Method & Search state & Operators and additional settings \\
\hline\hline
EoH-ind. & Population size \(10\) &
Evolution operators E1, E2, M1, M2, M3 \\
\hline
ReEvo-ind. & Population size \(10\) &
Role-local short and long reflection \\
\hline
MCTS-AHD-ind. & Tree depth \(10\); exploration \(0.1\) &
Widening \(k=2,\alpha=0.5\); actions E1, E2, M1, M2, S1 \\
\hline
HiFo-ind. & Population size \(8\); insight pool \(30\) &
Role-local hindsight and foresight \\
\hline
MOTIF & Inner exploration \(0.01\) &
Reward mix \(0.7\); reward scale \(10\); system-aware round disabled \\
\hline
RELIC & Private-tree depth \(8\); public prune threshold \(0.5\) &
\(c_{\mathrm{node}}=c_{\mathrm{op}}=c_{\mathrm{pub}}=1\);
Lift/Bridge/Reflect chosen by operator UCT \\
\hline
\end{tabularx}
\end{table}

\appsubsection{LLM Invocation Accounting and Compute Fairness}
\label{app:llm-call-accounting}

\paragraph{What is counted.}
We distinguish a \emph{proposal call}, whose response contains a new
executable role program that is charged to the common proposal budget, from an
\emph{auxiliary call}, whose response contains reflection or distilled text
but no scored candidate.  One logical invocation means one algorithm-level
request to the configured LLM.  Provider-internal retries, if any, are not
additional algorithmic decisions and are excluded.  Conversely, a malformed
or invalid candidate still counts as its proposal call.

\paragraph{RELIC call count.}
Let \(R=3\) roles, \(B=5\) batches, and \(H=10\) proposals per role and batch.
Lift, Bridge, and Reflect each use exactly one proposal call, hence RELIC makes
\(RBH=150\) candidate-generation calls.  At the end of every batch, Reveal
abstracts one incumbent per role and cross-distillation produces one
recipient-specific summary per role, adding \(2RB=30\) fixed knowledge calls.
Finally, let
\[
 D_{+}=\sum_{b=1}^{B}\sum_{i=1}^{R}
 \mathbf{1}\{\text{role \(i\) has a newly improved node to distill in batch \(b\)}\}.
\]
Positive-node distillation adds \(D_{+}\) calls, with
\(0\leq D_{+}\leq RB=15\).  The full-communication RELIC run therefore makes
\[
  150+(30+D_{+})=180+D_{+}\in[180,195]
\]
logical LLM invocations.  The value of \(D_{+}\) is run-dependent because an
extra distillation is requested only when a qualifying improvement exists.

\begin{table}[H]
\centering
\caption{Logical LLM invocations per complete three-role run under the
reported \(R=3\), \(B=5\), \(H=10\) protocol.  All methods have the same 150
scored candidate proposals; differences arise only from auxiliary text
operations.}
\label{tab:app-llm-calls}
\relicExperimentTableSetup
\begin{tabularx}{\linewidth}{|
    >{\raggedright\arraybackslash}p{0.17\linewidth}|
    >{\centering\arraybackslash}p{0.10\linewidth}|
    >{\centering\arraybackslash}p{0.10\linewidth}|
    >{\centering\arraybackslash}p{0.10\linewidth}|
    >{\raggedright\arraybackslash}X|}
\hline
Method & Candidate & Aux. & Total & Auxiliary operation \\
\hline\hline
EoH-ind. & 150 & 0 & 150 & None \\
ReEvo-ind. & 150 & 138 & 288 & Role-local short/long reflection after warm-up \\
MCTS-AHD-ind. & 150 & 0 & 150 & None \\
HiFo-ind. & 150 & 15 & 165 & One role-local insight distillation per role and batch \\
MOTIF & 150 & 0 & 150 & System-aware cross-role round is disabled \\
\textbf{RELIC} & \textbf{150} & \(\mathbf{30+D_{+}}\) &
\(\mathbf{180+D_{+}}\) & Reveal, cross-distillation, and conditional
positive-node distillation; \(0\leq D_{+}\leq15\) \\
\hline
\end{tabularx}
\end{table}

\paragraph{Interpretation.}
The experimental control is equal \emph{executable-proposal} compute, not
equal raw API-request count.  Relative to EoH-ind., MCTS-AHD-ind., and MOTIF,
RELIC uses 20--30\% more logical calls to maintain its text-only knowledge
channel; relative to HiFo-ind., it uses 15--30 additional calls.  Even its
maximum of 195 is 32.3\% below ReEvo-ind.'s 288 calls.  The extra RELIC calls
cannot directly improve a score by proposing an additional executable:
Reveal and distillation only produce non-executable principles, and every
method remains capped at 150 evaluated candidates.  This separation is
important because counting only proposal calls explains search opportunity,
whereas the total column better reflects API latency and monetary cost.
Token counts may differ with prompt length and are therefore not inferred
from call counts.

\appsubsection{Scope of Each Experimental Study}
\label{app:study-scope}

Table~\ref{tab:app-study-scope} makes explicit which factor changes in each
reported study.  Unless a row states otherwise, all settings in
Tables~\ref{tab:app-common-settings} and~\ref{tab:app-method-settings} remain
fixed.

\begin{table}[H]
\centering
\caption{Experimental study matrix.  ``Changed factor'' is the only intended
intervention within each comparison.}
\label{tab:app-study-scope}
\relicExperimentTableSetup
\begin{tabularx}{\linewidth}{|
    >{\raggedright\arraybackslash}p{0.205\linewidth}|
    >{\raggedright\arraybackslash}p{0.285\linewidth}|
    >{\raggedright\arraybackslash}X|}
\hline
Study & Benchmarks & Changed factor \\
\hline\hline
Original heterogeneous interfaces &
MAPP-PC/AR/DS; MAS-JD/FS/ST &
Search method under the original A/B/C role signatures \\
\hline
Alternative interfaces &
The same six MAPP/MAS tasks &
All methods receive the alternative A/B/C signatures; data and objectives
are unchanged \\
\hline
Combinatorial generalization &
HMTSP, FSTSP, FJAGV, MRCPS &
Problem family and scale (\(50\) at train; \(100/200\) at test) \\
\hline
Backbone robustness &
DGC, DiMES, SensorDCSP, SHDS &
LLM backbone among GPT-4o-mini, DeepSeek-V4-Flash, and Qwen-3.6-Flash \\
\hline
Component ablations &
MAPP-PC, MAPP-AR, MAS-JD, MAS-ST &
One RELIC knowledge/search component removed or replaced \\
\hline
Communication interventions &
The six heterogeneous MAPP/MAS tasks &
Delivered principle content or cross-role channel altered; proposal budget
and evaluator unchanged \\
\hline
\end{tabularx}
\end{table}

\paragraph{Ablations.}
Component ablations are re-run from scratch rather than applied to a trained
archive.  Removing an operator also removes its opportunity from the
operator-level UCT selector; removing public communication prevents the
corresponding principle from appearing in later prompts.  This preserves the
causal meaning of the intervention and avoids leaving cached information from
the full system.

\paragraph{Communication interventions.}
Misleading-principle and no-peer conditions modify only the payload available
through the public channel.  The recipient still operates under its own role
signature and never receives foreign executable code.  When mitigation is
enabled, the recipient uses its own contextual evidence and Reflect-style
diagnosis to decide whether the delivered principle is consistent with
observed team behavior; principle text is never treated as ground truth.

\appsubsection{Execution Safeguards and Reproducibility}
\label{app:execution-reproducibility}

\paragraph{A shared sandboxed evaluator.}
RELIC and every baseline call the same problem adapter, validator, simulator,
and scoring routine.  Candidate source is parsed and loaded under the required
signature; returned objects are checked for type, shape, range, and finite
values before the simulator consumes them.  The adapter prevents a proposal
from obtaining extra observations by changing its callable interface.  The
DCOP tasks additionally execute exactly 50 synchronous sweeps with a
two-second per-call budget, while heterogeneous combinatorial optimization
uses a \(0.25\)-second per-decision budget.  Task-specific fallback and
feasibility semantics are stated beside each interface in
Section~\ref{app:benchmarks}.

\paragraph{Failure handling.}
Syntax errors, missing functions, exceptions, timeouts, illegal indices,
malformed arrays, non-finite values, and prohibited state mutation cannot
silently enter a population or tree.  They receive the problem adapter's
invalid score and consume their proposal slot.  This policy is identical
across methods and prevents repeated resampling from turning invalid
generations into an unreported compute advantage.

\paragraph{Reproducible artifacts.}
A run configuration records the framework, problem, communication condition,
LLM provider/model/temperature, run seed, proposal budget, replay horizon, and
method hyperparameters.  Stored instance files fix the train/test split and
all simulator-side exogenous randomness.  Search outputs retain the selected
role programs and batch-level evaluation summaries, which are sufficient to
re-evaluate the frozen final team without invoking the LLM.  Reproduction
should compare raw simulator outputs first and derive means, standard
deviations, and relative gaps only afterward, using the definitions in
Section~\ref{app:evaluation-metrics}.

\paragraph{Cost reporting boundary.}
We report logical calls because they are invariant to transient provider
behavior.  Wall-clock time is not used to rank methods: it mixes algorithmic
work with provider queueing, network latency, and heterogeneous generated-code
runtime.  For deployment, only the three selected programs are required.
Reveal, cross-distillation, private/public trees, and all auxiliary LLM calls
are training-time costs; inference itself makes no LLM request.

\clearpage
\appsection{Extended Discussion}
\label{app:extended-discussion}

\lstdefinestyle{relicbestcode}{
    language=Python,
    basicstyle=\scriptsize\ttfamily,
    keywordstyle=\color{relicpromptframe!85!black}\bfseries,
    commentstyle=\color{black!55},
    stringstyle=\color{reliclinkgreen!65!black},
    numbers=left,
    numberstyle=\tiny\color{black!45},
    numbersep=7pt,
    xleftmargin=2.2em,
    framexleftmargin=1.8em,
    frame=single,
    rulecolor=\color{relicpromptframe!75},
    backgroundcolor=\color{relicpromptback!55},
    breaklines=true,
    breakatwhitespace=false,
    columns=fullflexible,
    keepspaces=true,
    showstringspaces=false,
    tabsize=4,
    captionpos=t,
    aboveskip=8pt,
    belowskip=10pt,
    literate={—}{{\textemdash}}1
}

\appsubsection{Reading and Interpreting the Generated Programs}
\label{app:reading-generated-programs}

\paragraph{Selection and presentation.}
This section examines five representative teams obtained from RELIC runs,
chosen to span substantially different coordination structures: uncertain
spatial exploration, pipeline scheduling, heterogeneous vehicle
synchronization, hierarchical project scheduling, and distributed constraint
optimization.  For each problem, we present the final role-A/B/C programs
selected by the search procedure without post-hoc manual modification.  The
listings are self-contained and illustrate both the reusable coordination
patterns induced by RELIC and the idiosyncrasies that remain in
LLM-generated executable heuristics.

\paragraph{Interpretation protocol.}
The listings should be read as empirical outcomes of contextual program
search, not as manually derived algorithms or independently verified optimal
policies.  We therefore examine each team at three levels: the decision signal
used by each role, the coordination pattern shared across roles, and residual
artifacts such as redundant branches or problem-specific constants.  A
plausible-looking expression is not itself evidence of utility.  Its status as
a selected artifact means only that it satisfied the interface and execution
checks in Section~\ref{app:information-boundary} and survived the team-level
evaluation protocol in Section~\ref{app:data-evaluation-protocol}.

\paragraph{Why expose complete programs.}
Aggregate objective values establish whether a method is competitive, but do
not reveal what kind of cooperation was synthesized.  Complete programs make
it possible to distinguish genuinely role-conditioned behavior from three
copies of the same generic heuristic.  They also expose a practically
important property of RELIC: principle exchange encourages compatible
behavior without requiring the final implementations to share state
representations, control flow, or source code.

\appsubsection{Case Study: Adaptive-Reconnaissance Multi-Agent Path Planning}
\label{app:case-mapp-ar}

\paragraph{Qualitative analysis of the LLM-generated programs.}
The three programs recover a common uncertainty-per-travel-cost principle but
instantiate it at different information scopes.  Role A augments the target's
uncertainty with an inverse-distance neighborhood spillover, thereby preferring
regions containing several unresolved nodes.  Role B turns teammate positions
into an explicit division-of-coverage rule: it yields when another robot is
closer and penalizes near-overlap.  With only a distance row and an uncertainty
vector, role C compresses the same idea to a robust value-density score and a
feasible fallback.  The result is behaviorally coherent without textual code
matching; it also exposes typical generated-code artifacts, such as role A's
coarse use of the remaining budget as a neighborhood radius and role C's
largely redundant fallback branch.

\begin{lstlisting}[
    style=relicbestcode,
    caption={MAPP-AR, role A: neighborhood-aware reconnaissance program.},
    label={lst:case-mapp-ar-a}
]
import numpy as np

def select_next_A(current: int, uncertainty_dict: dict, dist_mat: np.ndarray, budget_left: float) -> int:
    """
    Select the next node for the robot to scan based on uncertainty and neighboring influences.
    
    Args:
        current: current node id of my robot.
        uncertainty_dict: {node_id: uncertainty} for nodes not yet scanned.
            Scanned nodes are excluded. Higher = more uncertain.
        dist_mat: (N, N) distance matrix.
        budget_left: remaining travel budget for this robot.
    """
    best_node = 0
    best_score = -1.0
    
    for node, unc in uncertainty_dict.items():
        if node == current or node == 0:
            continue
        
        cost_to_node = dist_mat[current, node]
        total_cost = cost_to_node + dist_mat[node, 0]
        if total_cost > budget_left:
            continue

        neighbor_uncertainty = 0.0
        neighbor_count = 0
        
        for neighbor, neighbor_unc in uncertainty_dict.items():
            if neighbor == node or dist_mat[node, neighbor] < 1e-9:
                continue
            if dist_mat[node, neighbor] < budget_left:
                neighbor_uncertainty += neighbor_unc * (1.0 / (dist_mat[node, neighbor] + 1e-9))
                neighbor_count += 1

        # Combine uncertainty and neighbor influence using spillover effect
        total_score = unc + neighbor_uncertainty
        
        # Normalize score with cost to node
        score_with_cost = total_score / (cost_to_node + 1e-9)

        # Adjust final score based on surrounding uncertainties
        if neighbor_count > 0:
            score_with_cost += 0.1 * neighbor_count
        
        if score_with_cost > best_score:
            best_score = score_with_cost
            best_node = node
            
    return best_node
\end{lstlisting}

\begin{lstlisting}[
    style=relicbestcode,
    caption={MAPP-AR, role B: teammate-aware coverage program.},
    label={lst:case-mapp-ar-b}
]
import numpy as np

def select_next_B(my_state: dict, teammate_positions: list, graph: dict, budget_left: float) -> int:
    current = my_state["current"]
    dist_mat = graph["dist_mat"]
    nodes = graph["nodes"]

    best_node = 0
    best_score = -float("inf")
    
    teammate_set = set(teammate_positions)

    for (node_id, _x, _y, unc) in nodes:
        if node_id == current or node_id == 0 or unc <= 0.0:
            continue
        
        cost = dist_mat[current, node_id] + dist_mat[node_id, 0]
        if cost > budget_left:
            continue
        
        my_dist = dist_mat[current, node_id]

        # Skip if a teammate is heading to a nearby node
        teammate_dist = min(dist_mat[tp, node_id] for tp in teammate_positions) if teammate_set else float("inf")
        if teammate_dist < my_dist:
            continue

        score = unc / (cost + 1e-6)  # Score based on uncertainty relative to cost

        # Deprioritize if a teammate is very close to the node
        if teammate_dist < 2.0 * my_dist:
            score -= 0.5 * unc

        # Update best node if current score is higher
        if score > best_score:
            best_score = score
            best_node = node_id

    return best_node
\end{lstlisting}

\begin{lstlisting}[
    style=relicbestcode,
    caption={MAPP-AR, role C: compressed uncertainty-density program.},
    label={lst:case-mapp-ar-c}
]
import numpy as np

def select_next_C(current: int, dist_row: np.ndarray, uncertainty_vector: np.ndarray, remaining_budget: float) -> int:
    """
    Evaluate potential nodes for scanning based on uncertainty and travel costs,
    prioritizing exploration of regions with high uncertainty while respecting the budget.

    Args:
        current: current node id.
        dist_row: (N,) distances from current to every node.
        uncertainty_vector: (N,) remaining uncertainty at each node
            (0 if scanned). Depot = 0.
        remaining_budget: remaining travel budget.
    
    Returns:
        int: the selected node to scan, or 0 for depot.
    """
    n = len(uncertainty_vector)
    best_node = 0
    best_score = -float("inf")

    for j in range(n):
        if j != current and j != 0:  # Skip current and depot
            travel_cost = dist_row[j]  # Cost to reach j
            if travel_cost <= remaining_budget:
                # Calculate effective score considering uncertainty
                score = uncertainty_vector[j] / (travel_cost + 1e-6)  # Avoid division by zero
                if score > best_score:
                    best_score = score
                    best_node = j

    # If no suitable node was found, opt for the node with the highest uncertainty
    if best_score < 1e-3:
        high_uncertainty_nodes = np.where(uncertainty_vector > 0)[0]
        for j in high_uncertainty_nodes:
            travel_cost = dist_row[j]
            if travel_cost <= remaining_budget:
                best_node = j
                break

    return best_node
\end{lstlisting}

\appsubsection{Case Study: Multi-Agent Flow-Shop Scheduling}
\label{app:case-mas-fs}

\paragraph{Qualitative analysis of the LLM-generated programs.}
All three stages converge to weighted shortest processing time as a stable
local priority, while using their distinct observations to modulate how
aggressively short jobs are favored.  Role A is deliberately compact because
the first stage has no upstream state.  Role B conditions its exponent on
upstream slack and downstream congestion, whereas role C uses the number of
unfinished upstream jobs as a pipeline-pressure signal.  This team is a useful
example of soft coordination: no program constructs a global schedule, yet
their compatible priorities reduce blocking and weighted completion time.
The listings also show that LLM evolution need not remove semantically
redundant branches---two branches in role B compute the same score---when the
surviving program is nevertheless valid and effective in team evaluation.

\begin{lstlisting}[
    style=relicbestcode,
    caption={MAS-FS, role A: first-stage weighted-SPT program.},
    label={lst:case-mas-fs-a}
]
"""
Seed skill for Company A — Myopic-Greedy (Flow-Shop, stage 0).

Lens: local exploitation. Company A owns stage 0 (the first stage).
No upstream; no teammate info. Picks the job in its buffer with the
shortest processing time — classic SPT (shortest-processing-time-first),
adjusted by weight: pick max(w / p).

Conventions:
  - Return -1 to idle.
  - Return a job_id from `my_buffer`.
"""
import numpy as np


def select_next_A(
    my_buffer: dict,
    my_busy_until: float,
    t_now: float,
    dummy: float,
) -> int:
    """
    Weighted shortest-processing-time-first (WSPT).

    Args:
        my_buffer: {job_id: (p_here, w)} for jobs waiting at my stage.
        my_busy_until: equals t_now when called.
        t_now: current simulation time.
        dummy: placeholder (currently unused, reserved).
    """
    if not my_buffer:
        return -1
    best_job = -1
    best_ratio = -1.0
    for jid, (p, w) in my_buffer.items():
        ratio = w / (p + 1e-9)
        if ratio > best_ratio:
            best_ratio = ratio
            best_job = jid
    return best_job
\end{lstlisting}

\begin{lstlisting}[
    style=relicbestcode,
    caption={MAS-FS, role B: congestion-conditioned middle-stage program.},
    label={lst:case-mas-fs-b}
]
import numpy as np

def select_next_B(my_buffer: dict, upstream_busy: float, downstream_buf_size: int, my_busy_until: float) -> int:
    if not my_buffer:
        return -1

    # Identify if we are in a congested downstream or have slack time
    congested_downstream = downstream_buf_size >= 3
    slack_time = upstream_busy > my_busy_until

    best_job = -1
    best_score = -float("inf")

    for jid, (p, w) in my_buffer.items():
        if congested_downstream:
            # If congested downstream, prioritize shorter jobs
            score = w / (p + 1e-9)  # Higher weight per unit time for short jobs
        elif not congested_downstream and not slack_time:
            # Prioritize shorter jobs if both conditions are false (not congested & not slow upstream)
            score = w / (p + 1e-9)
        else:
            # If we have slack and low congestion, prioritize heavier jobs
            score = w / ((p + 1e-9) ** 1.2)

        if score > best_score:
            best_score = score
            best_job = jid

    return best_job
\end{lstlisting}

\begin{lstlisting}[
    style=relicbestcode,
    caption={MAS-FS, role C: pipeline-pressure-aware final-stage program.},
    label={lst:case-mas-fs-c}
]
import numpy as np

def select_next_C(my_buffer: dict, pipeline_state: np.ndarray, job_table: np.ndarray, t_now: float) -> int:
    """
    Selects the next job to process based on a weighted shortest-processing-time-first (WSPT) strategy.

    Args:
        my_buffer: {job_id: (p_here, w)} for jobs waiting at stage 2.
        pipeline_state: (3,) busy-until for stages 0, 1, 2.
        job_table: (N, 5) with columns
            (stage_current, remaining_A, remaining_B, remaining_C, weight).
        t_now: current time.
    """
    if not my_buffer:
        return -1

    # Count jobs in stages 0 and 1 (upstream pressure)
    upstream_jobs = np.sum(job_table[:, 0] < 2)

    # Determine scoring function based on upstream pressure
    if upstream_jobs >= 5:
        # High pressure: prioritize completing jobs quickly
        score_fn = lambda weight, p_here: weight / (p_here + 1e-9)
    else:
        # Low pressure: maximize the weighted completion time
        score_fn = lambda weight, p_here: weight / (p_here ** 1.3)

    best_job = -1
    best_score = -float("inf")

    # Evaluate each job in the buffer
    for job_id, (p_here, weight) in my_buffer.items():
        score = score_fn(weight, p_here)

        if score > best_score:
            best_score = score
            best_job = job_id

    return best_job
\end{lstlisting}

\appsubsection{Case Study: Flying Sidekick Traveling Salesman Problem}
\label{app:case-fstsp}

\paragraph{Qualitative analysis of the LLM-generated programs.}
Here the three roles correspond to different decisions within one synchronized
truck--drone sortie rather than three interchangeable agents.  Role A favors
customers for which drone service is substantially cheaper than a truck
detour and scores feasibility through the slower of the two alternatives.
Role B protects the drone reservation and selects a nearby truck customer,
while role C explicitly minimizes rendezvous makespan and arrival-time
mismatch.  The shared qualitative pattern is therefore synchronization around
the bottleneck vehicle, expressed through three non-isomorphic signatures.
The hard-coded threshold in role B and the unused endurance argument in role C
are characteristic imperfections of generated heuristics; RELIC does not
claim symbolic optimality, and the constraint-owning evaluator prevents such
local choices from violating sortie feasibility.

\begin{lstlisting}[
    style=relicbestcode,
    caption={FSTSP, role A: drone-customer nomination program.},
    label={lst:case-fstsp-a}
]
import numpy as np

def select_next_A(candidate_ids: np.ndarray, drone_roundtrip_times: np.ndarray,
                  truck_detours: np.ndarray, endurance: float) -> int:
    scores = np.maximum(drone_roundtrip_times, truck_detours)
    cost_effectiveness = drone_roundtrip_times < 0.5 * truck_detours
    filtered_scores = scores[cost_effectiveness]
    filtered_ids = candidate_ids[cost_effectiveness]

    if filtered_scores.size > 0:
        min_score_index = np.argmin(filtered_scores)
        if filtered_scores[min_score_index] < endurance:
            return int(filtered_ids[min_score_index])

    min_score_index = np.argmin(scores)
    if scores[min_score_index] < endurance:
        return int(candidate_ids[min_score_index])

    return -1
\end{lstlisting}

\begin{lstlisting}[
    style=relicbestcode,
    caption={FSTSP, role B: truck-customer nomination program.},
    label={lst:case-fstsp-b}
]
import numpy as np

def select_next_B(current: int, candidate_ids: np.ndarray, truck_distances: np.ndarray, drone_customer: int) -> int:
    # Exclude the drone customer from candidates
    valid_candidates = candidate_ids[candidate_ids != drone_customer]
    
    if len(valid_candidates) == 0:
        return -1  # No valid candidates

    # Calculate distances for valid candidates
    valid_distances = truck_distances[valid_candidates]
    current_distance = truck_distances[current]

    # Score calculation based on the principle
    scores = np.maximum(valid_distances, current_distance) + np.abs(valid_distances - current_distance)

    # Find the candidate with the lowest score within the endurance threshold
    endurance_threshold = 10  # Define an endurance threshold for valid scores
    valid_scores = scores[scores <= endurance_threshold]
    
    if len(valid_scores) == 0:
        return -1  # No valid candidates within endurance threshold

    best_candidate_index = np.argmin(valid_scores)
    best_candidate = valid_candidates[scores <= endurance_threshold][best_candidate_index]
    
    return int(best_candidate)
\end{lstlisting}

\begin{lstlisting}[
    style=relicbestcode,
    caption={FSTSP, role C: synchronized rendezvous program.},
    label={lst:case-fstsp-c}
]
import numpy as np

def select_next_C(candidate_ids: np.ndarray, truck_etas: np.ndarray,
                  drone_etas: np.ndarray, endurance: float) -> int:
    differences = np.abs(truck_etas - drone_etas)
    max_times = np.maximum(truck_etas, drone_etas)
    
    # Preserve behavior for dominant cases: minimize the maximum time with a small difference
    scores_preserving = max_times + differences
    best_candidate_preserving = np.argmin(scores_preserving)

    # Reshape behavior for redundant cases: focus on minimizing the maximum ETA
    scores_reshaping = max_times
    
    best_candidate_reshaping = np.argmin(scores_reshaping)
    
    # If the waiting is higher in the reshaping case, use the preserving score
    if scores_reshaping[best_candidate_reshaping] < scores_preserving[best_candidate_preserving]:
        return int(candidate_ids[best_candidate_reshaping])
    else:
        return int(candidate_ids[best_candidate_preserving])
\end{lstlisting}

\appsubsection{Case Study: Multi-Project Multi-Skilled Resource-Constrained Project Scheduling}
\label{app:case-mrcps}

\paragraph{Qualitative analysis of the LLM-generated programs.}
The MRCPS team demonstrates hierarchical specialization across three different
identifier spaces.  Role A balances project due-date pressure against the
amount of unfinished work; role B prefers short tasks but rewards successor
count, so completing a highly connected task can unlock more of the precedence
graph; and role C trades worker skill against cost before emphasizing highly
skilled candidates.  These policies are complementary rather than copies:
they successively decide which project matters, which ready task creates the
most progress, and which worker offers comparative advantage.  Role C's
even-task reservation rule and diversity adjustment are visibly ad hoc, which
underscores why RELIC selects programs by contextual execution instead of
treating fluent LLM rationale as a correctness certificate.

\begin{lstlisting}[
    style=relicbestcode,
    caption={MRCPS, role A: project-priority program.},
    label={lst:case-mrcps-a}
]
import numpy as np

def select_next_A(project_ids: np.ndarray, due_dates: np.ndarray, remaining_task_counts: np.ndarray, t_now: float) -> int:
    # Calculate the tightness of deadlines prioritizing projects with close due dates and less remaining work
    priorities = (due_dates - t_now) + (remaining_task_counts * 4)
    # Find the optimal project id by selecting the one with the minimal adjusted priority that still has remaining tasks
    feasible_indices = np.where(remaining_task_counts > 0)[0]
    selected_index = feasible_indices[np.argmin(priorities[feasible_indices])]
    return int(project_ids[selected_index])
\end{lstlisting}

\begin{lstlisting}[
    style=relicbestcode,
    caption={MRCPS, role B: precedence-aware task-selection program.},
    label={lst:case-mrcps-b}
]
import numpy as np


def select_next_B(project_id: int, task_ids: np.ndarray,
                  durations: np.ndarray, successor_counts: np.ndarray) -> int:
    score = durations - 0.5 * successor_counts
    return int(task_ids[int(np.argmin(score))])
\end{lstlisting}

\begin{lstlisting}[
    style=relicbestcode,
    caption={MRCPS, role C: skill--cost worker-assignment program.},
    label={lst:case-mrcps-c}
]
import numpy as np

def select_next_C(task_id: int, worker_ids: np.ndarray, skill_levels: np.ndarray, worker_costs: np.ndarray) -> int:
    # Calculate scores based on skill levels and costs
    score = skill_levels / np.maximum(worker_costs, 1e-9)

    # Define thresholds for selecting workers
    high_skill_threshold = np.percentile(skill_levels, 75)
    median_cost = np.median(worker_costs)

    # Mask for currently free and qualified workers
    qualified_workers = (skill_levels >= high_skill_threshold)

    # Adjust scoring for reserved tasks focusing on speed advantage
    reserved_tasks = (task_id % 2 == 0)
    speed_advantage_workers = qualified_workers & (worker_costs > median_cost)

    # Modify scores for reserved workers to boost selection
    adjusted_score = np.where(reserved_tasks & speed_advantage_workers, score * 1.5, score)
    
    # Introduce cooperative aspect by highlighting diverse skills
    diversity_factor = (skill_levels - np.mean(skill_levels)) ** 2  # Penalize for low diversity
    adjusted_score += np.where(qualified_workers, diversity_factor, 0)

    final_score = np.where(qualified_workers, adjusted_score, -np.inf)

    # Select the highest score among qualified workers
    return int(worker_ids[np.argmax(final_score)])
\end{lstlisting}

\appsubsection{Case Study: Distributed Graph Coloring}
\label{app:case-dgc}

\paragraph{Qualitative analysis of the LLM-generated programs.}
The DGC programs illustrate decentralized coordination through stochastic
commitment rather than constructive assignment.  Role A increases its move
probability for conflicted vertices with positive repair gain; role B combines
opportunity gain with color flexibility; and role C discourages crowded colors
while damping updates when peer churn is high.  All three use the
evaluator-supplied tensor \(u\) for deterministic tie-breaking and partial
commitment, which prevents every locally similar vertex from moving in
lockstep.  Thus a transferable repair-and-stabilize principle becomes three
role-specific vector policies.  Global tests such as role C's
\texttt{is\_dominant} branch and role B's explicit uniqueness loop also reveal
that the generated code can mix per-variable and batch-level reasoning; strict
shape, domain, determinism, and rollout checks are consequently essential.

\begin{lstlisting}[
    style=relicbestcode,
    caption={DGC, role A: conflict-repair program.},
    label={lst:case-dgc-a}
]
import numpy as np

def select_values_A(
    cur: np.ndarray,
    delta: np.ndarray,
    u: np.ndarray,
    t: int,
    violation_now: np.ndarray,
    repair_gain: np.ndarray,
) -> np.ndarray:
    m, domain = delta.shape
    rows = np.arange(m)

    # Determine potential candidates based on the minimum deltas and tie-breaking
    min_delta = delta.min(axis=1, keepdims=True)
    tied = delta <= min_delta + 1e-9
    candidate_colors = np.where(tied, u[:, :domain], np.inf).argmin(axis=1)

    # Identify conflicted nodes
    is_conflicted = violation_now > 1e-12
    current_gain = repair_gain[rows, candidate_colors]

    # Prioritize high repair_gain for conflicted nodes
    prioritized_gain = current_gain * is_conflicted
    relative_gain = prioritized_gain / (np.abs(violation_now) + np.abs(delta[rows, candidate_colors]) + 1.0)

    # Compute commit probabilities with tie-breaking jitter included
    commit_p = np.clip(0.30 + 0.55 * np.maximum(relative_gain, 0.0), 0.30, 0.85)

    # Adjust decision-making based on collective trends
    mean_gain = np.mean(repair_gain, axis=1)
    collective_influence = (mean_gain[rows] > 0).astype(np.float32)

    # Conditions to switch to a new color
    lateral = (np.abs(current_gain) <= 1e-12) & is_conflicted & (candidate_colors != cur)
    move_conditions = ((prioritized_gain > 0) | lateral) & (u[:, domain] < commit_p * (1 + collective_influence))

    # Select new colors based on move conditions or retain current color
    return np.where(move_conditions, candidate_colors, cur).astype(np.int64)
\end{lstlisting}

\begin{lstlisting}[
    style=relicbestcode,
    caption={DGC, role B: opportunity-and-flexibility program.},
    label={lst:case-dgc-b}
]
import numpy as np

def select_values_B(cur: np.ndarray, delta: np.ndarray, u: np.ndarray, t: int, opportunity_gain: np.ndarray, flexibility: np.ndarray) -> np.ndarray:
    m, domain = delta.shape
    rows = np.arange(m)

    # Current costs and preferred colors based on the available deltas
    current_cost = delta[rows, cur]
    preferred_choice = np.argmin(delta + np.where(delta.min(axis=1, keepdims=True) + 1e-9 > delta, 0, np.inf), axis=1)

    # Calculate potential gains from switching colors
    gain = opportunity_gain[rows, preferred_choice]
    scale = np.ptp(delta, axis=1) + 1.0

    # Commitment probability adjusted with flexibility
    robust = np.clip(flexibility / max(1, domain), 0.0, 1.0)
    commit_p = np.clip(0.35 + 0.35 * np.maximum(gain / scale, 0.0) + 0.15 * robust, 0.30, 0.82)

    # Peer influence to adjust commitment probabilities
    peer_contributions = np.mean(u, axis=1)
    adjusted_commit_p = np.clip(commit_p + 0.1 * (peer_contributions - 0.5), 0.30, 0.82)

    # Identify effective movements and potential cheap recolors
    movement_mask = (gain > 0.1) & (u[:, domain] < adjusted_commit_p)
    cheap_recolor_mask = (gain > 0.1) & (u[:, domain] > 0.5)

    # Select preferred choices for movement based on computed masks
    final_selection = np.where(movement_mask, preferred_choice, cur)

    # Directly apply cheap recoloring while avoiding multiple nodes jumping to the same color in one sweep
    unique_choices = np.unique(preferred_choice[cheap_recolor_mask])
    for choice in unique_choices:
        # Check how many nodes are currently choosing this color
        nodes_for_choice = np.where(preferred_choice == choice)[0]
        if len(nodes_for_choice) > 1:
            # For simplicity, only allow one node to opt for this choice
            selected_node = nodes_for_choice[0]
            final_selection[selected_node] = choice
        else:
            # If only one node prefers this choice, apply the recoloring
            final_selection[nodes_for_choice] = choice

    # Ensure nodes stay with their current color if no valid move was made
    final_selection = np.where((movement_mask | cheap_recolor_mask), final_selection, cur)

    return final_selection.astype(np.int64)
\end{lstlisting}

\begin{lstlisting}[
    style=relicbestcode,
    caption={DGC, role C: churn-aware stabilization program.},
    label={lst:case-dgc-c}
]
import numpy as np

def select_values_C(cur: np.ndarray, delta: np.ndarray, u: np.ndarray, t: int, peer_value_hist: np.ndarray, peer_churn_rate: np.ndarray, local_trend: np.ndarray) -> np.ndarray:
    m, domain = delta.shape
    rows = np.arange(m)

    crowd = peer_value_hist / np.maximum(1.0, peer_value_hist.sum(axis=1, keepdims=True))
    scale = np.maximum(1.0, np.ptp(delta, axis=1))[:, None]
    adjusted = delta + 0.05 * scale * crowd

    tied = adjusted <= adjusted.min(axis=1, keepdims=True) + 1e-9
    choice = np.where(tied, u[:, :domain], np.inf).argmin(axis=1)

    gain = delta[rows, cur] - delta[rows, choice]
    commit_p = np.clip(0.72 * (1.0 - 0.65 * peer_churn_rate) / (1.0 + 0.02 * t), 0.12, 0.72)
    
    peer_count = peer_value_hist.sum(axis=1)
    is_dominant = np.any(gain > 1e-12)

    if is_dominant:
        lateral = (np.abs(gain) <= 1e-12) & (delta[rows, cur] > 1e-12) & (choice != cur)
        move = ((gain > 1e-12) | lateral) & (u[:, domain] < commit_p)

        redundant_move = (u[:, domain] >= commit_p) & (peer_count < np.median(peer_count))

        if np.any(redundant_move):
            alternative_choice = (choice + np.argmax(peer_value_hist, axis=1)) % domain
            adjusted_choice = np.where(redundant_move, alternative_choice, choice)
        else:
            adjusted_choice = choice

        return np.where(move, adjusted_choice, cur).astype(np.int64)

    # For redundant cases, change the behavior to explore
    slow_update = (peer_count >= np.median(peer_count)) & (u[:, domain] < commit_p)
    horizon_adjustment = (cur + 1) % domain
    modified_horizon = np.where(slow_update, horizon_adjustment, cur)

    return modified_horizon.astype(np.int64)
\end{lstlisting}

\appsubsection{Cross-Domain Synthesis}
\label{app:cross-domain-synthesis}

\paragraph{Coordination without a shared code template.}
Across the five domains, RELIC does not force the roles toward a shared code
template.  Instead, reusable ideas reappear in interface-compatible forms:
value density becomes spatial coverage, weighted processing priority becomes
pipeline control, bottleneck minimization becomes truck--drone rendezvous,
urgency becomes hierarchical resource allocation, and controlled commitment
becomes distributed conflict repair.  This is the intended level of transfer:
roles reuse a coordination principle while reconstructing its implementation
under their own callable interface and observation scope.

\paragraph{Three recurring forms of complementarity.}
The programs exhibit three broad mechanisms.  First, \emph{coverage
complementarity} allocates spatial or combinatorial opportunities by relative
access, as in MAPP-AR.  Second, \emph{temporal complementarity} regulates
upstream pressure, downstream congestion, or rendezvous delay, as in MAS-FS
and FSTSP.  Third, \emph{functional complementarity} assigns different
decision layers to different roles, as in MRCPS and DGC.  These mechanisms
cannot generally be obtained by copying one role's executable heuristic into
another because their inputs, return spaces, and responsibilities differ.

\paragraph{Specialization does not imply unnecessary complexity.}
Several selected programs remain close to simple priority rules such as
uncertainty density or weighted shortest processing time.  Their value lies in
where and how the rule is conditioned, not necessarily in syntactic
complexity.  Conversely, longer code is not automatically more coordinated:
some branches are redundant or only weakly connected to the objective.
Contextual team evaluation, rather than program length or explanation
fluency, supplies the selection pressure.

\appsubsection{Failure Modes and the Validation Boundary}
\label{app:discussion-validation}

\paragraph{Observed generated-program failure modes.}
The five teams illustrate benign imperfections that can survive search:
duplicated scoring branches, unused arguments, coarse feasibility proxies,
hard-coded thresholds, and batch-level conditions applied where a
per-decision condition might be cleaner.  More severe candidates can also
raise exceptions, mutate inputs, return identifiers outside the feasible set,
produce arrays of the wrong shape, violate determinism, or exceed the
decision-time limit.  These are failures of a concrete program proposal, not
failures of the textual principle from which it may have been instantiated.

\paragraph{What validation guarantees.}
The trusted evaluator owns state transitions and problem constraints.
Interface checks and sandboxed execution prevent malformed programs from
entering the private search tree, while deterministic repair or rejection
handles invalid actions according to the benchmark contract described in
Sections~\ref{app:benchmarks} and~\ref{app:execution-reproducibility}.  This boundary
guarantees executable compatibility and benchmark feasibility; it does not
prove that a heuristic is optimal, interpretable, or robust outside the
evaluated distribution.

\paragraph{Why team-level testing remains necessary.}
A locally sensible update may duplicate a teammate's behavior, destabilize a
synchronous process, or improve one component while worsening the joint
objective.  RELIC therefore evaluates each candidate in the context of the
current incumbent team.  The case studies make this distinction concrete:
individual expressions are often recognizable heuristics, but their relevance
comes from the complementary role they play in the resulting team.

\appsubsection{Generalization, Deployment, and Limitations}
\label{app:discussion-generalization}

\paragraph{Sources of generalization.}
The generated programs operate on variable-size dictionaries or arrays rather
than memorized instance identifiers, and the shared principles are stated
independently of a teammate's concrete signature.  These design choices
support transfer across held-out instances and the larger problem sizes
described in Section~\ref{app:data-evaluation-protocol}.  Nevertheless, size transfer
within one generator should not be conflated with transfer to a new objective,
constraint family, observation schema, or data distribution.

\paragraph{Deployment profile.}
After training, deployment requires only the three selected programs and the
constraint-owning evaluator.  Public/private trees, principle archives,
revision prompts, and LLM calls are training-time machinery; inference makes
no LLM request.  The resulting controller is therefore inspectable and has a
fixed executable cost, although that cost can still vary with the loops and
array operations synthesized in each program.

\paragraph{Limitations and future directions.}
The qualitative evidence in this section comes from five representative
selected teams and should not be read as a frequency estimate over all runs.
RELIC currently optimizes scalar team feedback, so semantically redundant or
opaque code may survive whenever it performs well enough empirically.
Promising extensions include explicit complexity regularization, behavioral
equivalence tests for removing dead branches, uncertainty-aware selection
across random seeds, and adversarial distribution-shift evaluation.  A further
direction is to augment principle credit with causal interventions that test
whether a shared principle---rather than an unrelated code change---actually
produced the observed team improvement.

\end{document}